\documentclass[journal]{IEEEtran}

\usepackage{bm}
\usepackage{capt-of}
\usepackage{mathtools}
\usepackage{siunitx}
\usepackage[figurename=Figure]{caption}
\usepackage{scrextend}
\usepackage{xspace}
\usepackage{xcolor}
\usepackage{multirow}
\usepackage{wrapfig}
\usepackage{amsmath}
\usepackage{amssymb}
\usepackage{amsthm}
\usepackage{thmtools}
\usepackage{algorithm,algpseudocode}
\usepackage{booktabs}
\usepackage[export]{adjustbox}
\usepackage{url}
\usepackage{subfig}
\usepackage{lineno}
\usepackage{cite}
\usepackage{tabu}

\newcounter{algsubstate}
\renewcommand{\thealgsubstate}{\alph{algsubstate}}

\DeclareMathOperator*{\argmax}{arg\,max}

\DeclareMathOperator*{\minimize}{minimize}

\def\REVISION#1{\textcolor[RGB]{0,0,0}{#1}}

\newtheorem*{theorem*}{Theorem}

\newcommand{\secref}[1]{\mbox{Section~\ref{#1}}}

\newcommand{\figref}[1]{\mbox{Fig.~\ref{#1}}}
\renewcommand{\algref}[1]{\mbox{Algorithm~\ref{#1}}}
\renewcommand{\eqref}[1]{\mbox{Eq.~(\ref{#1})}}

\newcommand{\tabref}[1]{\mbox{Table~\ref{#1}}}

\makeatletter
\DeclareRobustCommand\onedot{\futurelet\@let@token\@onedot}
\def\@onedot{\ifx\@let@token.\else.\null\fi\xspace}
\def\eg{\emph{e.g}\onedot}
\def\ie{\emph{i.e}\onedot}

\def\etal{\emph{et al}\onedot}
\makeatother

\newcolumntype{L}[1]{>{\raggedright\arraybackslash}p{#1}}
\newcolumntype{C}[1]{>{\centering\arraybackslash}p{#1}}
\newcolumntype{R}[1]{>{\raggedleft\arraybackslash}p{#1}}

\ifCLASSOPTIONcompsoc
  \usepackage[nocompress]{cite}
\else
  \usepackage{cite}
\fi

\begin{document}

\title{Learning to Predict Gradients for\\ Semi-Supervised Continual Learning}

\author{Yan~Luo,~\IEEEmembership{Member,~IEEE,}
        Yongkang~Wong,~\IEEEmembership{Member,~IEEE,}
        Mohan~Kankanhalli,~\IEEEmembership{Fellow,~IEEE,}
        and~Qi~Zhao,~\IEEEmembership{Senior Member,~IEEE,}
\IEEEcompsocitemizethanks
			{
                \IEEEcompsocthanksitem Y. Luo is with the Harvard Ophthalmology AI Lab, Harvard University, Boston, MA, 02114. This work was done when Y. Luo was at the University of Minnesota. E-mail: yluo16@meei.harvard.edu
			\IEEEcompsocthanksitem Q. Zhao is with the Department of Computer Science and Engineering, University of Minnesota, Minneapolis, MN, 55455. E-mail: qzhao@cs.umn.edu
			\IEEEcompsocthanksitem Y. Wong and M. Kankanhalli are with the School of Computing, National University of Singapore, Singapore, 117417. E-mail: yongkang.wong@nus.edu.sg, mohan@comp.nus.edu.sg
			\IEEEcompsocthanksitem Qi Zhao is the corresponding author.
			}
}

\markboth{IEEE TRANSACTIONS ON NEURAL NETWORKS AND LEARNING SYSTEMS,~VOL.~xx, NO.~x, August~202x}%
{Shell \MakeLowercase{\textit{et al.}}: Bare Demo of IEEEtran.cls for Computer Society Journals}

\IEEEtitleabstractindextext{%
\begin{abstract}

A key challenge for machine intelligence is to learn new visual concepts without forgetting the previously acquired knowledge. 
Continual learning is aimed towards addressing this challenge.
However, there still exists a gap between continual learning and human learning. In particular, humans are able to continually learn from the samples associated with known or unknown labels in their daily life, whereas existing continual learning and semi-supervised continual learning methods assume that the training samples are associated with known labels.
Specifically, we are interested in two questions: 1) how to utilize unrelated unlabeled data for the semi-supervised continual learning task, and 2) how unlabeled data affect learning and catastrophic forgetting in the continual learning task.
To explore these issues,
we formulate a new semi-supervised continual learning method, which can be generically applied to existing continual learning models.
Furthermore, we propose a novel gradient learner to learn from labeled data to predict gradients on unlabeled data. In this way, the unlabeled data can fit into the supervised continual learning framework.
We extensively evaluate the proposed method on mainstream continual learning methods, adversarial continual learning, and semi-supervised learning tasks. The proposed method achieves state-of-the-art performance on classification accuracy and backward transfer in the continual learning setting while achieving desired performance on classification accuracy in the semi-supervised learning setting.
This implies that the unlabeled images can enhance the generalizability of continual learning models on the predictive ability on unseen data and significantly alleviate catastrophic forgetting.
The code is available at \url{https://github.com/luoyan407/grad_prediction.git}.


\end{abstract}

}

\maketitle

\IEEEdisplaynontitleabstractindextext

\IEEEpeerreviewmaketitle


\section{Introduction}\label{sec:introduction}

Continual learning (CL) models observe sets of labeled data through a sequence of tasks~\cite{Ring_Thesis_1994,Thrun_IRS_1995}. 
The tasks may vary over time, \eg~images with novel visual concepts (\ie~classes) or addressing different problems from the previous tasks~\cite{Schlimmer_AAAI_1986}. 
CL is analogous to human learning.
Humans are able to continually acquire, adjust, and transfer knowledge and experiences throughout their lifespan. 
The key challenges are two-fold. 
First, the learning models can abruptly forget previously absorbed knowledge while learning new information in novel tasks, \ie~suffer from catastrophic forgetting~\cite{Mccloskey_PLM_1989}. Second, how to employ the knowledge learned from previous tasks to quickly adapt to novel tasks. 

\begin{figure}[!t]
	\centering
	\includegraphics[width=1\columnwidth]{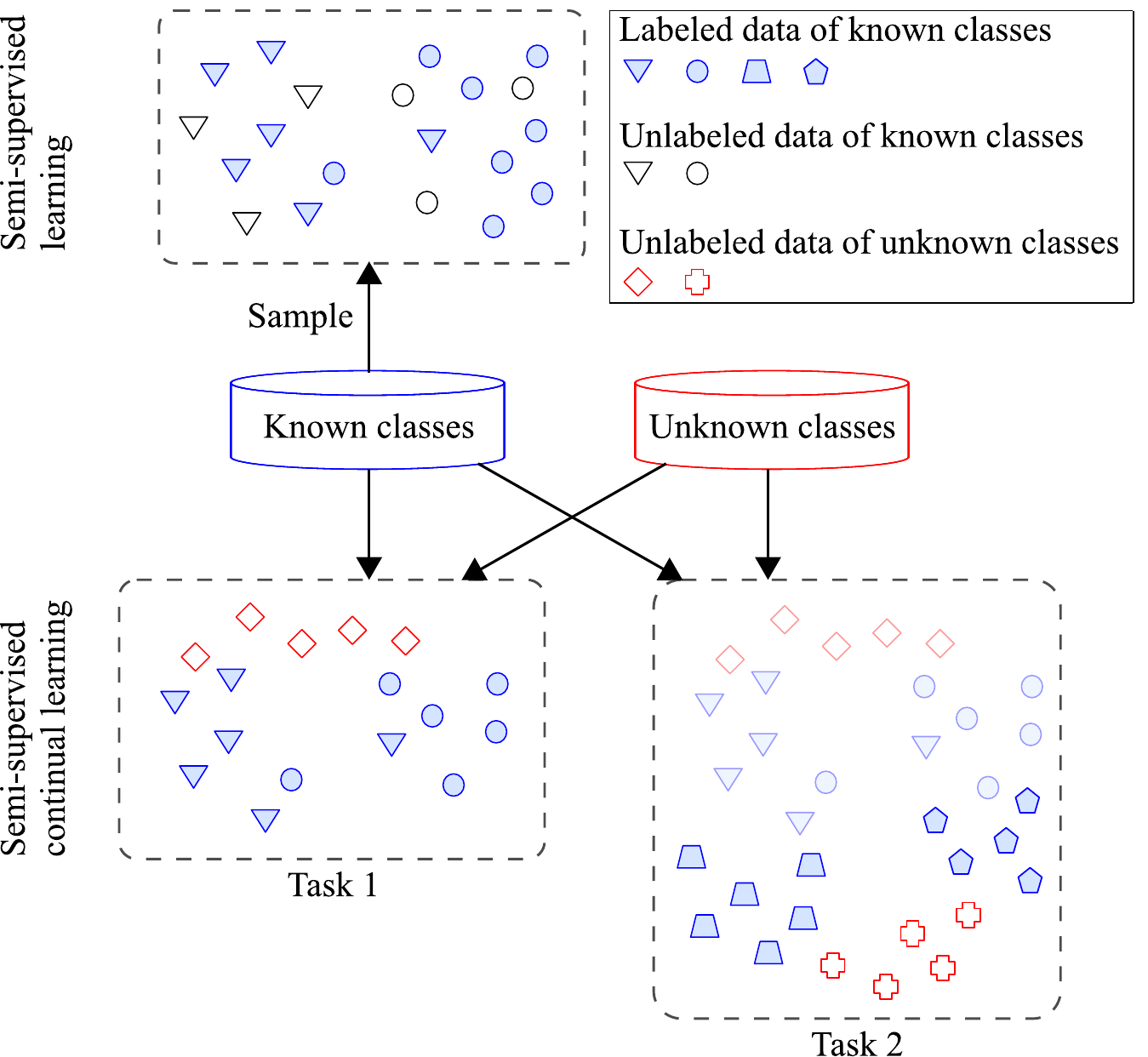}
	\caption{\label{fig:diff}
	    Conceptual comparison between the challenge in the novel \textit{semi-supervised continual learning} (SSCL) problem and the one in the \textit{semi-supervised learning} (SSL) problem.
        The key difference is that the underlying classes w.r.t. the unlabeled data could be unknown in the SSCL problem, while the ones in SSL are assumed to be from the known classes.
        To suitably adapt to the continual learning paradigm, we do not impose such a constraint in the novel SSCL problem.
        Instead, the underlying labels of unlabeled data can be from either known classes or unknown classes.
        The faded-out samples in the task 2 indicate that the samples in the task 1 are not available in the task 2 according to the protocol.
	}
\end{figure}

Previous CL methods presume that the labels associated with the data are known \cite{Lopez_NIPS_2017,Shin_NIPS_2017,Cuong_ICLR_2018,Riemer_ICLR_2018,Serra_ICML_2018,Sinha_ICCV_2019,Lee_2019_ICCV}. This assumption may be divergent from human learning, where a considerable amount of labels associated with the unlabeled data could be novel and unrelated to the known labels. Furthermore, large-scale labeled data may not always be available due to the limits of labor-intensive and expensive human annotations. Moreover, the classes of a task are distinct from the ones in the other tasks, or the task's labels may be of a different form, \eg~category vs. bounding box. Therefore, we do not presume any constraint that restricts the correlation between the labels associated with unlabeled data and the ones associated with labeled data. Instead, the unlabeled data could have either the same or different class labels as the labeled data, which is shown in \figref{fig:diff}.
As a result, the fundamental challenge lies in the generalizability of learning in this semi-supervised continual learning (SSCL) setting. The CL models do not only generalize the knowledge learned from preceding tasks to the current task, but also should leverage unlabeled data that are associated with unknown labels to boost the learning process.

The labels that are known to the learning process play an important role in an end-to-end learning paradigm, even in the semi-supervised learning setting. Through the labels and the pre-defined loss functions, the gradients are computed to back-propagate to neurons in each layer. This gradient-based learning process is key to updating the models to make more precise predictions \cite{Chaudhry_ICLR_2018,Chaudhry_arXiv_2019,Ebrahimi_ECCV_2020,Lopez_NIPS_2017,Luo_TPAMI_2019,Cuong_ICLR_2018,Riemer_ICLR_2018,Serra_ICML_2018,Shin_NIPS_2017,Sinha_ICCV_2019}. However, when the underlying labels of unlabeled data are unknown, it is a challenge to generate the gradients that improve the generalizability of models in the SSCL setting.



Conventionally, pseudo labeling, \ie~predicting labels by a teacher network for the unlabeled data and taking them as the ground-truth labels for training a student network, is widely used for semi-supervised learning (SSL)~\cite{Chen_ICML_2020,Chen_NIPS_2020,Lee_ICMLW_2013,Xie_CVPR_2020}. 
However, it may not work in the SSCL setting as the classes of a task are distinct from the ones in the other tasks or the task's labels may be of a different form.
In contrast to the pseudo labeling methods, learning to predict pseudo gradients on unlabeled samples is straightforward and effective as predicting labels is skipped. Moreover, the pseudo gradients are aligned with the knowledge learned from samples in various categories, while the gradients generated by pseudo labeling methods are aligned with a specific category as an unlabeled sample is conventionally labeled as a category in the CL setting. 

\begin{figure}[!t]
	\centering
	\includegraphics[width=1.0\columnwidth]{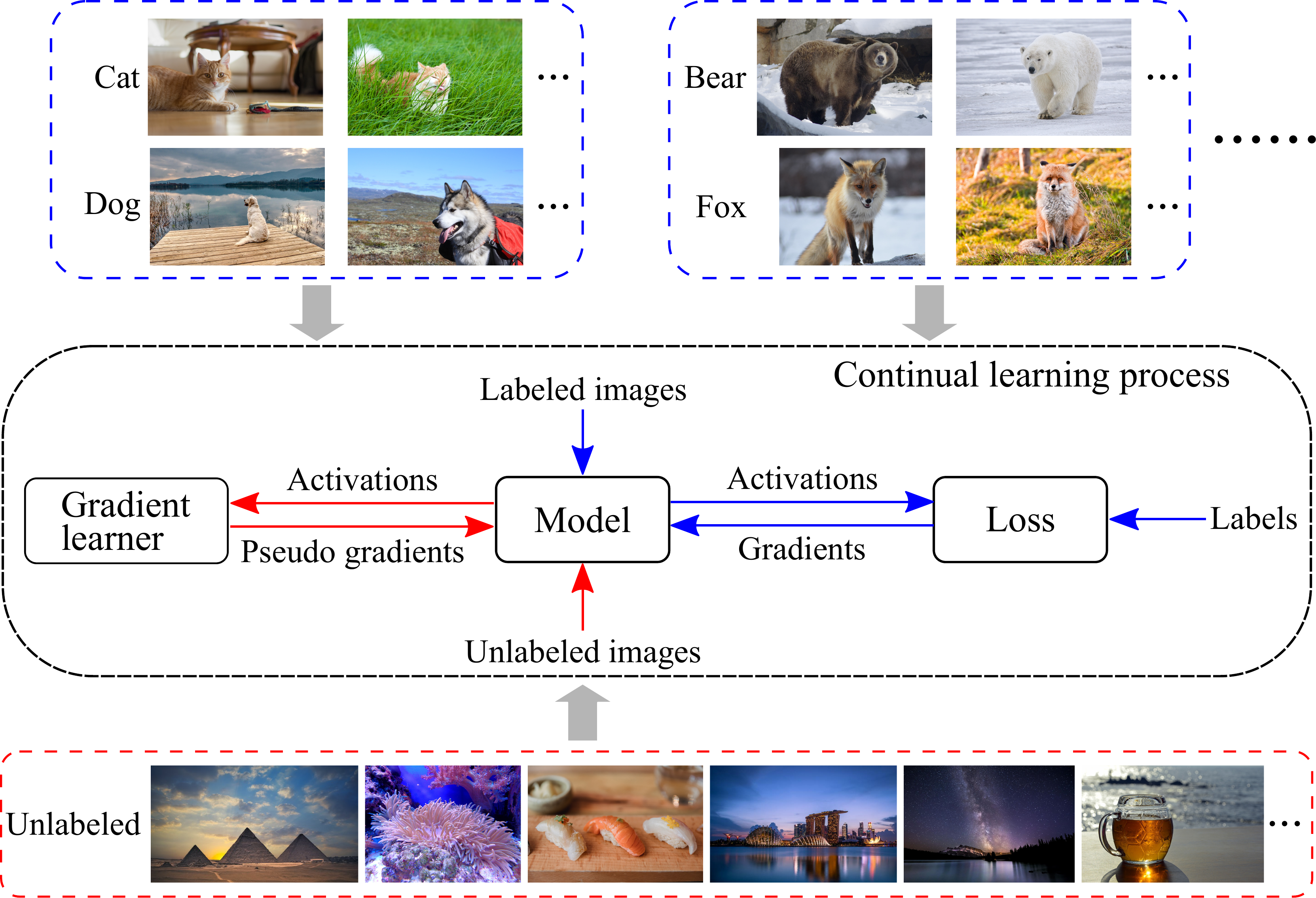}
	\caption{\label{fig:teaser}
	    Problem of semi-supervised \mbox{continual learning}. Conventional supervised continual learning requires labels to compute gradients for model update (see \textcolor{blue}{blue flows}). In contrast, this work proposes to \textit{predict gradients} so that the unlabeled images can be incorporated in the continual learning paradigm for better generalizability (see \textcolor{red}{red flows}).
	}
\end{figure}

To utilize unlabeled data into the supervised CL framework, we propose a novel gradient-based learning method that learns from the labeled data to predict \textit{pseudo gradients} for the unlabeled data, as shown in \figref{fig:teaser}.
Specifically, a novel gradient learner learns the mapping between features and the corresponding gradients generated with labels.
We follow \cite{Ebrahimi_ECCV_2020,Lopez_NIPS_2017} to conduct extensive experiments on CL benchmarks, \ie~MNIST-R, MNIST-P, iCIFAR-100, CIFAR-100, and miniImageNet.
To verify the generalization ability of the proposed method, we follow \cite{Zhang_AISTATS_2021} to evaluate the proposed method on SVHN, CIFAR-10, and CIFAR-100.
The main contributions of this work are summarized as follows.
\begin{itemize}
    \item We propose a novel semi-supervised continual learning method that leverages the rich information from unlabeled data to improve the generalizability of CL models.
    \item We propose a learning method that learns to predict gradients for unlabeled data. To the best of our knowledge, this is the first work that generates pseudo gradients without ground-truth labels.
    \item Extensive experiments and ablation study show that the proposed method improves the generalization performance on all metrics.
    This implies that learning with unlabeled data is helpful for improving the predictive ability and alleviating catastrophic forgetting of CL models.
    \item We provide empirical evidence to show that the proposed method can generalize to the SSL task.
\end{itemize}

\section{Related Work}
\label{sec:related}

\subsection{Continual Learning}


Continual learning (CL) is a learning paradigm where a model learns through a sequence of tasks \cite{Ring_Thesis_1994,Thrun_IRS_1995,Thrun_Springer_1998}. CL is a branch of online learning \cite{Bottou_NIPS_2004,Hoi_JMLR_2014}, where the challenge is to balance the retention of knowledge from preceding tasks with the acquisition of new information for future tasks. However, catastrophic forgetting is a common issue in this paradigm \cite{Mccloskey_PLM_1989}. There are three types of CL, namely task-incremental learning, domain-incremental learning, and class-incremental learning \cite{Ven_arXiv_2019}.
Task-incremental learning solves a sequence of distinct tasks, when the task ID is provided along the process.
Domain-incremental learning adapts to changing input data distributions while maintaining performance on the original task.
Class-incremental learning solves a sequence of distinct tasks and infers the task ID.
Compared to domain-incremental learning, task-incremental learning and class-incremental learning emphasize recognizing and classifying new classes without forgetting previous knowledge, which are closely related to the proposed method.

There are a number of works that can be cast into the category of  task-incremental learning~\cite{Chaudhry_ICLR_2018,Ebrahimi_ECCV_2020,Kirkpatrick_PNAS_2017,Lopez_NIPS_2017,Luo_TPAMI_2019,Cuong_ICLR_2018,Rebuffi_CVPR_2017,Riemer_ICLR_2018,Chaudhry_NeurIPS_2020}. Specifically, Lopez-Paz and Ranzato propose a memory-based method, namely GEM, to impose a constraint on the gradients w.r.t to the training samples and the memory~\cite{Lopez_NIPS_2017}. Along the same line, Luo~\etal introduce a gradient alignment method DCL that enhances the correlation between the gradient and the accumulated gradient~\cite{Luo_TPAMI_2019}. Recently, Ebrahimi~\etal propose an adversarial continual learning (ACL) approach that aims to factorize task-specific and task-invariant features simultaneously~\cite{Ebrahimi_ECCV_2020}. Unlike GEM, where the training samples are observed one by one, the training process of ACL would repeat multiple times on every task. All the aforementioned works follow supervised continual learning paradigm that requires the ground-truth labels. However, how unlabeled data may influence the continual learning problem remain unclear. In this work, we propose the SSCL paradigm, where the model occasionally observes unlabeled data. 
The class-incremental learning problem aims to learn visual concepts in new tasks while retaining the visual concepts learned in the previous tasks~\cite{Rebuffi_CVPR_2017,Lee_2019_ICCV}.
Correspondingly, the samples in the coresets would be repeatedly observed in this problem, whereas CL only observes each sample once.
In this setting, Carvalho \etal introduce a catastrophic forgetting solution based on knowledge amalgamation (CFA) that learns a student network from multiple heterogeneous teacher models.
Lee \etal leverage unlabeled data with a knowledge distillation method to boost the class-incremental learning \cite{Lee_2019_ICCV}.
Notably, the experimental protocol in \cite{Lee_2019_ICCV} is different from that of SSCL. The class-incremental learning with unlabeled data maintains three sets of samples through the learning process, \ie the samples that have been seen in the previous tasks, the samples that are related to the current task and have not been seen before, and the unlabeled samples are selected by a confidence-based strategy from a data pool. In contrast, SSCL only observes the samples that are related to the current task and the unlabeled samples randomly selected from the data pool.
For a fair comparison, we utilize the knowledge distillation method in \cite{Lee_2019_ICCV} to generate pseudo labels for unlabeled samples as the baselines.
This work follows the experimental protocols used in GEM \cite{Lopez_NIPS_2017} and ACL \cite{Ebrahimi_ECCV_2020}, which are widely-adopted task-incremental learning schemes.

Except for the aforementioned methods, the task incremental learning problem and class incremental learning problem can be solved by Dark Experience Replay (DER) \cite{Buzzega_NeurIPS_2020} and eXtended-DER (X-DER) \cite{Boschini_arXiv_2022} simultaneously. DER exploits a buffer (\ie dark experience) storing data from previous tasks to train a student model. X-DER leverages memory update and future preparation to improve DER.

From the perspective of the strategies, the common strategies tackling the continual problem can be divided into three categories: rehearsal-based, regularization-based, and knowledge distillation-based methods.
Rehearsal-based methods address catastrophic forgetting by replaying training samples stored in a memory buffer \cite{Hou_CVPR_2019,Buzzega_NeurIPS_2020}. In contrast, regularization-based methods prevent catastrophic forgetting by regularizing the model's parameters so that they do not change much when new data is presented \cite{Buzzega_NeurIPS_2020}. Moreover, knowledge distillation can be used to prevent catastrophic forgetting by transferring knowledge from a previous model (teacher) to a new model (student) \cite{Hou_CVPR_2019,Buzzega_NeurIPS_2020}.

\subsection{Semi-supervised Learning (SSL)}


\REVISION{
Semi-supervised learning, a machine learning technique, involves training a model using both labeled and unlabeled data \cite{Fierimonte_TNNLS_2017,Duan_TNNLS_2022,Cui_TNNLS_2023,Zhu_TNNLS_2023}. This task aims to utilize a small set of labeled data along with a larger set of unlabeled data, enabling the model to establish connections and make predictions on unseen data.
For example, Fierimonte \etal propose a fully decentralized approach to semi-supervised learning using privacy-preserving matrix completion, specifically addressing the challenge of distributed learning \cite{Fierimonte_TNNLS_2017}. Duan \etal introduce a novel method that incorporates low-confidence samples into semi-supervised learning through mutex-based consistency regularization \cite{Duan_TNNLS_2022}. Another approach by Yang \etal leverage a contrastive learning-based loss function and augmented samples generated via an interpolation-based approach to guide training~\cite{Yang_TNNLS_2022}.
}

Existing methods are mainly based on pseudo-labeling or self-training, \ie~leveraging the labeled data to predict artificial labels for the unlabeled data~\cite{Scudder_TIT_1965,Shahshahani_TGRS_1994}. 
Most modern deep learning based models follow this line of research~\cite{Chen_NIPS_2020,Lokhande_CVPR_2020,Ren_NIPS_2020,Sohn_NIPS_2020,Zhang_AISTATS_2021}. 
Particularly, the noisy student model~\cite{Xie_CVPR_2020} employs the teacher-student method to train on ImageNet~\cite{Deng_CVPR_2009} with unlabeled images from JFT~\cite{Hinton_arXiv_2015}, which is an in-house dataset at Google and has 100 million labeled images with 15,000 labels, to achieve state-of-the-art performance.
In addition, Zhang \etal propose a meta-objective to alternatingly optimize the weights and the pseudo labels such that the learning process can leverage unlabeled data \cite{Zhang_AISTATS_2021}.
\REVISION{To utilize the abundant unlabeled data, these semi-supervised learning models assign predicted labels to unlabeled data to generate gradients for back-propagation. In contrast, the proposed method instead predict pseudo gradients for back-propagation, bypassing the need for a loss function with pseudo-labeled data.}
Different from conventional (semi-)supervised learning, where visual concepts are unchanging during the learning process, the visual concepts of a task in CL are different from that of the other tasks through the whole learning process.
As a result, unlabeled data that are labeled as known classes would break the protocol of the split of classes in various tasks of CL \cite{Lopez_NIPS_2017}.
Instead, CL is in favor of a more generic hypothesis of unlabeled data, that is, the underlying labels of unlabeled data could be unknown.
A natural choice is to sample unlabeled images from external datasets, rather than treating training images as unlabeled images. 
In conventional semi-supervised continual leanring (SSCL) \cite{Zhang_AISTATS_2021}, the visual concepts that are related to the unlabeled samples are presumed to be known for computing gradients. Different from SSCL, the proposed SSCL in this work does not require this hypothesis. As a result, without known labels, it is unable to compute the gradients for back-propagating the errors. Instead of computing the gradients with the labels, we study how to predict the pseudo gradients by measuring the suitability between unlabeled samples and predicted pseudo gradients in learning a certain visual concept.

\begin{figure*}[!t]
	\centering
	\includegraphics[width=0.96\linewidth]{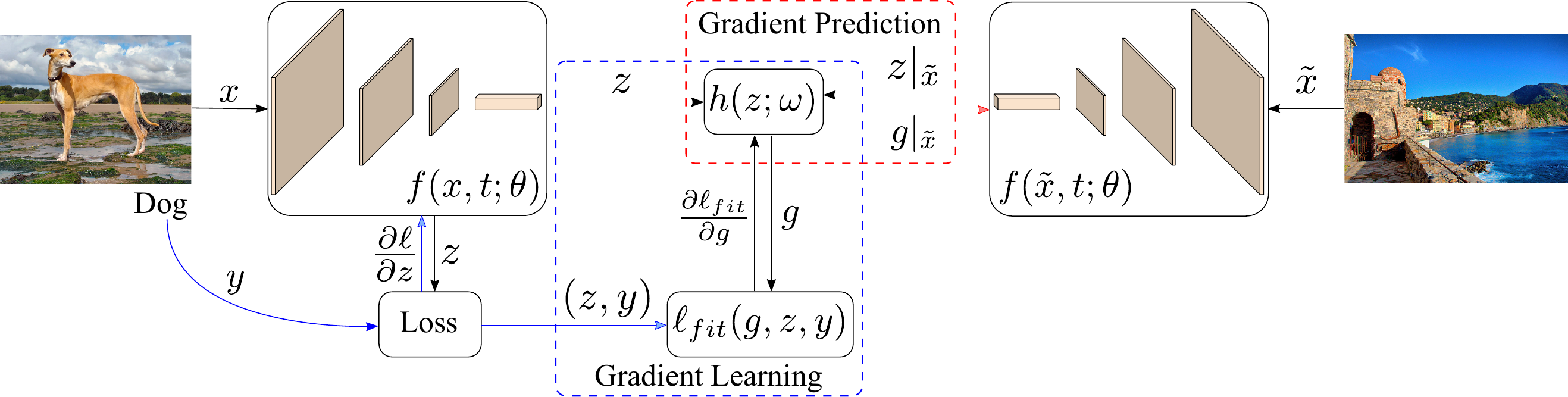}
	\caption{\label{fig:framework}
	    Overview of the proposed gradient learning and gradient prediction process with the gradient learner $h(\cdot;\omega)$. The backbone network is shared between the two processes.}
\end{figure*}

\subsection{Gradient-based Methods}

Gradient-based methods refer to a family of optimization algorithms used to find the parameters of a machine learning model that minimize a certain objective function \cite{Bottou_COMPSTAT_2010}. These methods rely on computing the gradients of the objective function with respect to the model parameters and using those gradients to iteratively update the parameters until convergence. The gradient is a measure of how the loss function changes as a function of the model's parameters \cite{Andrychowicz_NIPS_2016,Lopez_NIPS_2017,Luo_TPAMI_2019,Ren_NIPS_2020}.


The learning process is composed by the forward propagation and back-propagation. 
Jaderberg \etal propose a learning framework with the synthetic gradients to allow layers to be updated in an asynchronous fashion \cite{Jaderberg_ICML_2017}. The proposed pseudo gradients can be used as ground-truth gradients in such a learning framework when the labels of training images are missing.
In particular, \cite{Lopez_NIPS_2017,Luo_TPAMI_2019} use the information of gradients to form a constraint to improve the performance of CL. 
\cite{Andrychowicz_NIPS_2016} have a similar flavor, that is, they aim to learn an optimizer to adaptively compute the step length for the vanilla gradients.
\REVISION{In contrast to these gradient-based methods that rely on labeled data to compute gradients, the proposed method predicts pseudo gradients by maximizing their fitness within the loss function applied to labeled data.}



Stochastic optimization methods often use gradients to update a model's parameters. 
In the literature, stochastic gradient descent (SGD)~\cite{Robbins_AMS_1951} takes the anti-gradient as the parameters' update for the descent, using the first-order approximation~\cite{Boyd_CUP_2004}. 
In a similar manner, several first- and second-order methods are devised to guarantee convergence to local minima under certain conditions~\cite{Jin_ICML_2017,Carmon_SIAM_2018,Reddi_ICAIS_2018}. 
Nevertheless, these methods are computationally expensive and may be not feasible for learning settings with large-scale high-dimensional data.
In contrast, adaptive methods, such as Adam~\cite{Kingma_arXiv_2014}, RMSProp~\cite{Hinton_RMSProp_2012}, and Adabound~\cite{Luo_ICLR_2018}, show remarkable efficacy in a broad range of machine learning tasks~\cite{Hinton_RMSProp_2012,Luo_ICLR_2018}. 
Moreover, Zhang~\etal propose an optimization method that wraps an arbitrary optimization method as a component to improve the learning stability~\cite{Zhang_NIPS_2019}. 
These methods are contingent on vanilla gradients to update a model. 
In this work, we study how the predicted gradients influence the learning process.

\section{Problem Set-Up}

The training process of supervised learning methods generally requires a training dataset {\small $D_{tr}=\{(x_{i},y_{i})\}_{i=1}^{i}$} that consists of samples {\small $s_{i} = (x_{i},y_{i})$}, where $x_{i}\in \mathcal{X}$ represents a sample and $y_{i}\in \mathcal{Y}$ represents a target vector, where $\mathcal{Y}$ is the target label space.
The samples presumably are identically and independently distributed variables that follow a fixed underlying distribution $\mathcal{D}$ \cite{Lopez_NIPS_2017}.
With all samples, supervised learning methods attempt to find a model {\small $f: \mathcal{X} \xrightarrow{\theta} \mathcal{Y}$} to map feature vectors to the target vectors, where $\theta$ are the parameters of $f$.
In contrast to supervised learning, SCL is more human-like and will observe the continuum of data 
\begin{align*}
    D_{tr} = \{(x_{i}, t_{i}, y_{i})| (x_{i}, y_{i})\sim \mathcal{D}_{t_{i}}, t_{i} \in \mathcal{T} \} ,
\end{align*}
where $t_{i}$ indicates the $i$-th task and $\mathcal{T}$ is a set of tasks. A task is a specific learning problem.
Different from supervised learning, which has a fixed distribution, each task is associated with an underlying distribution in the SCL setting.
The SCL models are defined as {\small $f: \mathcal{X} \times \mathcal{T} \xrightarrow{\theta} \mathcal{Y}$}.
Correspondingly, the loss of SCL is defined as 
\begin{align}
    \mathcal{L}(f_{\theta}, D_{tr}) = \frac{1}{|D_{tr}|} \sum_{(x_{i}, t_{i}, y_{i})\in D_{tr}} \ell(f_{\theta}(x_{i}, t_{i}), y_{i}),
    \label{eqn:scl}
\end{align}
where $f(\cdot; \theta)$ is simplified as $f_{\theta}(\cdot)$. 
With the loss function $\ell$ and a training sample $(x_{i},t_{i},y_{i})$, the gradient can be computed, \ie~{\small $\frac{\partial \ell}{\partial z_{i}} \frac{\partial z_{i}}{\partial \theta}$}, where {\small $z_{i} = f_{\theta}(x_{i}, t_{i})$}.
Typically, $\ell$ is the cross entropy loss in the classification task.
Note that we follow the convention of classification literature \cite{Krizhevsky_NIPS_2012,He_CVPR_2016,Tan_ICML_2019} to define the input of $\ell$ as logits $z$ and ground-truth labels $y$, instead of predicted labels and ground-truth labels.
Finally, the model is updated with the computed gradient, that is,
\begin{align}
    \theta \leftarrow \theta - \eta \frac{\partial \ell}{\partial z_{i}} \frac{\partial z_{i}}{\partial \theta},
    \label{eqn:org_update}
\end{align}
where $\eta$ is the learning rate for updating $\theta$. Let $\Omega(X)$ be the set of classes associated with all labeled data $X$, and $\Omega(\tilde{X})$ be the set of classes associated with all unlabeled data $\tilde{X}$. We assume that $	\Omega(X)\subset\Omega(\tilde{X})$. In other words, the underlying labels associated with unlabeled data are likely to be unknown to the learning process.



In this study, we introduce the concept of semi-supervised continual learning (SSCL), which involves the use of both labeled and unlabeled data to train CL models. If the input is unlabeled data, the model update shown in \eqref{eqn:org_update} cannot be performed. This is because the underlying labels associated with the unlabeled data are unknown to the learning process. We assume that the set of classes associated with all labeled data $X$ is a subset of the set of classes associated with all unlabeled data $\tilde{X}$, denoted as $\Omega(X)\subset\Omega(\tilde{X})$.
When training with unlabeled data, there is no label available to feed into the loss function, which makes it impossible to compute the update shown in \eqref{eqn:org_update}. Thus, it is crucial to use unlabeled samples to predict pseudo gradients, represented as $\frac{\partial g|{\tilde{x}_{i}}}{\partial \theta}$, which can then be used to update the model through back-propagation.



\section{Methodology}
\label{sec:method}

In this section, we introduce how to train a gradient learner in a CL framework, and how to use the resulting gradient learner to predict gradients of unlabeled data.
We also discuss the sampling policy for unlabeled data and the geometric interpretation of the proposed gradient prediction.
\figref{fig:framework} shows an overview of the proposed SSCL method, which includes \textit{gradient learning} and \textit{gradient prediction} process.

\subsection{Gradient Learning}

In a continual learning framework, a model is designed to learn the mapping from raw data to the logits that minimize the pre-defined continual loss.
During the training process, at the $i$-th training step or episode, the generated logits $z_{i}$ w.r.t. the input $x_{i}$ is passed to the continual loss $\ell$. With the corresponding $y_{i}$, $\ell(z_{i}, y_{i})$ is computed to yield the gradient $\frac{\partial \ell}{\partial z_{i}}$.
We aim to compute the pseudo gradient $\bar{g}$ and use it to back-propagate the error and update the parameters $\theta$ by the chain rule. They can be mathematically summarized as
\begin{align}
    \text{Forward: }\;\; z_{i} &= f_{\theta}(x_{i},t_{i}), \\
    \text{Backward: } \frac{\partial \ell}{\partial \theta} &= \frac{\partial \ell}{\partial z_{i}} \frac{\partial z_{i}}{\partial \theta}, \ 
      \theta \leftarrow \theta - \eta \frac{\partial \ell_{fit}}{\partial \bar{g}|_{\tilde{x}_{i}}} \frac{\partial \bar{g}_{i}}{\partial \omega}.
  \label{eqn:fwd_bwd}
\end{align}

When the learning process is fed with unlabeled data $\tilde{x}_{i}$, it is desirable to have the logits and corresponding gradient so that $\tilde{x}_{i}$ can straightforwardly fit into the SCL framework.
Therefore, we propose a gradient learner $h$ that aims to learn the mapping from the logits $z_{i}$ to the gradients $\frac{\partial \ell}{\partial z_{i}}$, that is,
\begin{align}
    g_{i} = h(z_{i};\omega),
    \label{eqn:grad_learner}
\end{align}
where $\omega$ is the parameters of $h$ and $g_{i}$ is the predicted gradient that is expected to work as $\frac{\partial \ell}{\partial z_{i}}$ for back-propagation. 

To guarantee that the predicted gradients can mimic the gradients' efficacy in the learning process, we formulate the fitness of the predicted gradients w.r.t. the logits as a learning problem.
We define the fitness loss function to quantify the effect of the predicted gradients by fitting them back in the loss, \ie
\begin{align}
    \ell_{fit}(z_{i}, g_{i}, y_{i}) = \ell(z_{i}-\eta g_{i}, y_{i}).
    \label{eqn:fit_loss}
\end{align}
By observing triplet $(z_{i}, g_{i}, y_{i})$ at each training step, the minimization of $\ell_{fit}$ will iteratively update the proposed gradient learner $h(\cdot;\omega)$ through back-propagation.
As depicted in \eqref{eqn:fit_loss}, the predicted gradients aim to minimize the fitness loss, rather than mimicking the vanilla gradients $\frac{\partial \ell}{\partial z}$ in terms of direction and magnitude.

However, the gradients are sensitive in the learning process and a small change in gradients could lead to a divergence of training.
To obtain robust predicted gradients, instead of directly using the output of $h(\cdot;\omega)$ in the fitness loss (\ref{eqn:fit_loss}), we reference the magnitude $\tau_{i}$ of the vanilla gradient. With $\tau_{i}$, the predicted gradient can be accordingly normalized, \ie
\begin{align}
    \bar{g}_{i} = \alpha \tau_{i} g_{i}/\|g_{i}\|, \ \ \ \tau_{i} = \| \frac{\partial \ell}{\partial z_{i}} \|,
    \label{eqn:nml}
\end{align}
where $\alpha \in [0,1]$ is a hyperparameter that controls the proportion of the magnitude of the predicted gradient to $\tau_{i}$ and $z_{i}$ is generated by $(x_{i}, t_{i}, y_{i})$.
On the other hand, the output of the proposed gradient learner is a gradient that is subtle and crucial to the learning process. To properly update the proposed gradient learner, we apply a simple yet practically useful version of the loss scale technique \cite{Chen_ICML_2018,Lin_ICCV_2017,Zhao_arXiv_2019} to the fitness function. Specifically, the left hand side in \eqref{eqn:fit_loss} is multiplied with a pre-defined coefficient $\lambda$.
Finally, the fitness loss is computed with more robust $\bar{g}_{i}$, that is
\begin{align}
    \ell_{fit}(z_{i}, \bar{g}_{i}, y_{i}) = \lambda \ell(z_{i}-\eta \bar{g}, y_{i}).
    \label{eqn:robust_fit_loss}
\end{align}
Once the fitness loss is set, triplet $(z, \bar{g}, y)$ at each training step suffices to fit into the model learning. 
Specifically, the proposed gradient learner would be updated with $\frac{\partial \ell_{fit}}{\partial \bar{g}_{i}}$, \ie 
\begin{align}
    \omega \leftarrow \omega - \hat{\eta} \frac{\partial \ell_{fit}}{\partial \bar{g}_{i}} \frac{\partial \bar{g}_{i}}{\partial \omega}.
    \label{eqn:update_learner}
\end{align}
where $\hat{\eta}$ is the learning rate for updating $\omega$.
The model learning formed by the fitness loss (\ref{eqn:fit_loss}) and the update function (\ref{eqn:update_learner}) is generic and any gradient-based methods, \eg~multilayer perceptron (MLP)~\cite{Hastie_Springer_2009}, deep networks~\cite{He_CVPR_2016,Krizhevsky_NIPS_2012,Tan_ICML_2019}, or transformer~\cite{Vaswani_NIPS_2017}, can be used.
Without loss of generality, we use the baseline gradient-based method, \ie~MLP, in this work.

The process of learning to predict pseudo gradients is described in lines 6--10 in \algref{alg:gl}.

\begin{algorithm}[!t]
	\caption{Gradient Learning \& Prediction}\label{alg:gl} 
	\begin{algorithmic}[1]
		\State \textbf{Input}: $(x_{i}, t_{i}, y_{i})\in D_{tr}$, $\tilde{x}_{i}$, $\theta$, $\omega$, $\alpha$, $\lambda$, $\eta$, $\hat{\eta}$
		\State $z_{i}=f(x_{i},t_{i}; \theta)$
		\State $\ell_{i}=\ell(z_{i},y_{i})$
		\State Compute the gradient w.r.t. $z_{i}$, \ie $\frac{\partial \ell_{i}}{\partial z_{i}}$
		\State Update the model $\theta \leftarrow \theta - \eta \frac{\partial \ell_{i}}{\partial z_{i}} \frac{\partial z_{i}}{\partial \theta_{i}}$
		\State $g_{i} = h(z_{i};\omega)$
		\State $\bar{g}_{i} = \alpha \tau_{i} g_{i}/\|g_{i}\|, \ \ \ \tau_{i} = \| \frac{\partial \ell}{\partial z_{i}}, \|$
		\State $\ell_{fit} = \lambda \ell(z_{i}-\eta \bar{g}, y_{i})$
		\State Compute the gradient w.r.t $\bar{g}_{i}$, \ie $\frac{\partial \ell_{fit}}{\partial \bar{g}_{i}}$
		\State Update the gradient learner $\omega \leftarrow \omega - \hat{\eta} \frac{\partial \ell_{fit}}{\partial \bar{g}_{i}} \frac{\partial \bar{g}_{i}}{\partial \omega}$
		\If {$\tilde{x}_{i}$ is not equal to $\varnothing$}
		\State $z|_{\tilde{x}_{i}}=f(\tilde{x}_{i},t_{i}; \theta)$,  $\ g|_{\tilde{x}_{i}} = h(z|_{\tilde{x}_{i}};\omega)$
		\State $\bar{g}|_{\tilde{x}_{i}} = \alpha \tau_{i} g|_{\tilde{x}_{i}}/\|g|_{\tilde{x}_{i}}\|, \ \ \ \tau_{i} = \| \frac{\partial \ell}{\partial z_{i}} \|$
		\State $\theta \leftarrow \theta - \eta \bar{g}|_{\tilde{x}_{i}} \frac{\partial \bar{g}|_{\tilde{x}_{i}}}{\partial \theta}$
		\EndIf
	\end{algorithmic} 
\end{algorithm}

\begin{table}[!t]
	\centering
	\caption{\label{tbl:notation}
		Symbols and notations used in \algref{alg:gl}.
	}
	\vspace{-2ex}
	\adjustbox{width=1.0\columnwidth}{
	\begin{tabular}{L{6ex} L{45ex}}
		\toprule
		Symbol & Definition \\
		\cmidrule(lr){1-1} \cmidrule(lr){2-2} 
            $x_{i}$ & the $i$-th training sample \\
            $y_{i}$ & the label of the $i$-th training sample \\
            $t_{i}$ & the task of the $i$-th training sample \\
            $\tilde{x}_{i}$ & the $i$-th unlabeled sample \\
            $\ell$ & the loss function \\
            $z_{i}$ & the logit of the $i$-th training sample \\
            $\theta$ & the parameters of the classification model $f$ \\
            $\omega$ & the parameters of the gradient learner $h$ \\
            $\eta$ & the learning rate for updating $\theta$ \\
            $\hat{\eta}$ & the learning rate for updating $\omega$ \\
            $\alpha$ & the hyperparameter controlling the proportion of the magnitude \\
            $\lambda$ & the coefficient w.r.t. the fitness loss \\
		\bottomrule	
	\end{tabular}}
\end{table}

\subsection{Gradient Prediction}

To avail the additional unlabeled data in the learning process for better generalizability, the proposed gradient learner $h$ will predict gradients when the learning process is fed with unlabeled data $\tilde{x}$. 
Given $\tilde{x}_{i}$, the predicted gradient is computed in a similar way as \eqref{eqn:grad_learner} and (\ref{eqn:nml}) describe, but we use {\small $\tau_{i-1} = \| \frac{\partial \ell}{\partial z_{i-1}} \|$} (\ie~the last labeled sample prior to the $n$-th step), rather than $\tau_{i}$, as the label of $\tilde{x}_{i}$ is absent to produce $\| \frac{\partial \ell}{\partial z_{i}} \|$. Once the predicted gradient $\bar{g}|_{\tilde{x}_{i}}$ is computed, the model can be updated as
\begin{align}
    \theta \leftarrow \theta - \eta \bar{g}|_{\tilde{x}_{i}} \frac{\partial \bar{g}|_{\tilde{x}_{i}}}{\partial \theta}.
    \label{eqn:update_model}
\end{align}

To maintain the flexibility in leveraging external unlabeled data, we follow the basic idea of probability theory to presume that unlabeled data are sampled from a distribution. In contrast to the use of labeled data, where we assume all labeled data will be used during the training process, it is possible that no unlabeled data are sampled at some learning steps. In other words, the training process will revert to supervised learning if no unlabeled data is used. Mathematically, it can be formulated as
\begin{align}
    \tilde{x} = \left.
    \begin{cases}
        \tilde{x}_{i}\sim \mathcal{D}_{\tilde{x}}, \ \ &\text{if } q < p \\
        \varnothing, & \text{otherwise}\\
    \end{cases}
    \right.
    \label{eqn:sample}
\end{align}
where $q$ is a random variable following a distribution and $p$ is a pre-defined threshold. Without loss of generality, we assume the distribution is a standard uniform distribution $\mathcal{U}(0,1)$. When $p$ is set to 1, it indicates that the learning process will always draw several unlabeled data from a set $\tilde{X}$ of unlabeled data. When $p$ is set to 0, it indicates that the learning process will not draw any unlabeled data. In other words, $p$ manages the transition from SCL to SSCL.

The process of predicting gradients is described in lines 11--14 in \algref{alg:gl}, and the symbols used in the algorithm are depicted in \tabref{tbl:notation}.

\subsection{Connection to Pseudo Labeling}
\label{subsec:cnn_pl}

Note that we do not assume that the underlying classes that are associated with the unlabeled data are the same as or similar to the known classes. As a result, the distributions of the unlabeled samples could be very different from the labeled samples. Hence, directly predicting pseudo label for back-propagation may not be suitable in the SSCL setting.

\begin{figure}[!t]
	\centering
	\includegraphics[width=1\linewidth]{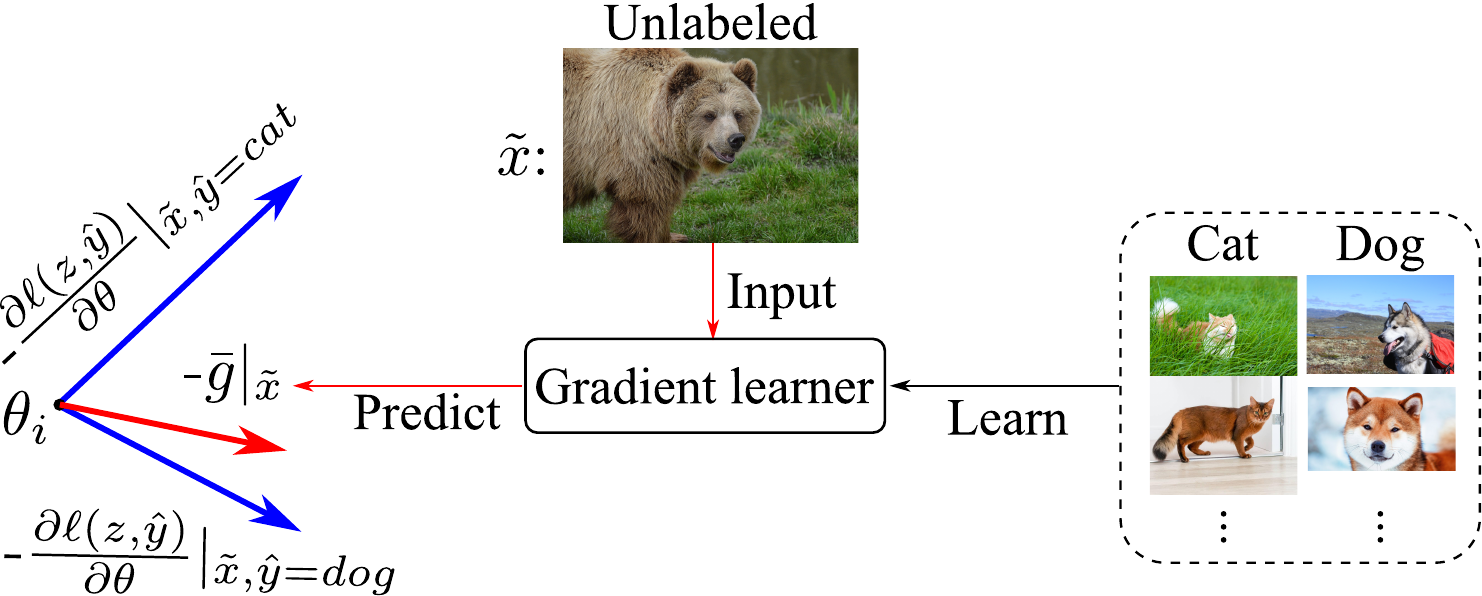}%
	\caption{\label{fig:comparison}
    	Comparison between the predicted gradient $\bar{g}$ and the gradients $\frac{\partial \ell(z,\hat{y})}{\partial \theta}$ generated with pseudo labels $\hat{y}$. Assume the proposed gradient learner is trained with the samples in categories cat and dog, given an unlabeled image $\tilde{x}$, the proposed gradient learner would take all learned class-specific knowledge (\ie w.r.t categories cat and dog) into account, instead of taking one category (\ie cat or dog) into account in pseudo labeling methods.
    	}
\end{figure}

When labels are unavailable, a common practice to utilize unlabeled samples is by training a teacher model with labeled samples and then predicting pseudo labels on unlabeled samples \cite{Li_ICCV_2017,Lee_2019_ICCV,Xie_CVPR_2020,Pham_arXiv_2020}.
Pseudo labeling \cite{Lee_2019_ICCV,Xie_CVPR_2020,Pham_arXiv_2020} is viewed as a teacher-student learning framework, \ie
\begin{align}
\minimize_{\theta'}\ \  &\ell(f_{\theta'}^{tch}(x_{i}, t_{i}), y_{i}) \label{eqn:teacher} \\
\hat{y} = \argmax_{j} &[ f^{tch}_{\theta'}(\tilde{x}, t_{i}) ]_{j} \label{eqn:teacher_pred} \\
\minimize_{\theta}\ \  &\ell(f_{\theta}^{stn}(x_{i}, t_{i}), \hat{y}) \label{eqn:student}
\end{align}
where \textit{tch} (resp. \textit{stn}) stands for teacher (resp. student), $\theta'$ (resp. $\theta$) are the weights of the teacher (resp. student), $((x_{i}, t_{i}), y_{i})$ is a labeled sample, and $\tilde{x}$ is a unlabeled sample. 
In short, the teacher would be trained with labeled samples by \eqref{eqn:teacher}.
When it comes across unlabeled sample $\tilde{x}$, the teacher first predicts an one-hot pseudo label $\hat{y}$ by \eqref{eqn:teacher_pred} and then $\hat{y}$ is viewed as the label for training the student by \eqref{eqn:student}.
A common alternative to one-hot pseudo labels in \eqref{eqn:teacher_pred} is the probabilities w.r.t. each class, which is used in leveraging unlabeled data in the class-incremental learning \cite{Lee_2019_ICCV}. 
To generate probabilistic labels, the softmax function is usually applied.
We denote the one-hot pseudo labeling method and the probabilistic pseudo labeling method as \textit{1-PL} and \textit{P-PL} for simplicity.

The difference between gradient prediction and pseudo labeling is shown in \figref{fig:comparison}. As teacher models have a chance of generating incorrect labels, the resulting gradients would vary with different pseudo labeling. Instead, the proposed gradient learner is trained with labeled samples so the predicted gradients are generated with implicit knowledge that maps visual appearance to various visual concepts, rather than one. For example, given training samples of cat and dog, when the proposed gradient learner observes a fox image to predict the pseudo gradient, the pseudo gradient would be aligned with the learned knowledge of both cat and dog, instead of only cat or only dog. Therefore, the pseudo gradients generated by the proposed gradient learner have better generalizability than the ones generated by pseudo labeling methods.
Last but not least, as indicated in \eqref{eqn:fit_loss}, the predicted gradients are generated to minimize the fitness loss, while the gradients generated by pseudo labeling methods aim to reproduce the gradients generated with ground-truth labels. Ideally, if the pseudo labels are identical to the ground-truth labels, the  gradients generated with pseudo labels would be identical to the gradients generated with ground-truth labels. However, this rarely happens in practice as unlabeled data have no labels or the underlying labels are unknown.

On the other hand, each task in SSCL has a limited number of labeled samples and the visual concepts of any two tasks are different. With limited labeled samples, it is difficult to predict correct pseudo labels. Thus, predicting pseudo gradients is more straightforward and effective in this case.

Furthermore, pseudo labeling methods have many more parameters than the proposed gradient learner. 
Although the outputs of the teacher model and the proposed gradient learner are supposed to be of the same dimension, the inputs are different. 
The former takes images as input whereas the latter takes CL models' output as input.
Thus, the teacher models usually have the same as or more parameters ($>1M$) than the student models \cite{Lee_2019_ICCV,Chen_NIPS_2020,Xie_CVPR_2020}, whereas the proposed gradient learner is a small MLP with fewer parameters ($<10K$)

\subsection{Geometric Interpretation}

\begin{figure}[!t]
	\centering
	\includegraphics[width=0.95\linewidth]{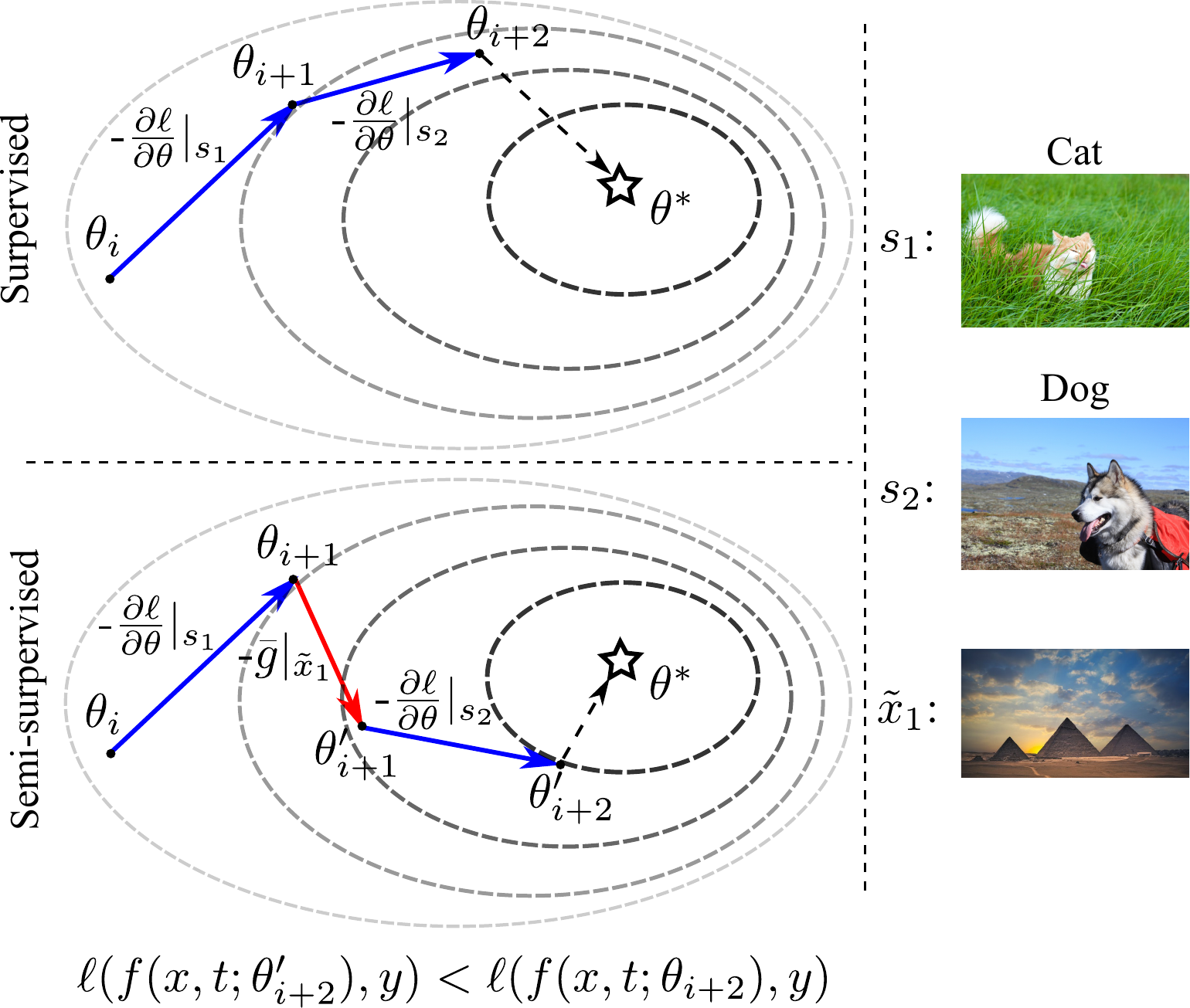}%
	\caption{\label{fig:geom}
    	Geometric interpretation of supervised learning (top) and semi-supervised learning (bottom). Through leveraging the semantics of unlabeled images, the generalizability of models is expected to be improved. 
    	Experimental results in Table~\ref{tbl:conn_rota}--\ref{tbl:acl_miniImageNet} validate this finding.
    	}
\end{figure}

\figref{fig:geom} shows the geometric interpretation of gradient prediction by comparing SSL (bottom) with supervised learning (top). 
In this illustrative example, given two labeled images, $s_{1}$ and $s_{2}$, and one unlabeled image, $\tilde{x}_{1}$, the predicted gradient $-\bar{g}|_{\tilde{x}_{1}}$ helps boost the convergence, \ie~$\theta'_{i+2}$ is closer to the underlying local minimum $\theta^{*}$ than $\theta_{i+2}$.
This also impacts on the generalizability. Given an unseen labeled triplet $(x,t,y)$, we have inequality {\small $\ell(f(x;\theta'_{i+2}), t, y) < \ell(f(x;\theta_{i+2}), t, y)$}.
This implies that the CL model with pseudo gradients is likely to be closer to a local minimum than the one that is not using it.

\subsection{Trade-off: Overwhelming vs. Generalizing}
\label{subsec:tradeoff}

It is desirable to use as many unlabeled data as possible, as long as the data improve the generalizability of the CL models. Unfortunately, this goal is difficult to achieve. The reasons are two-fold. 
Firstly, as shown in \figref{fig:diff}, since the underlying classes of the unlabeled images are unknown, the distributions of unlabeled data could be considerably different from the ones of the labeled data.
Secondly, gradient learning and prediction are challenging as it is a regression task in a high-dimensional space and the values of gradients are usually small but influential.
Last but not least, in contrast to classification task, where the labels are one-hot vectors that are in $[0,1]$, the ranges of vanilla gradients are determined by the labeled data and lie in $(-\infty,+\infty)$.
Therefore, gradient learning task is by nature very challenging.

As a result, when more unlabeled data are used in the learning process, it is more prone to accumulate prediction errors that harm the training quality. 
Specifically, predictive errors in the back-propagation could overwhelm the knowledge learned from the given labeled data. 
Therefore, achieving a good trade-off between overwhelming and generalizing is important in the SSCL problem. 
In this work, we use a probabilistic threshold $p$ to implement this trade-off.

\section{Experiment}
\label{sec:exp}

\subsection{Experimental Set-up}

We follow the experimental protocols used in GEM~\cite{Lopez_NIPS_2017} and ACL~\cite{Ebrahimi_ECCV_2020}, which are cast into the category of task-incremental learning.
In the training scheme of \cite{Lopez_NIPS_2017}, the models will observe training samples and no training samples will be observed for a second time. 
The training and test samples are randomly assigned to $n$ tasks according to classes and each task has training and test samples with different classes from the other tasks. 
Similarly, \cite{Ebrahimi_ECCV_2020} randomly assigns the samples into $n$ tasks, but in each task, there are multiple epochs that repeat the stochastic process over the training samples as the conventional supervised learning. 
After the training on every task is done, the trained models would be evaluated with all the samples of all tasks, including the tasks that the CL process has gone through and the tasks which have not been executed yet.
Furthermore, to understand the generalization ability of the proposed method in SSL task, we follow the experimental protocols used in \cite{Zhang_AISTATS_2021} to evaluate the proposed method.

\subsection{Datasets}

In the GEM training scheme, we use the following datasets.
\textbf{MNIST permutation (MNIST-P)}~\cite{Kirkpatrick_PNAS_2017} is a variant of MNIST~\cite{Lecun_IEEE_1998}, which consists of 70k images of size 28$\times$28. 
Each image is transformed by a fixed permutation of pixels.
\textbf{MNIST rotation (MNIST-R)}~\cite{Lopez_NIPS_2017} is similar to MNIST-P, but each image is rotated by a fixed angle between 0 and 180 degrees.
\textbf{Incremental CIFAR-100 (iCIFAR-100)}~\cite{Rebuffi_CVPR_2017} is a variant of the CIFAR-100~\cite{Krizhevsky_TR_2009}, which consists of 60k images of size 32$\times$32 that are split into multiple subsets by the classes.

In the ACL training scheme, we use \textbf{CIFAR-100}~\cite{Ebrahimi_ECCV_2020} and \textbf{miniImageNet}~\cite{Vinyals_NIPS_2016}.
CIFAR-100 is also split into multiple subsets like iCIFAR-100. 
Instead of being used once, the images in each task are repeatedly used in every epoch.
miniImageNet is a variant of ImageNet~\cite{Deng_CVPR_2009}, which consists of 60k images of size 84$\times$84 with 100 classes.

Following GEM and ACL, all training samples are split into 20 tasks. 
Briefly, each task on iCIFAR-100, CIFAR-100, and miniImageNet has 5 classes. 
For MNIST-P and MNIST-R, each task has 10 classes and is performed with different permutation or rotation from the other tasks.

For the experiments on MNIST-R, MNIST-P, iCIFAR-100, and CIFAR-100, we use Tiny ImageNet as unlabeled dataset. 
For the experiments on miniImageNet, we use the unlabeled images from MS COCO~\cite{Lin_ECCV_2014}. 
Both unlabeled dataset are widely-used large-scale real-world datasets. 
Hence, the unlabeled pool are representative and general for various CL tasks.

In the semi-supervised training scheme, we follow the same experimental protocols used in \cite{Zhang_AISTATS_2021} to evaluate the proposed method on SVHN \cite{Netzer_Report_2011}, CIFAR-10, and CIFAR-100 \cite{Krizhevsky_TR_2009}. The numbers of labeled data are 1k, 4k, and  10k for SVHN, CIFAR-10, and CIFAR-100, respectively.

\subsection{Metrics \& Methods}
\label{subsec:metrics}

To comprehensively validate the performance of the proposed method, we conduct experiments based on the training schemes of GEM and ACL. 
DCL~\cite{Luo_TPAMI_2019} achieves state-of-the-art performance on MNIST-P, MNIST-R and iCIFAR-100, and is considered as another baseline in the GEM training scheme.

CL has three key metrics, namely average accuracy (ACC), backward transfer (BWT), and forward transfer (FWT) \cite{Lopez_NIPS_2017}, \ie
\begin{align}
	ACC &= \frac{1}{T} \sum_{i=1}^{T} R_{T,i} \\
	BWT &= \frac{1}{T-1} \sum_{i=1}^{T-1} R_{T,i}-R_{i,i} \label{eq:BWT} \\
	FWT &= \frac{1}{T-1} \sum_{i=2}^{T} R_{i-1,i}-\bar{b}_{i}
	\label{eq:FWT}
\end{align}
where $R_{i,j}$ is the test classification accuracy that is evaluated on the test set of the $j$-th task when training on the $i$-th task, $T$ is the number of tasks, and $\bar{b}_{i}$ is the test classification accuracy at random initialization at the $i$-th task.
Average accuracy indicates the predictive ability of the trained models on all tasks. BWT measures the effect of how learning a task $t$ influences the performance on previous tasks $k < t$. A large negative score is referred as catastrophic forgetting while a positive score implies that learning new tasks generalizes to previous tasks. Correspondingly, FWT measures the effect of how learning a task $t$ influences the performance on future tasks $k > t$. A positive score implies that learning a task generalizes to future tasks, which is similar to zero-shot learning. In the ACL training scheme, we use the same metrics, \ie average accuracy and BWT, as \cite{Ebrahimi_ECCV_2020}. We denote a baseline as \textit{backbone} (if any) \textit{continual algorithm}, \eg ResNet GEM. Similarly, we denote the proposed method as \textit{backbone} (if any) \textit{continual algorithm} + proposed, which indicates the proposed method is used to leverage the information from unlabeled images.
Also, following \cite{Lee_2019_ICCV} and \cite{Zhang_AISTATS_2021}, we report the performance of 1-PL, P-PL, and MG for the purpose of comparison in the CL setting. 
Specifically, the teacher model takes images as input to predict pseudo labels when the learning process encounters unlabeled images. The teacher model is composed by the same backbone of the CL model and a linear transformation layer that generate pseudo labels in each task. 
In other words, 1-PL and P-PL have many more parameters than the baseline and the proposed method.

\subsection{Hyperparameters \& Implementation Details}

We use the same training hyperparameters in GEM~\cite{Lopez_NIPS_2017}, DCL~\cite{Luo_TPAMI_2019}, and ACL~\cite{Ebrahimi_ECCV_2020}. More details can be found in these works or in our code repository. Here, we focus on the hyperparameters that are related to the proposed method. 
There are five hyperparameters, namely threshold $p$, magnitude ratio $\alpha$,  loss scale $\lambda$, network architecture $h_{\omega}$, and batch size of unlabeled images.
The hyperparameters w.r.t. the proposed method used in the SSCL and SSL setting are reported in \tabref{tbl:hyper}. 
In particular, we follow \cite{Zhang_AISTATS_2021} to use the unlabeled data with $p=1.0$ in SSL.


\begin{table}[!t]
	\centering
	\caption{\label{tbl:hyper}
		Hyperparameters w.r.t. the proposed method. 
		BS denotes batch size of unlabeled images.
	}
 	\vspace{-2ex}
	\adjustbox{width=1.0\columnwidth}{
		\begin{tabular}{c l c c c c c c c}
			\toprule
			Setting & Dataset & Method & Backbone & BS & $p$ & $\alpha$ & $\lambda$ & $h_{\omega}$ \\ \cmidrule{1-1} \cmidrule{2-9}  
			\multirow{10}{*}{SSCL} & MNIST-R & GEM  & MLP & 4 & 0.15 & 0.001 & 0.30 & (64,16) \\
			& MNIST-R & DCL  & MLP & 4 & 0.15 & 0.001 & 0.30 & (64,16) \\ \cmidrule{2-9}
			& MNIST-P & GEM & MLP & 4 & 0.15 & 0.001 & 0.50 & (64,16) \\
			& MNIST-P & DCL & MLP & 4 & 0.15 & 0.001 & 0.49 & (64,16) \\ \cmidrule{2-9}
			& iCIFAR-100 & GEM & ResNet-18 & 4 & 0.30 & 0.005 & 2.00 & (128,32) \\
			& iCIFAR-100 & DCL & ResNet-18 & 4 & 0.30 & 0.005 & 2.50 & (128,32) \\
			& iCIFAR-100 & GEM & EffNet-B1 & 4 & 0.20 & 0.005 & 2.00 & (128,32) \\
			& iCIFAR-100 & DCL & EffNet-B1 & 4 & 0.35 & 0.005 & 2.00 & (128,32) \\ \cmidrule{2-9}
			& CIFAR-100 & ACL  & AlexNet & 64 & 0.30 & 0.001 & 0.20 & (128,32) \\ \cmidrule{2-9}
			& miniImageNet & ACL & AlexNet & 64 & 0.35 & 0.001 & 0.15 & (128,32) \\
			\cmidrule{1-9}
            \multirow{3}{*}{SSL} & SVHN & -  & Conv-Large \cite{Tarvainen_NeurIPS_2017} & - & - & 0.001 & 2.00 & (128,32) \\
            & CIFAR-10 & - & Conv-Large & - & - & 0.001 & 1.00 & (128,32) \\
		& CIFAR-100 & - & Conv-Large & - & - & 0.001 & 1.00 & (128,32) \\
            \bottomrule
	\end{tabular}}
\end{table}



Specifically, this work follows the same experimental protocol used in \cite{Lopez_NIPS_2017,Luo_TPAMI_2019} to evaluate the proposed method on MNIST-R, MNIST-P, and iCIFAR-100, while it follows the same experimental protocol used in \cite{Ebrahimi_ECCV_2020} to evaluate the proposed method on CIFAR-100 and miniImageNet.
All hyperparameters that are used with the baselines are used with the proposed method as well.

Without loss of generality, we use MLP as the gradient learner {\small $h(\cdot;\omega)$} ({\small $h_{\omega}$} for short). 
Assume the gradient is in $\mathbf{R}^{m}$, we denote (\textit{dimension of the 1st layer output}, \textit{dimension of the 2nd layer output}, \ldots, \textit{dimension of the penultimate layer output}) for simplicity. 
For instance, given $m=5$, architecture $(64, 16)$ indicates the MLP consists of three layers, the first one is a linear operation with a coefficient matrix of size $5\times 64$, the second one is with a coefficient matrix of $64\times 16$, and the last one is with a coefficient matrix of size $16\times 5$.

Similar to other supervised learning methods, few learning steps may not be adequate to train a good gradient learner. 
Hence, the gradient learner is trained from the very beginning, but the predicted gradients are used after 50 learning steps in the GEM and DCL training scheme, and after 5 learning steps in the ACL training scheme.

Note that restricted to the shared and private module design in ACL \cite{Ebrahimi_ECCV_2020}, which requires a fixed dimension of the input features, the batch size of unlabeled images has to be the same as the batch size of training samples, that is, 64.


\subsection{Generalization Performance} \label{subsec:perf}

\begin{table}[!t]
	\centering
	\caption{\label{tbl:conn_rota}
		Performance on MNIST-R. 
		All methods use MLP as the backbone network~\cite{Lopez_NIPS_2017,Luo_TPAMI_2019}. 
		The proposed gradient learner has 1824 parameters. Accuracy is in (\%). The top performance is highlighted in bold. MG and PG stand for meta gradient \cite{Zhang_AISTATS_2021} and predicted gradient (proposed), respectively. FT and COS indicate the fitness loss and the cosine similarity loss, respectively. 
	}
	\vspace{-2ex}
	\adjustbox{width=1.0\columnwidth}{
	\begin{tabular}{L{28ex} C{9ex} C{9ex} C{9ex}}
		\toprule
    	Methods                          & Accuracy       & BWT & FWT \\
		\cmidrule(lr){1-1} \cmidrule(lr){2-2} \cmidrule(lr){3-3} \cmidrule(lr){4-4}
		EWC~\cite{Kirkpatrick_PNAS_2017} & 54.61          & -0.2087         & 0.5574 \\
		GEM~\cite{Lopez_NIPS_2017}       & 83.35          & -0.0047         & 0.6521 \\
		DCL~\cite{Luo_TPAMI_2019}        & 84.08          & 0.0094          & 0.6423 \\ \midrule
		GEM + 1-PL            			 & 74.58          & -0.0782         & 0.6319 \\
		GEM + P-PL					     & 79.39          & -0.0380         & 0.6453 \\
		GEM reproduced                   & 83.03          & -0.0061         & 0.6482 \\
		GEM + MG                   & 84.97 & 0.0051 & 0.6552 \\ 
		GEM + proposed                   & \textbf{86.54} & \textbf{0.0227} & 0.6537 \\ 
		\midrule
		DCL + 1-PL			             & 82.12          & 0.0022         & 0.6275 \\
		DCL + P-PL					     & 83.34          & 0.0033         & 0.6359 \\
		DCL reproduced                   & 84.88          & 0.0088          & 0.6526 \\
		DCL + MG                   &     85.74      &     0.0168      & 0.6518 \\
		DCL + proposed                   & 86.26          & 0.0106          & \textbf{0.6620} \\
		\bottomrule	
	\end{tabular}}
\end{table}


Tables \ref{tbl:conn_rota}--\ref{tbl:conn_cifar} report the performance of the proposed method with comparison to the compared baselines on MNIST-R, MNIST-P, and iCIFAR-100.
EWC \cite{Kirkpatrick_PNAS_2017}, iCARL \cite{Rebuffi_CVPR_2017}, MAS \cite{Aljundi_ECCV_2018}, A-GEM \cite{Chaudhry_ICLR_2019}, LUCIR \cite{Hou_CVPR_2019}, BiC \cite{Wu_CVPR_2019}, HAL \cite{Chaudhry_AAAI_2021}, DER \cite{Buzzega_NeurIPS_2020}, X-DER \cite{Boschini_arXiv_2022}, CFA \cite{Carvalho_ECML_2022}, \REVISION{MutexMatch \cite{Duan_TNNLS_2022}, Interpolation-Based Contrastive Learning (ICL) \cite{Yang_TNNLS_2022}, and the Glimpse Network \cite{Li_TNNLS_2023}} use the same ResNet backbone.
Compared to these baselines, the proposed method achieves higher average accuracy and BWT, \eg~ResNet GEM + proposed. This implies that the proposed method effectively utilizes the information of unlabeled images to improve the predictive ability, alleviates catastrophic forgetting, and enhances zero-shot learning ability.
Moreover, the proposed method consistently improves the average accuracy, BWT, and FWT of the baselines with the same backbone, \eg~ResNet GEM reproduced vs. ResNet GEM + proposed.

In the ACL setting (\ie~\tabref{tbl:acl_cifar100} and \ref{tbl:acl_miniImageNet}), the average accuracy and BWT of the baseline are improved by the proposed method. Moreover, the standard deviation w.r.t. the proposed method over 5 runs is smaller than the corresponding baseline. 
This implies the proposed method is more stable than the baseline.

\begin{table}[!t]
	\centering
	\caption{\label{tbl:conn_perm}
		Performance on MNIST-P. 
		All methods use MLP as the backbone network~\cite{Lopez_NIPS_2017,Luo_TPAMI_2019}. 
		The proposed gradient learner has 1824 parameters.
	}
	\vspace{-2ex}
	\adjustbox{width=1.0\columnwidth}{
	\begin{tabular}{L{28ex} C{9ex} C{9ex} C{9ex}}
		\toprule
		Methods & Accuracy & BWT & FWT \\
		\cmidrule(lr){1-1} \cmidrule(lr){2-2} \cmidrule(lr){3-3} \cmidrule(lr){4-4}
		EWC~\cite{Kirkpatrick_PNAS_2017} & 59.31 & -0.1960 & -0.0075 \\
		GEM~\cite{Lopez_NIPS_2017}       & 82.44 & 0.0224  & -0.0095 \\
		DCL~\cite{Luo_TPAMI_2019}        & 82.58 & 0.0402  & -0.0092 \\ \midrule
		GEM + 1-PL			             & 80.61 & 0.0327 & -0.0014 \\
		GEM + P-PL					     & 80.58 & 0.0224 & -0.0039 \\
		GEM reproduced                   & 82.35 & 0.0251  & -0.0101 \\
		GEM + MG                   & 82.30  & 0.0332   & -0.0170  \\
		GEM + proposed                   & 82.91 & 0.0316  & -0.0072 \\ 
		\midrule
		DCL + 1-PL			             & 81.57 & \textbf{0.0479} & \textbf{0.0002} \\
		DCL + P-PL					     & 80.95 & 0.0219 & -0.0083 \\
		DCL reproduced                   & 82.83 & 0.0279  & -0.0100 \\
		DCL + MG                   &  82.48 &  0.0423  & -0.0078  \\
		DCL + proposed                   & \textbf{82.97} & 0.0402  & -0.0038 \\
		\bottomrule	 
	\end{tabular}}
\end{table}
\begin{table}[!t]
	\centering
	\caption{\label{tbl:conn_cifar}
	    Performance on iCIFAR-100. \textit{ResNet} indicates ResNet-18. \textit{EffNet} stands for EfficientNet (B1) \cite{Tan_ICML_2019}. The proposed gradient learner has 4896 parameters.}
	\vspace{-2ex}
	\adjustbox{width=1.0\columnwidth}{
	\begin{tabular}{L{28ex} C{9ex} C{9ex} C{9ex}}
		\toprule
		Methods & Accuracy & BWT & FWT \\
		\cmidrule(lr){1-1} \cmidrule(lr){2-2} \cmidrule(lr){3-3} \cmidrule(lr){4-4}
		EWC \cite{Kirkpatrick_PNAS_2017} & 48.33 & -0.1050 &  \textbf{0.0216} \\
		iCARL \cite{Rebuffi_CVPR_2017} & 51.56 & -0.0848 &  0.0000 \\
            MAS \cite{Aljundi_ECCV_2018} & 49.45 & -0.0674 & 0.0157 \\
            A-GEM \cite{Chaudhry_ICLR_2019} & 67.14 & 0.0037 & 0.0087 \\
            LUCIR \cite{Hou_CVPR_2019} & 58.71 & 0.0177 & -0.0067 \\
            BiC \cite{Wu_CVPR_2019} & 60.92 & -0.0010 & -0.0023 \\
            HAL \cite{Chaudhry_AAAI_2021} & 63.85 & 0.0017 & 0.0088 \\
            DER \cite{Buzzega_NeurIPS_2020} & 65.72 & 0.0011 & 0.0053 \\
            X-DER \cite{Boschini_arXiv_2022} & 68.32 & 0.0223 & 0.0017 \\
            CFA \cite{Carvalho_ECML_2022} & 67.41 & 0.0124 & -0.0026 \\
            MutexMatch \cite{Duan_TNNLS_2022} & 68.09 & 0.0156 & 0.0021 \\
            ICL \cite{Yang_TNNLS_2022} & 67.23 & 0.0084 & -0.0012 \\
            Glimpse \cite{Li_TNNLS_2023} & 66.87 & 0.0169 & -0.0031 \\
		ResNet GEM \cite{Lopez_NIPS_2017} & 66.67 & 0.0001 &  0.0108 \\
		ResNet DCL \cite{Luo_TPAMI_2019} & 67.92 & 0.0063 & 0.0102 \\
		EffNet GEM \cite{Luo_TPAMI_2019} & 80.80 & 0.0318 & -0.0050 \\
		EffNet DCL \cite{Luo_TPAMI_2019} & 81.55 & 0.0383 & -0.0048 \\ \midrule
		ResNet GEM + 1-PL & 65.44 & 0.0861 & -0.0030  \\
		ResNet GEM + P-PL & 65.55 & 0.0511 & -0.0033  \\
		ResNet GEM reproduced & 66.92 & 0.0132 & -0.0048  \\
		ResNet GEM + MG & 67.24  & 0.0614  &  -0.0001  \\
		ResNet GEM + proposed & 68.74 & 0.0619 & 0.0055  \\
		\midrule
		ResNet DCL + 1-PL & 66.43 & 0.0765 & 0.0051  \\
		ResNet DCL + P-PL & 67.78 & 0.0704 & 0.0078  \\
		ResNet DCL reproduced & 67.55 & 0.0048 & -0.0117  \\
		ResNet DCL + MG & 66.07  & 0.0524  &  0.0184  \\
        ResNet DCL + proposed & 68.53 & 0.0574 & -0.0038  \\
		\midrule
		EffNet GEM + 1-PL & 78.33 & 0.0855 & -0.0106  \\
		EffNet GEM + P-PL & 77.46 & 0.0535 & 0.0077  \\
		EffNet GEM reproduced & 81.44 & 0.0128 & 0.0105  \\
		EffNet GEM + MG &  83.95 & 0.0294  &  -0.0256  \\
		EffNet GEM + proposed & 85.51 & 0.0219 & 0.0148 \\
		\midrule
		EffNet DCL + 1-PL & 77.12 & \textbf{0.0862} & -0.0160  \\
		EffNet DCL + P-PL & 76.82 & 0.0821 & 0.0097  \\
		EffNet DCL reproduced & 83.47 & 0.0266 & -0.0185  \\
		EffNet DCL + MG & 85.06  & 0.0488  &  -0.0043  \\
		EffNet DCL + proposed & \textbf{85.70} & 0.0378 & 0.0017  \\
		\bottomrule	
	\end{tabular}}
\end{table}

\begin{table}[!t]
	\centering
	\caption{\label{tbl:acl_cifar100}
		Performance on CIFAR-100 in adversarial continual learning setting. The training process is repeated 5 times, and the average accuracy and standard deviation are reported~\cite{Ebrahimi_ECCV_2020}. ACL uses AlexNet~\cite{Krizhevsky_NIPS_2012} as backbone. The proposed gradient learner has 1427 parameters.
	}
	\vspace{-2ex}
	\adjustbox{width=1.0\columnwidth}{
	\begin{tabular}{L{22ex} C{15ex} C{15ex}}
		\toprule
		Methods & Accuracy & BWT \\
		\cmidrule(lr){1-1} \cmidrule(lr){2-2} \cmidrule(lr){3-3}
		A-GEM~\cite{Chaudhry_ICLR_2018}    & 54.38$\pm$3.84 & -0.2199$\pm$0.0405 \\
		ER-RES~\cite{Chaudhry_arXiv_2019} & 66.78$\pm$0.48 & -0.1501$\pm$0.0111 \\
		PNN~\cite{Rusu_arXiv_2016}         & 75.25$\pm$0.04 & 0  \\
		HAT~\cite{Serra_ICML_2018}         & 76.96$\pm$1.23 & 0.0001$\pm$0.0002  \\
		ACL~\cite{Ebrahimi_ECCV_2020}      & 78.08$\pm$1.25 & 0$\pm$0.0001 \\ \midrule
		ACL reproduced                     & 78.17$\pm$1.32 & \textbf{0.01}$\pm$0.0168   \\
		ACL + proposed                     & \textbf{78.46}$\pm$1.05 & \textbf{0.01}$\pm$0.0123  \\
		\bottomrule	
	\end{tabular}}
\end{table}


On the other hand, 1-PL and P-PL yield lower accuracies than the proposed method. This is because the pseudo labels are likely to be incorrect as the training samples are not adequate and the visual concepts vary from task to task (the analysis of pseudo labeling is provided in \secref{sec:analysis}). Incorrect pseudo labels lead to the gradients that guide the learning process in unpredictable directions.
Note that the BWTs of 1-PL and P-PL are higher than the others. This results from lower accuracy. As indicated in the definition of BWT (\eqref{eq:BWT}), when the test classification accuracy $R_{i,i}$ on the $i$-th task with the model trained in the $i$-th task is low, it will lead to high BWT. In other words, when overall ACC is high, BWT tends to be relatively low. 
Similarly, FWT tends to be high (\ie 0.0216) when the corresponding accuracies over tasks are low (\ie 48.33\%).

Since the proposed method is generic, we also evaluate it in the SSL setting \cite{Zhang_AISTATS_2021}.
The meta-objective defined in \cite{Zhang_AISTATS_2021} is used as the fitness loss to learn to predict pseudo gradients for unlabeled images.
The experimental results on SVHN \cite{Netzer_Report_2011}, CIFAR-10, and CIFAR-100 are reported in \tabref{tbl:semi_exp}.
The proposed method can improve the performance of the SSL task. This implies that the proposed method generally work with the unlabeled data with the pseudo labels that share the same or similar distributions as the labeled data.


\section{Analysis} \label{sec:analysis}

This section provides a series of experiments to analyze the proposed method. 
All analyses are based on iCIFAR-100.

\begin{table}[!t]
	\centering
	\caption{\label{tbl:acl_miniImageNet}
		Performance on miniImageNet in adversarial continual learning setting. The training process is repeated 5 times, and the average accuracy and standard deviation are reported~\cite{Ebrahimi_ECCV_2020}. ACL uses AlexNet~\cite{Krizhevsky_NIPS_2012} as backbone.
        The proposed gradient learner has 1427 parameters.
	}
	\vspace{-2ex}
	\adjustbox{width=1.0\columnwidth}{
	\begin{tabular}{L{22ex} C{15ex} C{15ex}}
		\toprule
		Methods & Accuracy & BWT \\
		\cmidrule(lr){1-1} \cmidrule(lr){2-2} \cmidrule(lr){3-3}
		A-GEM \cite{Chaudhry_ICLR_2018} & 52.43$\pm$3.10 & -0.1523$\pm$0.0145 \\
		ER-RES \cite{Chaudhry_arXiv_2019} & 57.32$\pm$2.56 & -0.1134$\pm$0.0232 \\
		PNN \cite{Rusu_arXiv_2016} & 58.96$\pm$3.50 & 0  \\
		HAT \cite{Serra_ICML_2018} & 59.45$\pm$0.05 & -0.0004$\pm$0.0003  \\
		ACL \cite{Ebrahimi_ECCV_2020} & 62.07$\pm$0.51 & 0$\pm$0 \\ \midrule
		ACL reproduced & 62.69$\pm$1.01 & \textbf{0}$\pm$0.0042   \\
		ACL + proposed & \textbf{63.88}$\pm$0.39 & \textbf{0}$\pm$0.0000 \\
		\bottomrule	
	\end{tabular}}
\end{table}

\begin{table}[!t]
	\centering
	\caption{\label{tbl:semi_exp}
	    Semi-supervised classification error rates (\%) of the Conv-Large \cite{Tarvainen_NeurIPS_2017} architecture on the SVHN, CIFAR-10, and CIFAR-100 datasets. The numbers of labeled data are 1k, 4k, and 10k for these three datasets, respectively. We follow the exact experimental protocol used in \cite{Zhang_AISTATS_2021} and use the official implementation code to conduct this experiment. The meta-objective defined in \cite{Zhang_AISTATS_2021} is used as the fitness loss to learn to predict pseudo gradients for unlabeled images.
	    }
	\vspace{-2ex}
	\adjustbox{width=1.0\columnwidth}{
	\begin{tabular}{L{18ex} C{8ex} C{10ex} C{11ex}}
		\toprule
		Method & SVHN & CIFAR-10 & CIFAR-100 \\
		\cmidrule(lr){1-1} \cmidrule(lr){2-2} \cmidrule(lr){3-3} \cmidrule(lr){4-4}
		Co-training \cite{Qiao_ECCV_2018} & 3.29 & 8.35 & 34.63  \\
		TNAR-VAE \cite{Yu_CVPR_2019}  & 3.74 & 8.85 & - \\
		ADA-Net \cite{Wang_ICCV_2019} & 4.62 & 10.30 & -  \\
		DualStudent \cite{Ke_ICCV_2019} & - & 8.89 & 32.77 \\
		MG \cite{Zhang_AISTATS_2021} & 3.15 & 7.78 & 30.74  \\
		 \midrule
		 MG reproduced & 3.53 & 7.82 & 30.74  \\
		 MG + proposed & 3.45 & 7.46 & 30.02  \\
		\bottomrule	
	\end{tabular}}
\end{table}

\subsection{Effects of Visual Diversity}

Here, we study the influence of the variance between training images and the unlabeled images on the model performance. In addition to the Tiny ImageNet from the previous section, we selected a variety of datasets, namely MS COCO~\cite{Lin_ECCV_2014}, CUB-200~\cite{Wah_report_2011}, FGVC-aircraft~\cite{Maji_report_2013}, and Stanford-cars~\cite{Krause_ICCVW_2013}, as the source of unlabeled images. The classes in these datasets overlap with the ones in CIFAR-100 to various degrees. The performance are reported in \tabref{tbl:diversity}. Overall, the proposed method shows improvement with all unlabeled images source. The images in Tiny ImageNet are similar to the ones in MS COCO, where both are natural images but having different image resolution. The resolution of images in Tiny ImageNet is closer to that in CIFAR-100 than MS COCO. Therefore, using Tiny ImageNet images leads to the most performance improvement. In contrast, the images in FGVC-aircraft are the most dissimilar to the ones in CIFAR-100 and the accuracy improvement is marginal. On the other hand, using CUB-200 leads to higher accuracy than using MS COCO. This is because CUB-200 shares similar visual concepts with CIFAR (\ie bird) and both the two datasets are object-centered, whereas the images of MS COCO contain multiple objects and are non-object-centered. 

\begin{table}[!t]
	\centering
	\caption{\label{tbl:diversity}
	    Effects of visual diversity of $\tilde{x}$ on the classification performance (\%) on iCIFAR-100 with ResNet GEM. 
	}
	\vspace{-2ex}
	\adjustbox{width=1.0\columnwidth}{
	\begin{tabular}{L{22ex} C{9ex} C{9ex} C{9ex}}
		\toprule
		Source & Accuracy & BWT & FWT \\
		\cmidrule(lr){1-1} \cmidrule(lr){2-2} \cmidrule(lr){3-3} \cmidrule(lr){4-4}
		$\tilde{x}=\varnothing$                & 66.92 & 0.0132 & -0.0048  \\ \midrule
		Tiny ImageNet \cite{Deng_CVPR_2009}    & 68.74 & 0.0619 &  0.0055 \\
		MS COCO \cite{Lin_ECCV_2014}           & 67.78 & 0.0562 &  0.0006 \\
		CUB-200 \cite{Wah_report_2011}         & 68.03 & 0.0460 &  0.0041 \\
		FGVC-aircraft \cite{Maji_report_2013}  & 67.05 & 0.0385 &  0.0159 \\
		Stanford-cars \cite{Krause_ICCVW_2013} & 67.41 & 0.0465 & -0.0028 \\
		\bottomrule	
	\end{tabular}}
\end{table}

	\begin{table}[!t]
	\centering
	\caption{\label{tbl:noise}
	    Effects of random noise on the performance (\%) with ResNet GEM. The noise follows a uniform distribution {\small $\mathcal{U}(-1,1)$} or a unit normal distribution {\small $\mathcal{N}(0,1)$}, and is used as predicted gradients.
	    The experimental details are described in Section \ref{subsec:noise}.
	}
	\vspace{-2ex}
	\adjustbox{width=1.0\columnwidth}{
	\begin{tabular}{L{22ex} C{9ex} C{9ex} C{9ex}}
		\toprule
		Setting & Accuracy  & BWT & FWT \\
		\cmidrule(lr){1-1} \cmidrule(lr){2-2} \cmidrule(lr){3-3} \cmidrule(lr){4-4}
		No noise            & 66.92 & 0.0132 & -0.0048  \\
		No noise + proposed & 68.74 & 0.0619 & 0.0055 \\
		\midrule
		$\mathcal{U}(-1,1)$  & 54.10 & 0.1978 & -0.0121 \\
		$\mathcal{U}(-1,1)$ + proposed  & 67.71 & 0.0533 & 0.0004 \\
		\midrule
		$\mathcal{N}(0,1)$  & 45.29 & 0.2121 & 0.0032 \\
		$\mathcal{N}(0,1)$ + proposed  & 67.08 & 0.0502 & 0.0007 \\
		\bottomrule	
	\end{tabular}}
\end{table}

\subsection{Using Random Noise as Pseudo Gradients} \label{subsec:noise}

To evaluate the efficacy of the proposed method, we use random noise as the predicted gradients. 
The random noise is either generated by a uniform distribution {\small $\mathcal{U}(-1,1)$} or a normal distribution {\small $\mathcal{N}(0,1)$}. 
The results with the same experimental set-up as \tabref{tbl:conn_cifar} are shown in \tabref{tbl:noise}.
Specifically, {\small $\mathcal{U}(-1,1)$} or {\small $\mathcal{N}(0,1)$} indicates that the corresponding noise is used to replace {\small $\bar{g}|_{\tilde{x}_{i}}$} (see line 13 in \algref{alg:gl}), while \textit{proposed} indicates that the corresponding noise is used to replace {\small $g|_{\tilde{x}_{i}}$} (see line 12 in \algref{alg:gl}) and they will be the input to the equations in line 13 in \algref{alg:gl}.
As shown, {\small $\mathcal{U}(-1,1)$} or {\small $\mathcal{N}(0,1)$} produces much lower accuracy than the other settings. Note that the random noise disturbs the training for all the tasks so that the accuracies of preceding tasks are low when computing the BWT scores for the current task. As discussed in \secref{subsec:perf}, this leads to high BWTs, according to the definition of BWT (\eqref{eq:BWT}).

\subsection{Effects of Number of Labeled/Unlabeled Images} \label{subsec:num_unlabeled}

To understand how the numbers of labeled and unlabeled images affect the performance of CL, we conduct an analysis to show the performance of using different amount (range from $0\%$ to $100\%$) of labeled and unlabeled images.
Without using any unlabeled images, it implies the methods is a regular SCL method.
We compare our proposed method with Meta-gradient \cite{Zhang_AISTATS_2021} and the results with different amount of labeled (unlabeled) images are reported in \tabref{tbl:cifar_diffsize} (\tabref{tbl:cifar_diffsize_unlabel}). 
For results in \tabref{tbl:cifar_diffsize}, we use 20\% of unlabeled images for training.
An observation is that the performance increases as more labeled images are used for training. On the contrary, using more unlabeled images, which follows very different distributions in comparison to the labeled images, does not always lead to better performance.
As discussed in Section \ref{subsec:tradeoff} and shown in \figref{fig:diff}, the distributions of unknown classes' samples could be very different from the ones of known classes' samples.
Therefore, using more unlabeled images of the unknown classes would lead to a performance drop.

\subsection{How Hyperparameters Range Across Datasets} \label{subsec:across_datasets}

\begin{figure}[!t]
	\centering
	\subfloat{\includegraphics[width=0.24\textwidth]{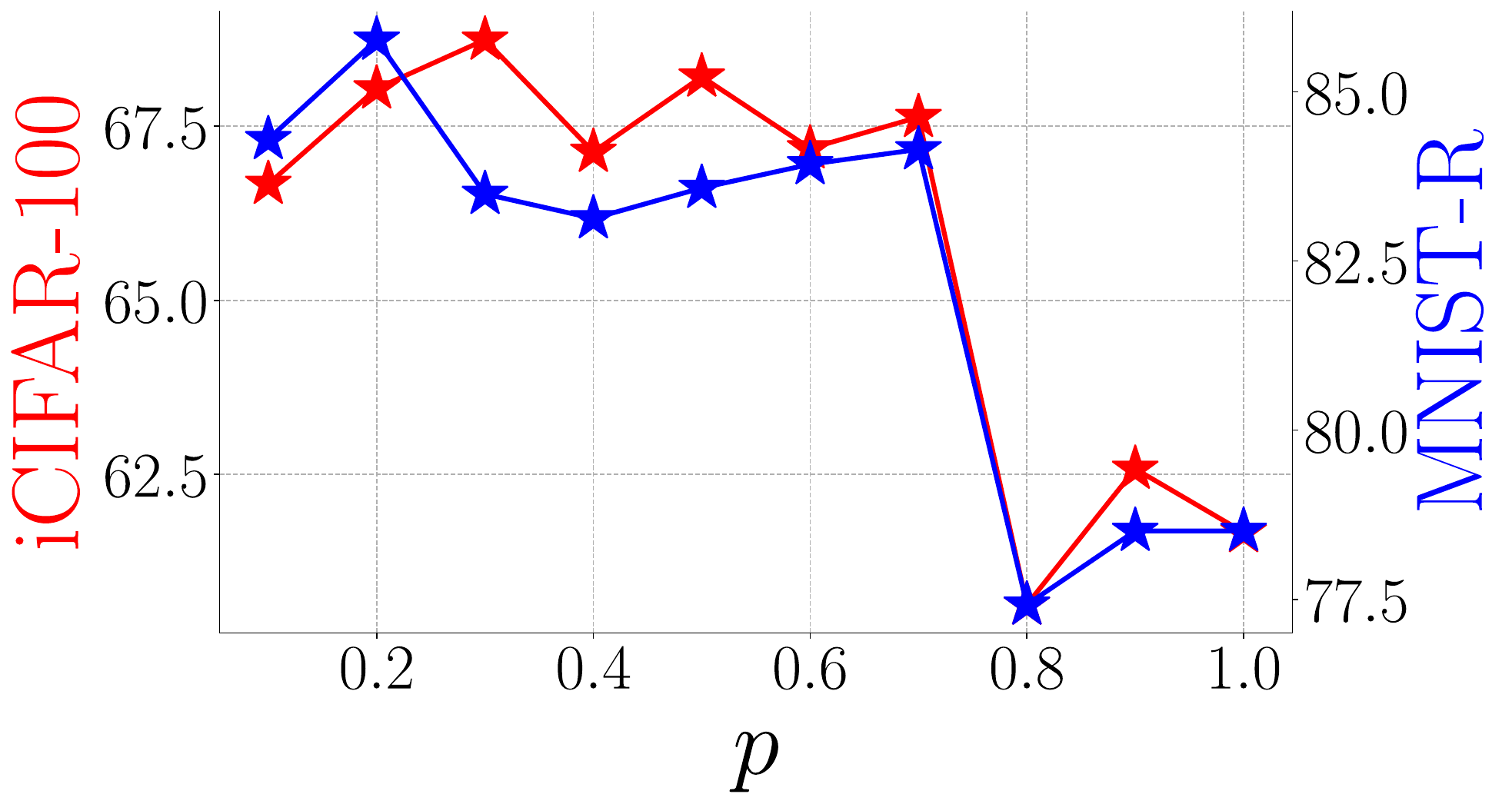}} 
	\subfloat{\includegraphics[width=0.24\textwidth]{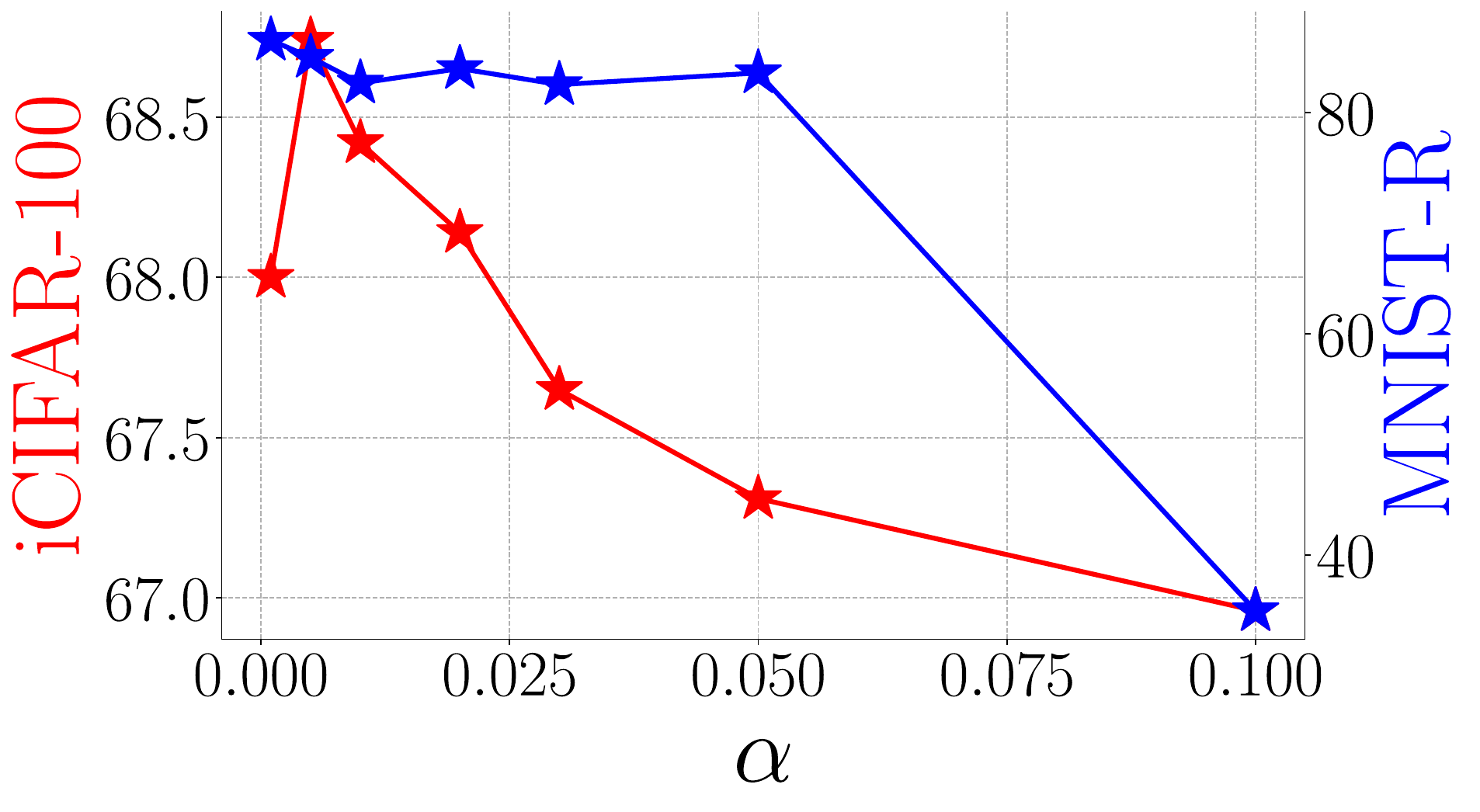}}
	\\
	\subfloat{\includegraphics[width=0.24\textwidth]{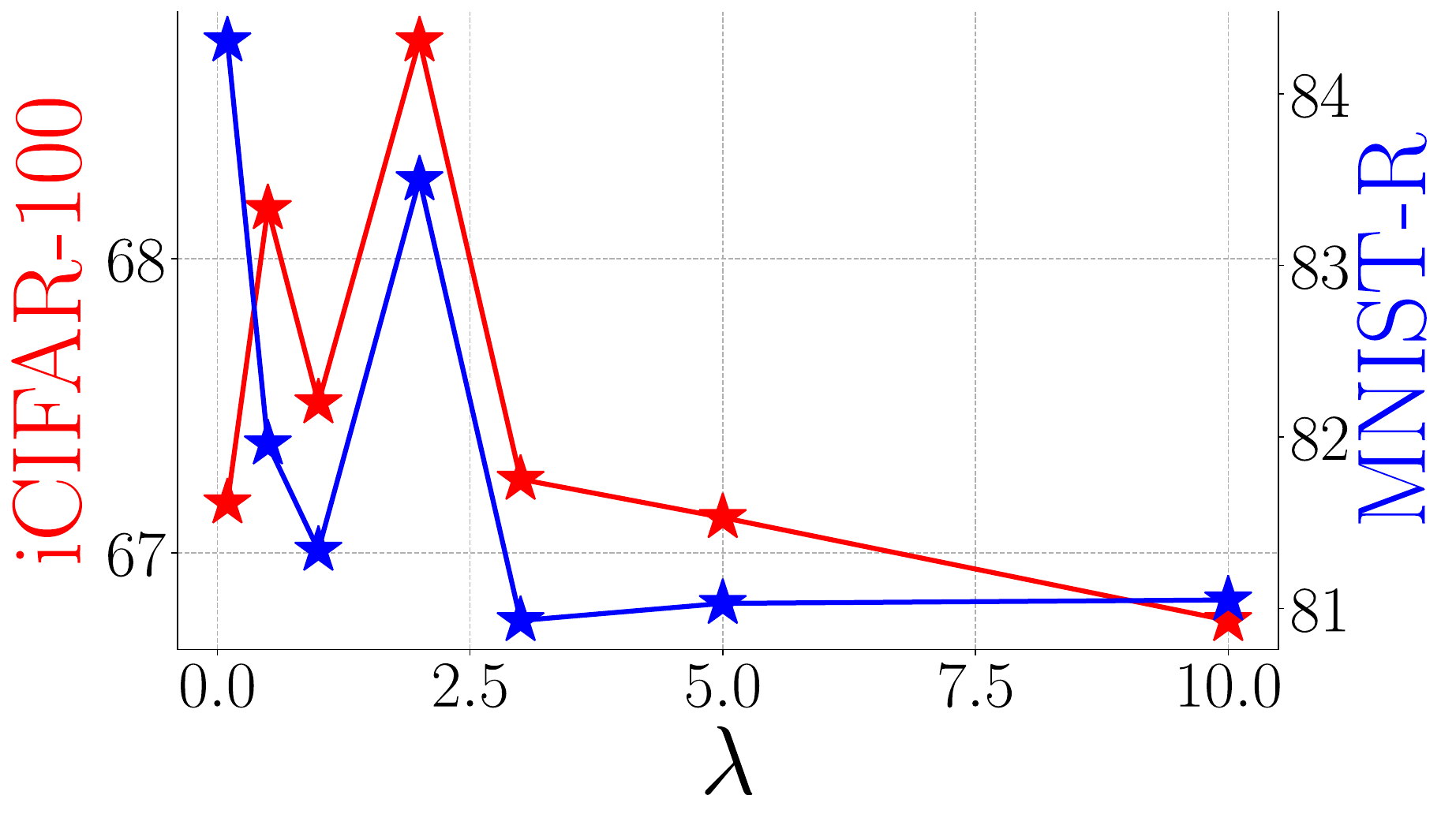}}
    \vspace{-1ex}
	\caption{\label{fig:across_ds}
		Effects of $p$ (top), $\alpha$ (middle), and $\lambda$ (bottom) on accuracy across datasets (\ie iCIFAR-100 and MNIST-R).
    	}
\end{figure}

In this section, we study how key hyperparameters $p$, $\alpha$, and $\lambda$ are robust to the training on different datasets when using the same unlabeled data. 
\figref{fig:across_ds} shows the curves of the accuracy w.r.t. $p$, $\alpha$, and $\lambda$. 
Overall, the curves w.r.t. iCIFAR-100 and MNIST-R are similar to each other. 
Specifically, as the values of $p$, $\alpha$, and $\lambda$ exceed a certain point, it would lead to a significant drop in accuracy. 
The hyperparameters used in this work (see \tabref{tbl:hyper}) are selected in the optimal range.

\begin{table}[!t]
	\centering
	\caption{\label{tbl:cifar_diffsize}
	    Effects of different numbers of labeled images on iCIFAR-100.
	    \textit{L-Ratio} indicates the amount of labeled images in iCIFAR-100 used for training. 
	    About 20\% of unlabeled images are sampled from Tiny ImageNet. 
	    The setting is the same as the one used in \tabref{tbl:conn_cifar} and ResNet GEM is used in this analysis.
	    }
	\vspace{-2ex}
	\adjustbox{width=1.0\columnwidth}{
	\begin{tabular}{L{9ex} C{12ex} C{9ex} C{9ex} C{9ex}}
		\toprule
		Method & L-Ratio & Accuracy & BWT & FWT \\
		\cmidrule(lr){1-1} \cmidrule(lr){2-2} \cmidrule(lr){3-3} \cmidrule(lr){4-4} \cmidrule(lr){5-5}
		\multirow{5}{*}{MG \cite{Zhang_AISTATS_2021}} & 20\% & 48.02  & 0.0420  &  0.0033  \\
		 & 40\% & 60.09 & 0.0822 & -0.0004  \\
		 & 60\% & 61.40 & 0.0699 & -0.0047  \\ 
		 & 80\% & 63.21 & 0.0541 & 0.0006  \\
		 & 100\% & 67.24  & 0.0614  &  -0.0001  \\ \midrule
		\multirow{5}{*}{Proposed} & 20\% & 50.38  & 0.0693  &  -0.0011  \\
		 & 40\% & 61.10 & 0.1045 & -0.0039  \\
		 & 60\% & 61.38 & 0.0582 & 0.0107  \\ 
		 & 80\% & 64.60 & 0.0552 & 0.0003  \\ 
		 & 100\% & 68.74 & 0.0619 & 0.0055  \\ 
		\bottomrule	
	\end{tabular}}
\end{table}


\begin{table}[t]
	\centering
	\caption{\label{tbl:cifar_diffsize_unlabel}
	    Effects of different numbers of unlabeled images on iCIFAR-100.
	    \textit{U-Ratio} is the amount of unlabeled images in Tiny ImageNet used for training. 
	    The setting is the same as the one used in \tabref{tbl:conn_cifar} and ResNet GEM is used in this analysis.}
	\adjustbox{width=1.0\columnwidth}{
	\begin{tabular}{L{9ex} C{12ex} C{9ex} C{9ex} C{9ex}}
		\toprule
		Method & U-Ratio & Accuracy & BWT & FWT \\
		\cmidrule(lr){1-1} \cmidrule(lr){2-2} \cmidrule(lr){3-3} \cmidrule(lr){4-4} \cmidrule(lr){5-5}
		\multirow{7}{*}{MG \cite{Zhang_AISTATS_2021}} & 0\% & 66.92 & 0.0132 & -0.0048  \\
		& 10\% & 66.43  & 0.0539  &  -0.0118  \\
		& 20\% & 67.24  & 0.0614  &  -0.0001  \\
		& 40\% & 67.03 & 0.0601 & -0.0010  \\
		 & 60\% & 66.95 & 0.0579 & -0.0037 \\ 
		 & 80\% & 66.71 & 0.0536 & -0.0059 \\
		 & 100\% & 66.27 & 0.0456 & 0.0018 \\ \midrule
		\multirow{7}{*}{Proposed} & 0\% & 66.92 & 0.0132 & -0.0048  \\
		& 10\% & 67.80  & 0.0525  &  0.0084  \\
		& 20\% & 68.74 & 0.0619 & 0.0055  \\
		 & 40\% & 67.53 & 0.0630 & 0.0140  \\
		 & 60\% & 67.99 & 0.0644 & 0.0099 \\ 
		 & 80\% & 67.91 & 0.0581 & -0.0021 \\ 
		 & 100\% & 66.96 & 0.0573 & 0.0000   \\ 
		\bottomrule	
	\end{tabular}}
\end{table}

\begin{figure}[!t]
	\centering
	\begin{minipage}{1.0\columnwidth}
	    \begin{minipage}{0.327\columnwidth}	\includegraphics[width=1\textwidth]{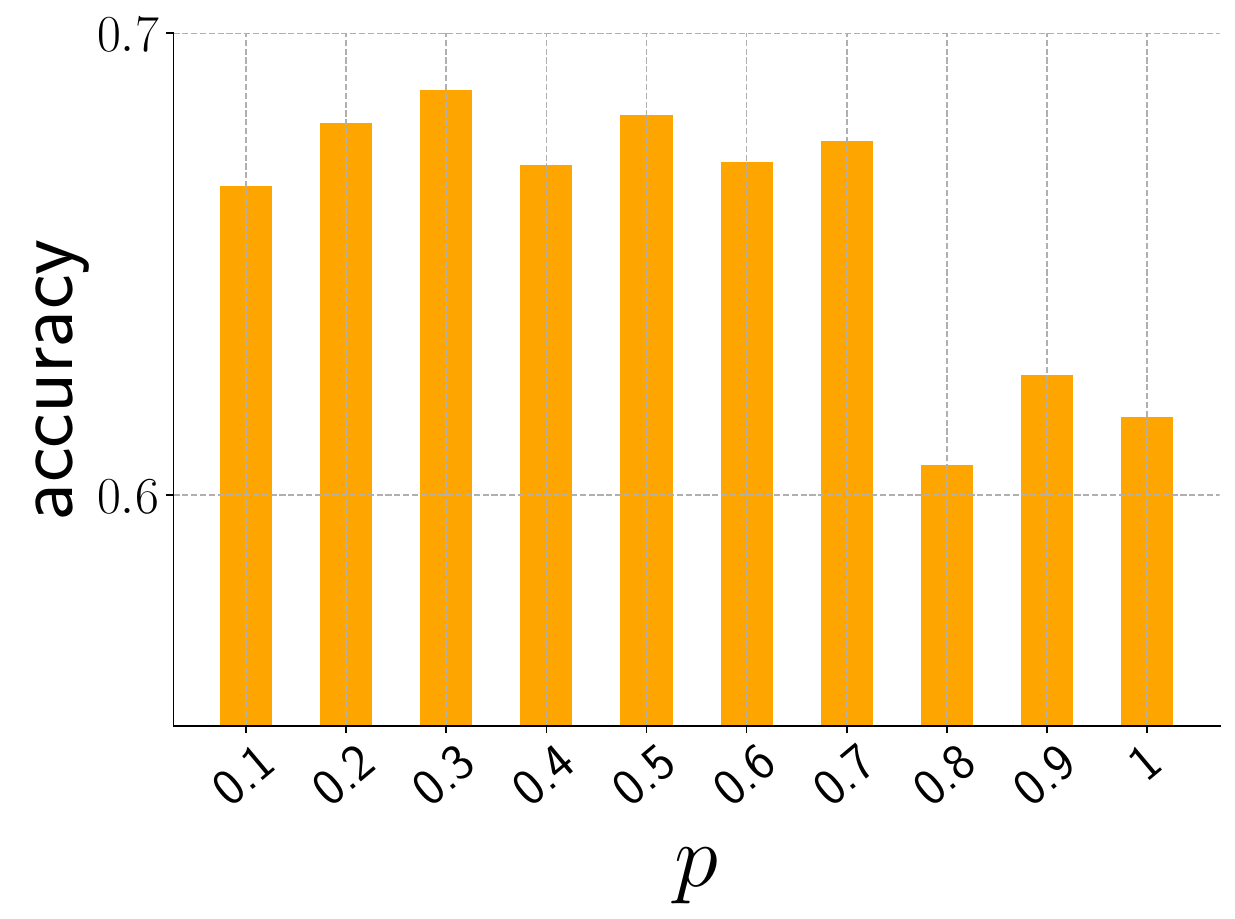}	    \end{minipage}
	    \begin{minipage}{0.327\columnwidth}	\includegraphics[width=1\textwidth]{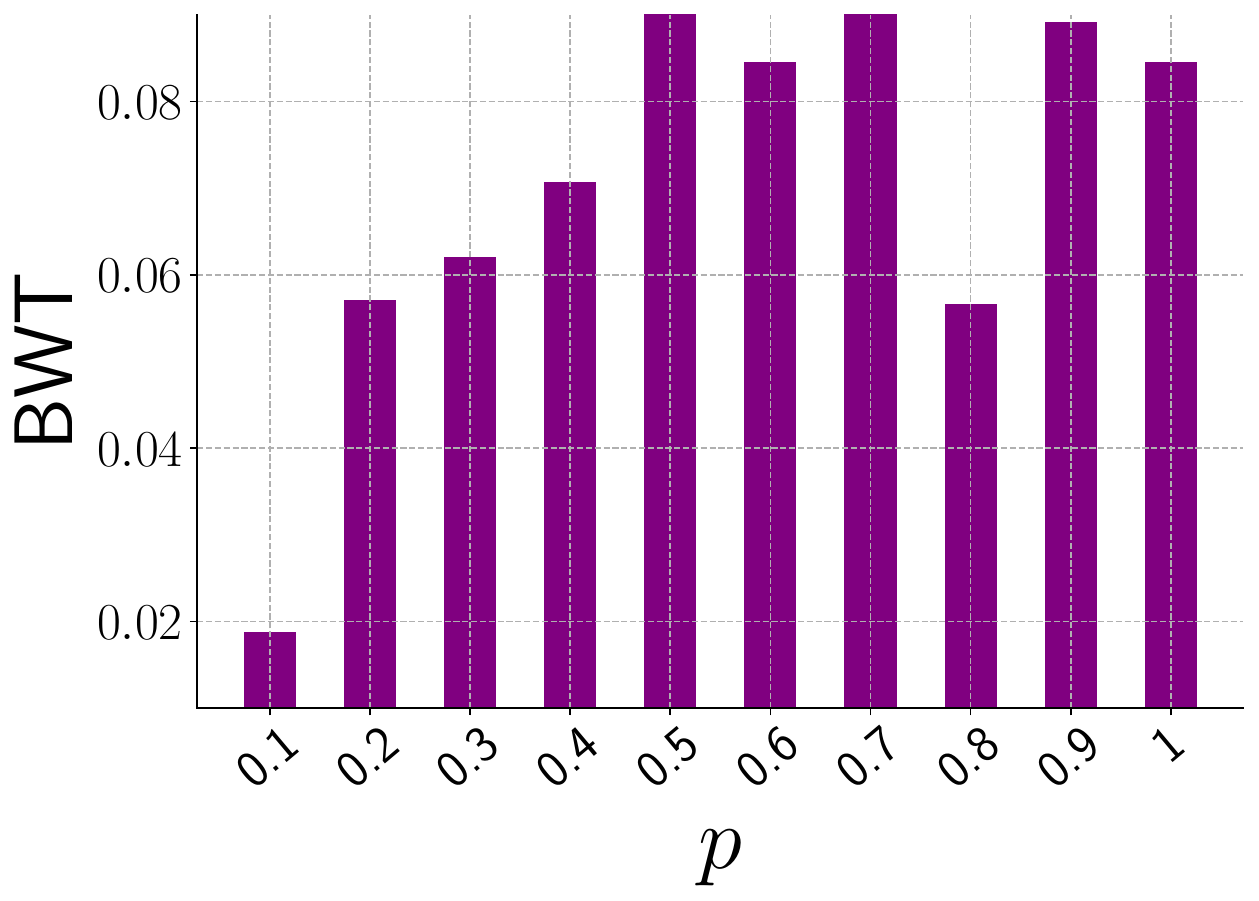}	            \end{minipage}
	    \begin{minipage}{0.327\columnwidth}	\includegraphics[width=1\textwidth]{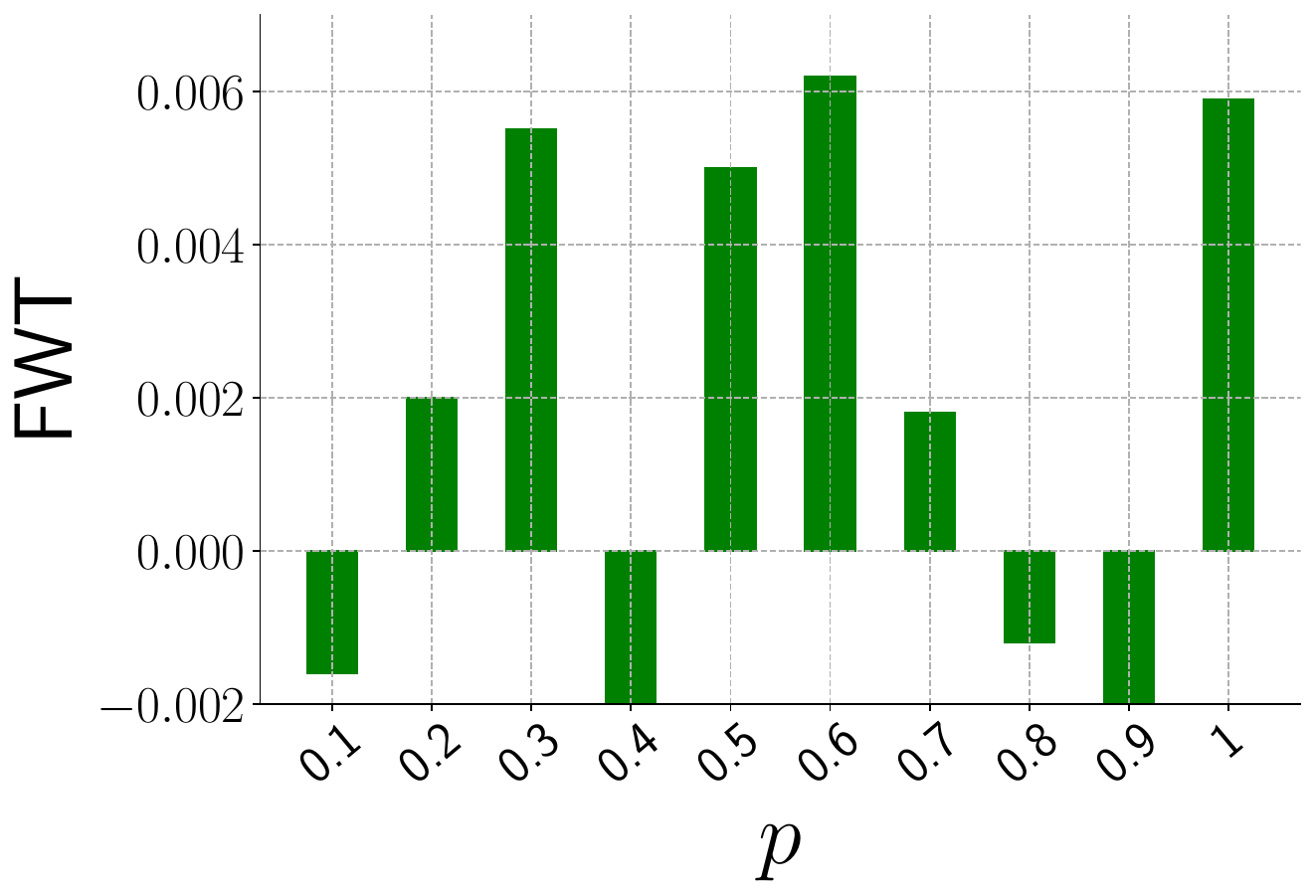}	            \end{minipage}
	\end{minipage}
	\begin{minipage}{1.0\columnwidth}
	    \begin{minipage}{0.327\columnwidth}	\includegraphics[width=1\textwidth]{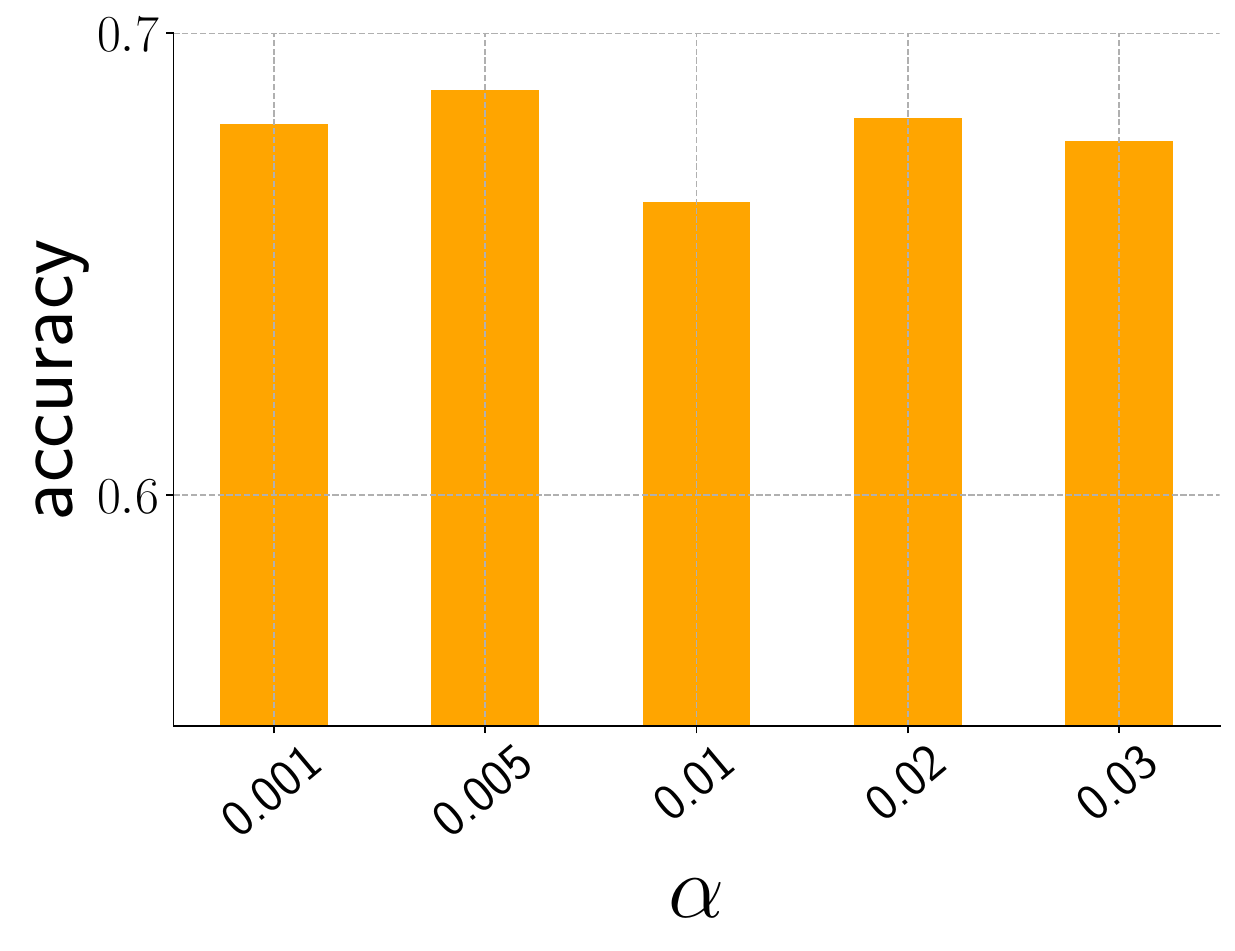}	\end{minipage}
	    \begin{minipage}{0.327\columnwidth}	\includegraphics[width=1\textwidth]{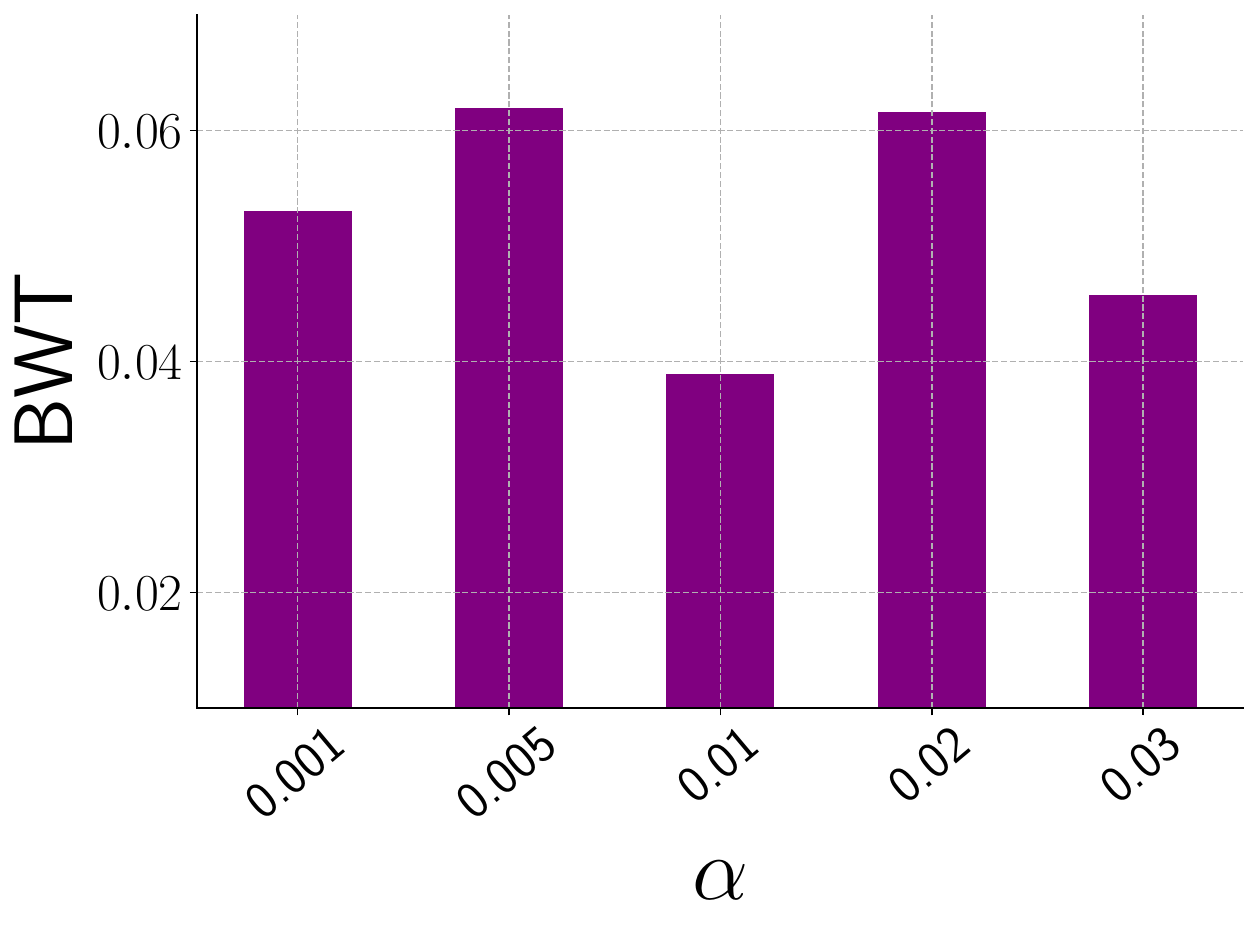}	        \end{minipage}
	    \begin{minipage}{0.327\columnwidth}	\includegraphics[width=1\textwidth]{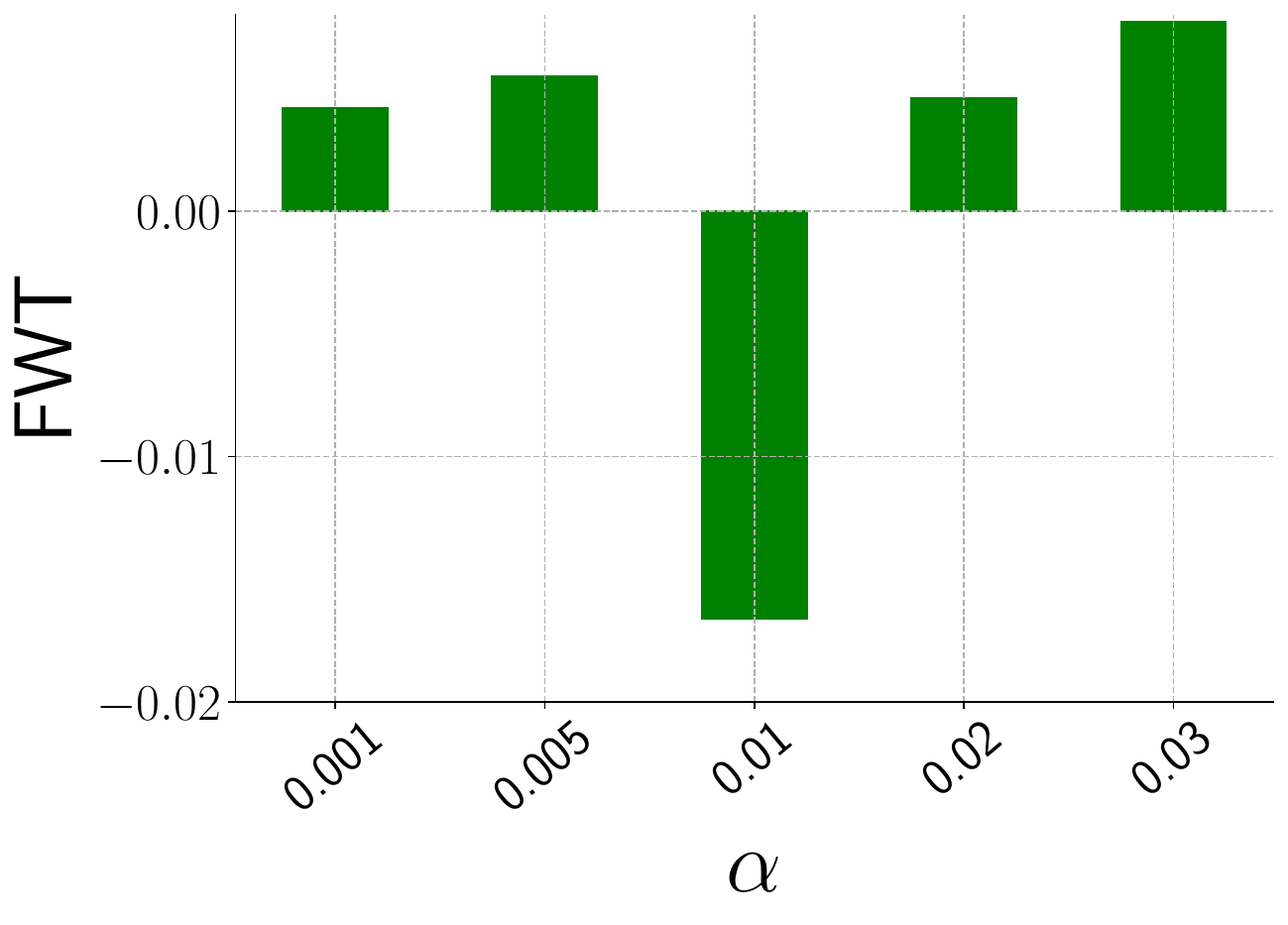}	        \end{minipage}
	\end{minipage}
	\begin{minipage}{1.0\columnwidth}
	    \begin{minipage}{0.327\columnwidth}	\includegraphics[width=1\textwidth]{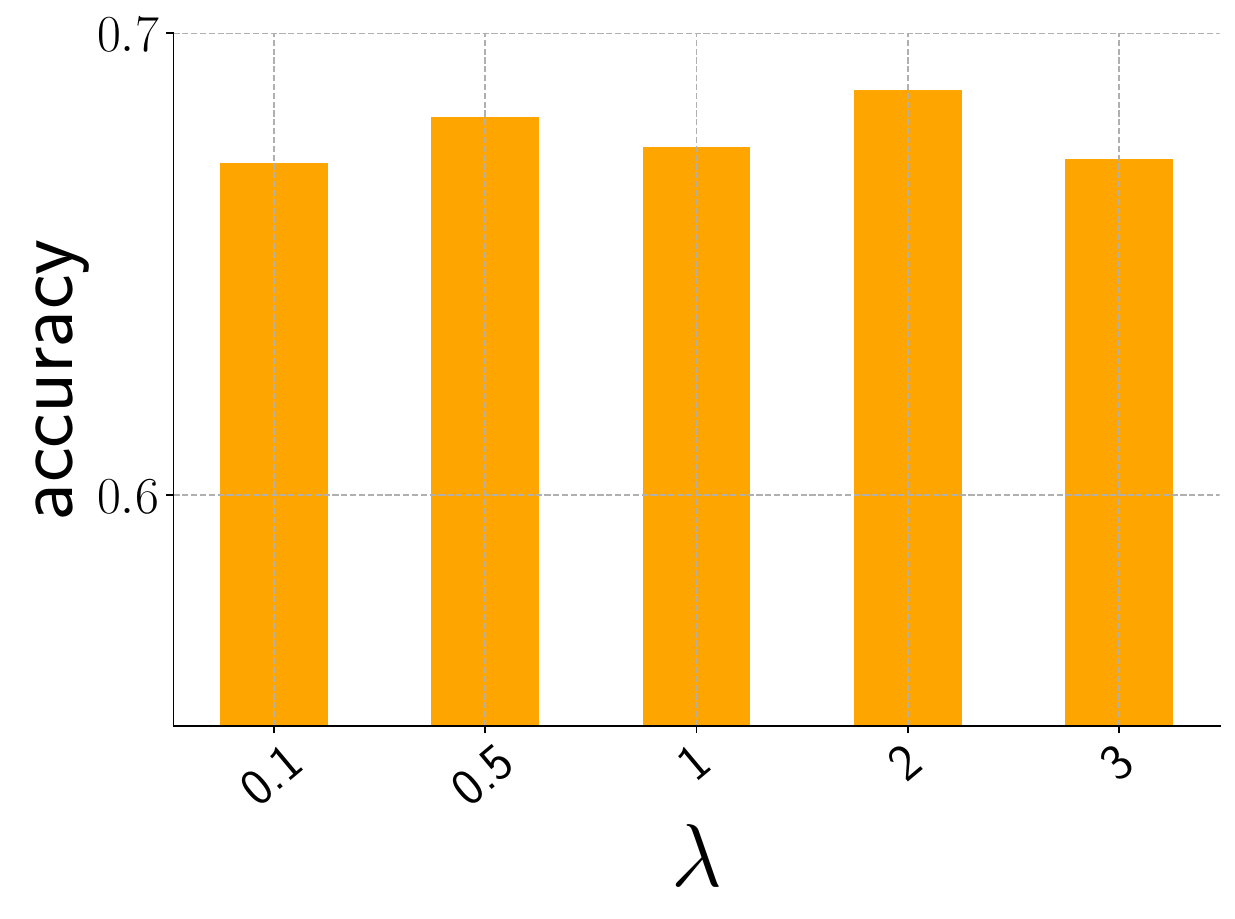}	\end{minipage}
	    \begin{minipage}{0.327\columnwidth}	\includegraphics[width=1\textwidth]{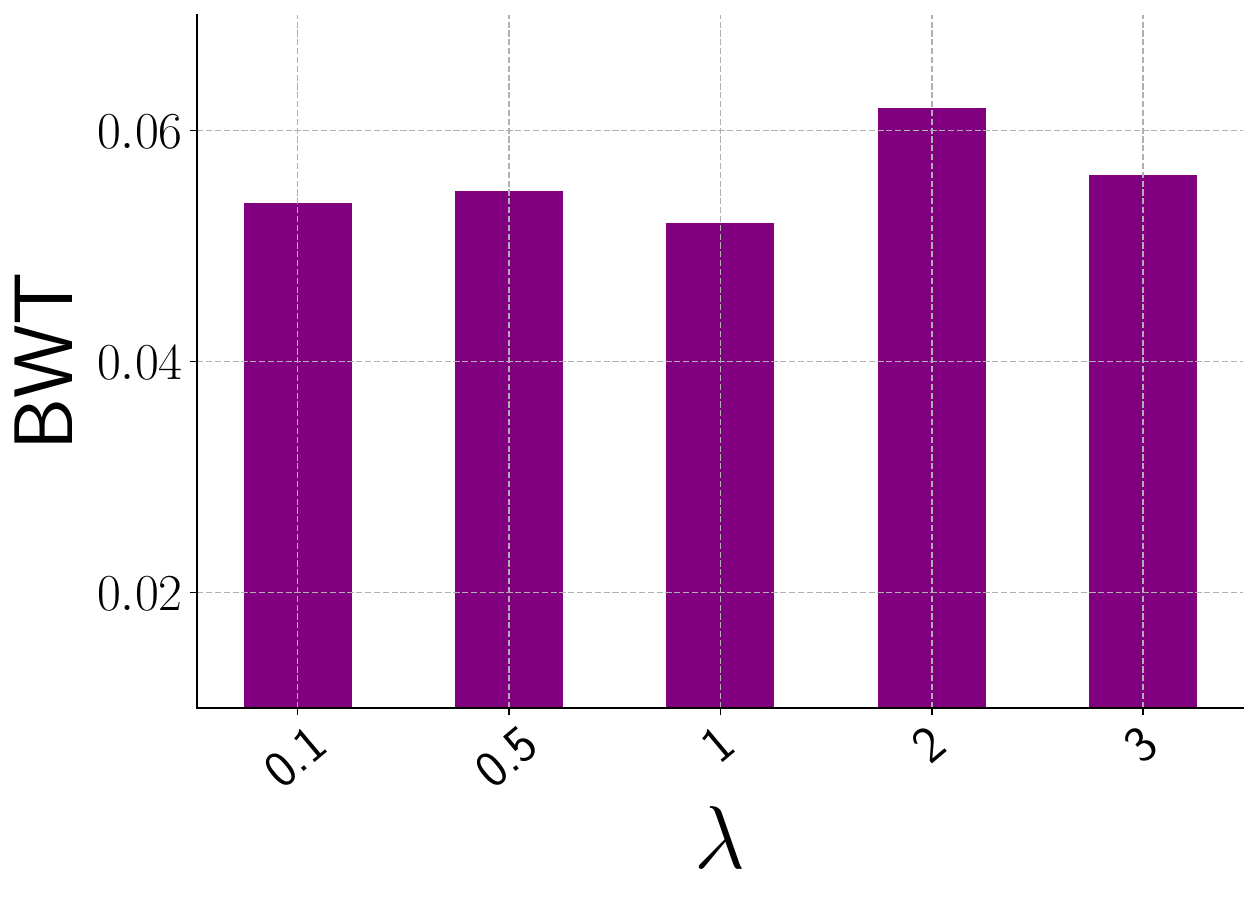}	    \end{minipage}
	    \begin{minipage}{0.327\columnwidth}	\includegraphics[width=1\textwidth]{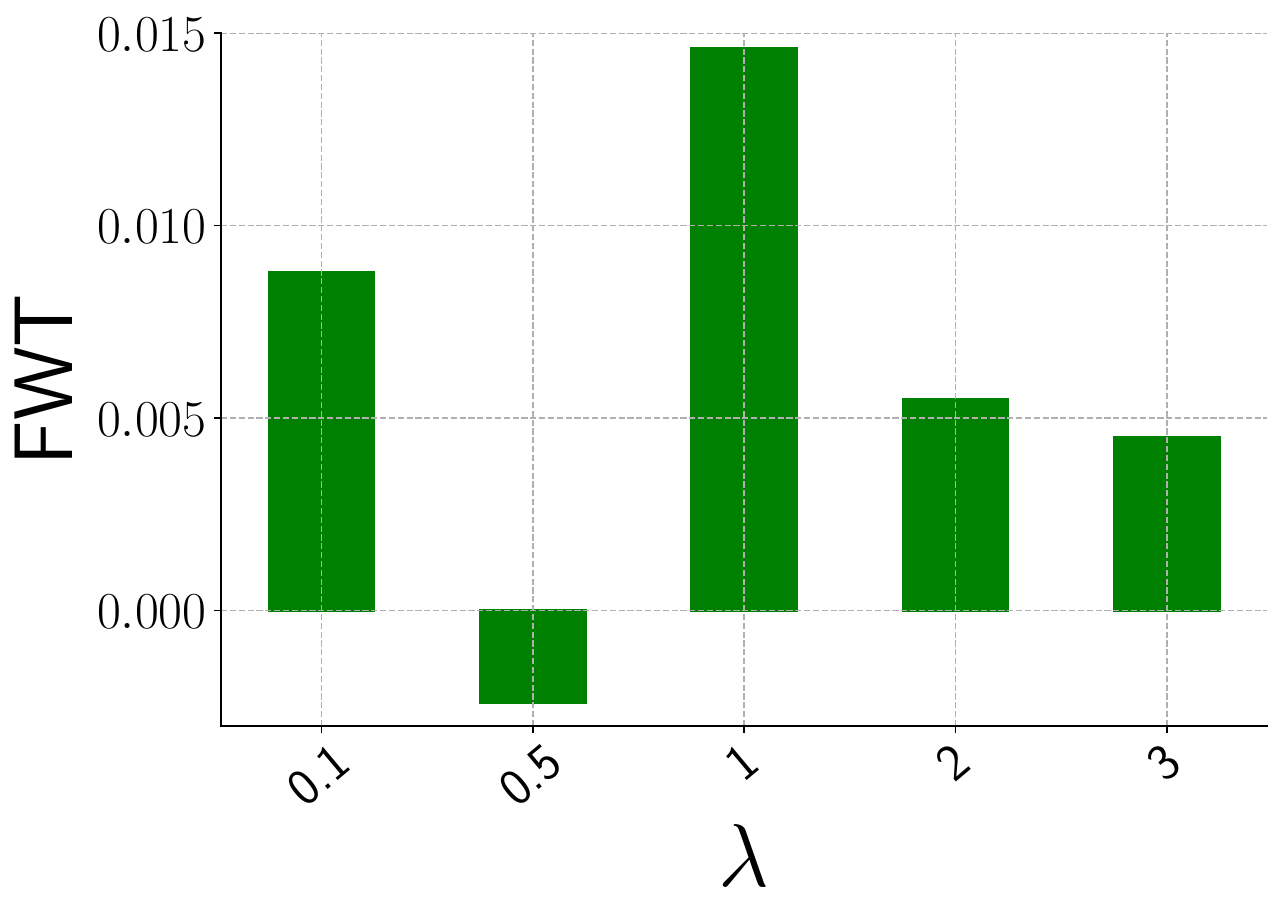}	    \end{minipage}
	\end{minipage}
	\begin{minipage}{1.0\columnwidth}
	    \begin{minipage}{0.327\columnwidth}	\includegraphics[width=1\textwidth]{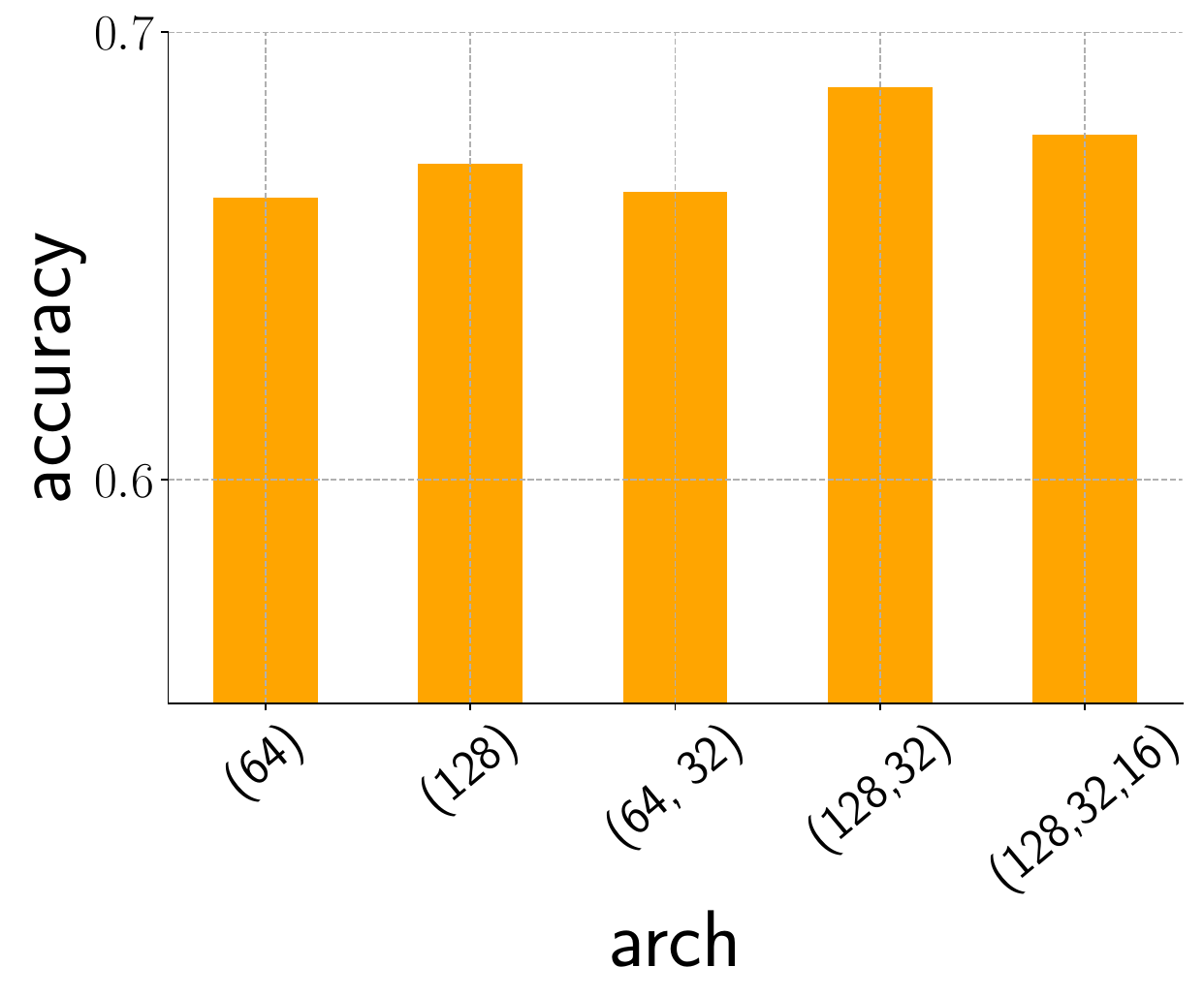}	    \end{minipage}
	    \begin{minipage}{0.327\columnwidth}	\includegraphics[width=1\textwidth]{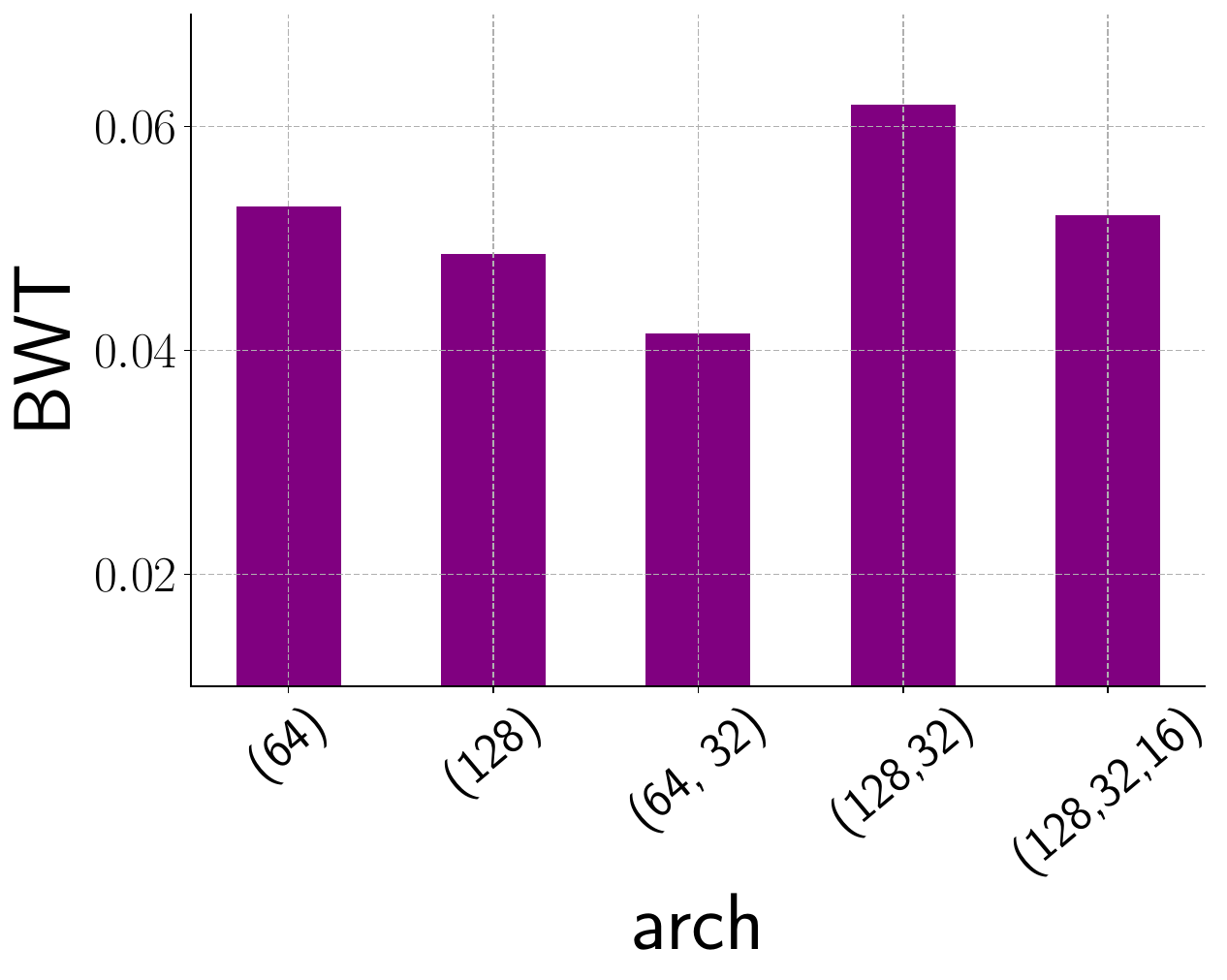}	        \end{minipage}
	    \begin{minipage}{0.327\columnwidth}	\includegraphics[width=1\textwidth]{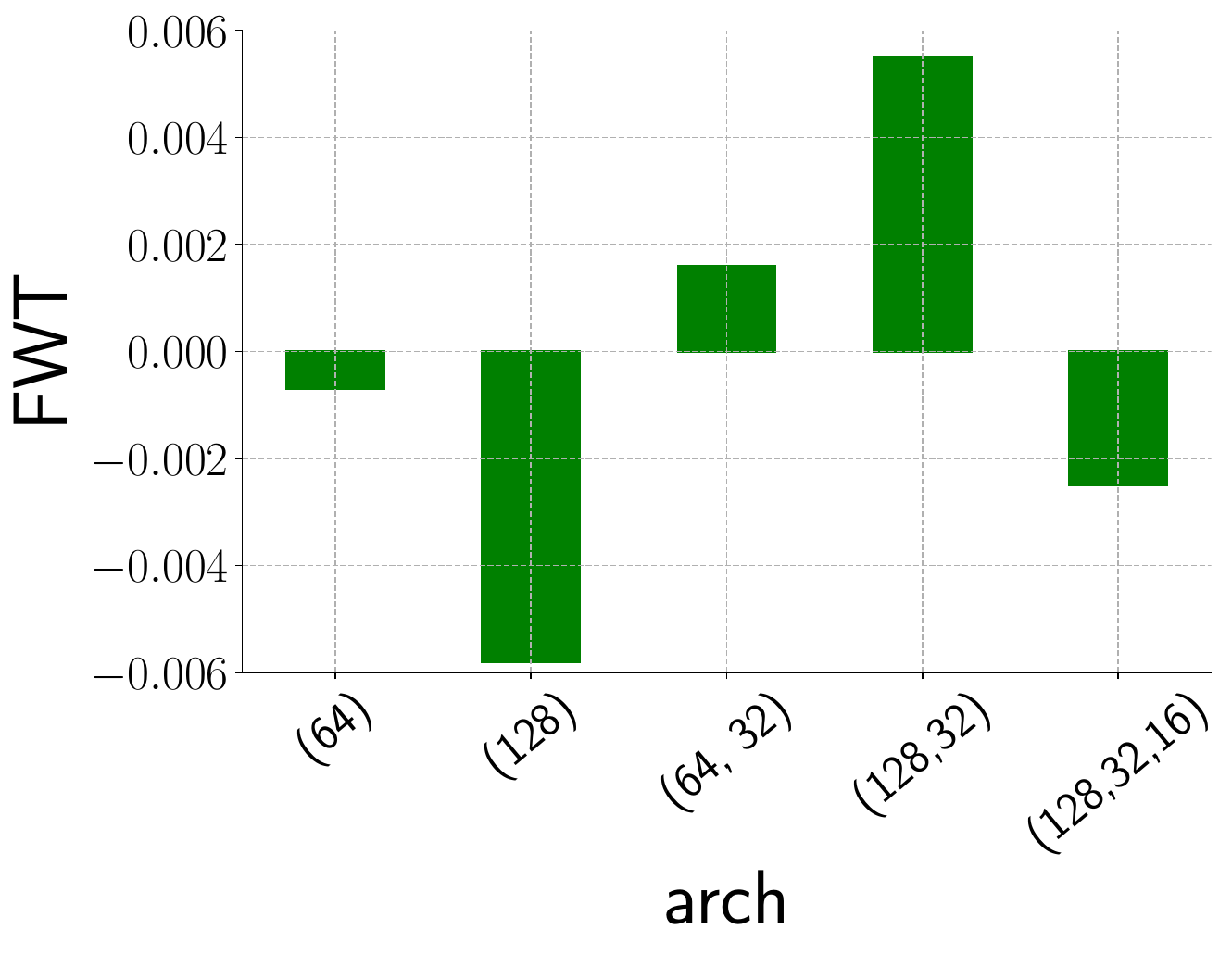}	        \end{minipage}
	\end{minipage}
	\begin{minipage}{1.0\columnwidth}
	    \begin{minipage}{0.327\columnwidth}	\includegraphics[width=1\textwidth]{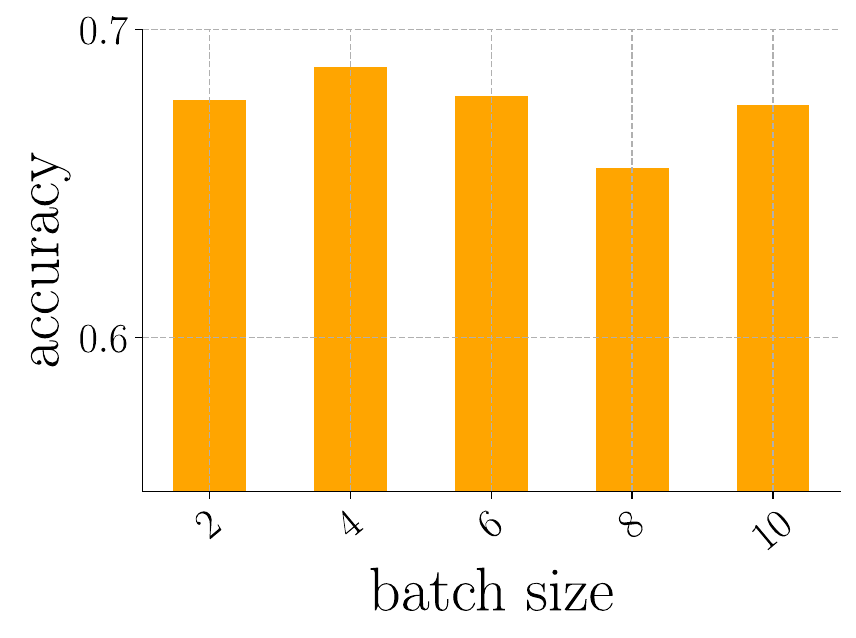}\end{minipage}
	    \begin{minipage}{0.327\columnwidth}	\includegraphics[width=1\textwidth]{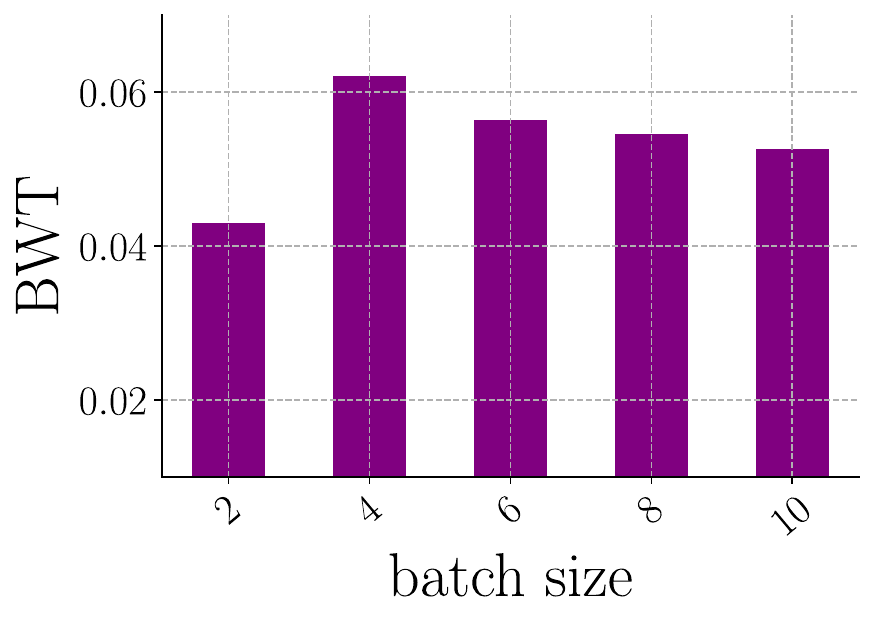}	    \end{minipage}
	    \begin{minipage}{0.327\columnwidth}	\includegraphics[width=1\textwidth]{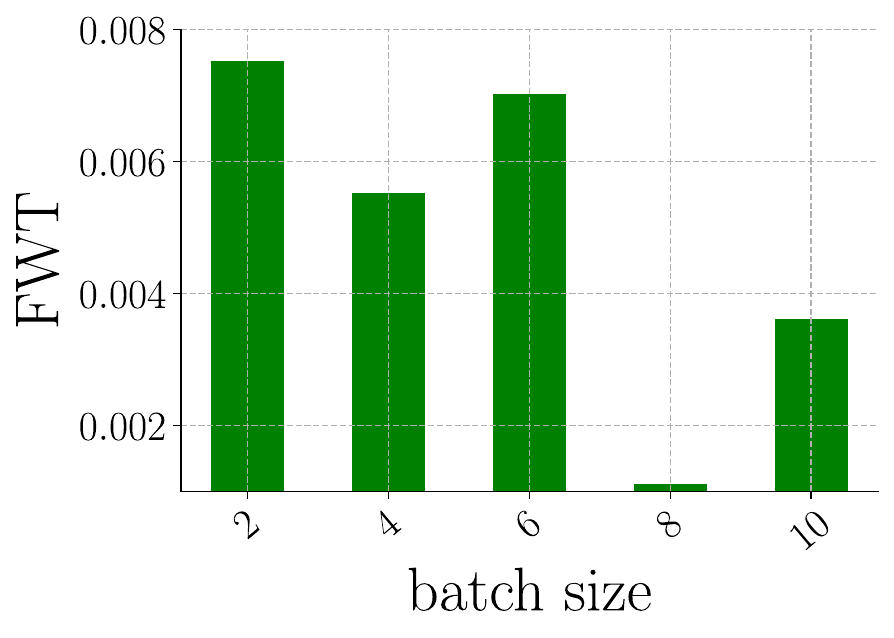}	    \end{minipage}
	\end{minipage}
	\vspace{-1ex}
	\caption{\label{fig:abl}
    	Ablation study of the proposed method with various hyperparameters detailed in Section~\ref{sec:exp}. 
    	The metrics are classification accuracy (left), BWT (middle), and FWT (right). 
    	ResNet GEM is used for the analysis.
    	}
\end{figure}

\begin{figure*}[!t]
	\centering
	\begin{tabular}{cccc}
	\footnotesize{~~~~~~~ResNet GEM} & \footnotesize{~~~~~~~ResNet DCL} & \footnotesize{~~~~~~~EffNet GEM} & \footnotesize{~~~~~~~EffNet DCL} \\
	\subfloat{\includegraphics[width=0.225\textwidth]{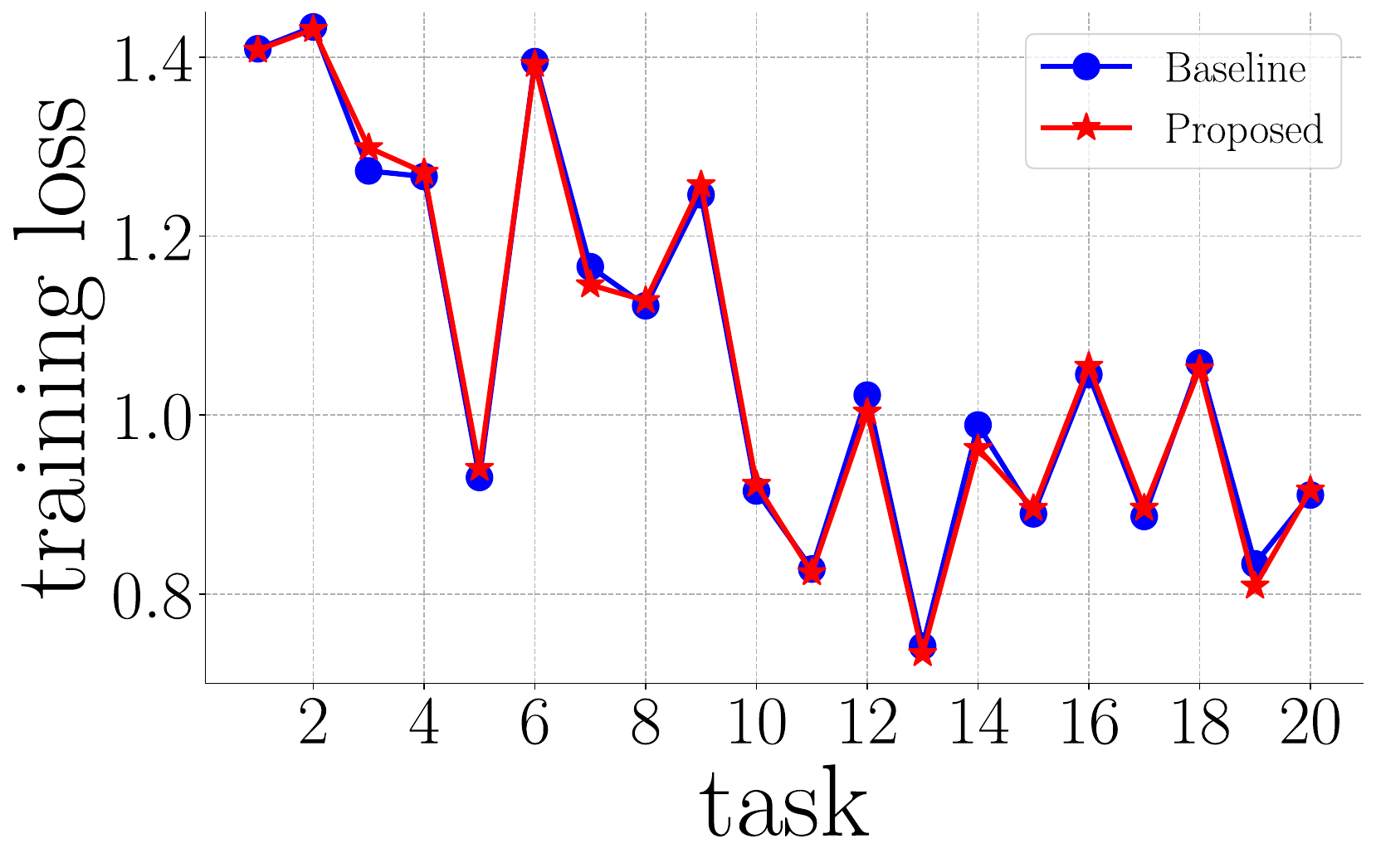}    } &
	\subfloat{\includegraphics[width=0.225\textwidth]{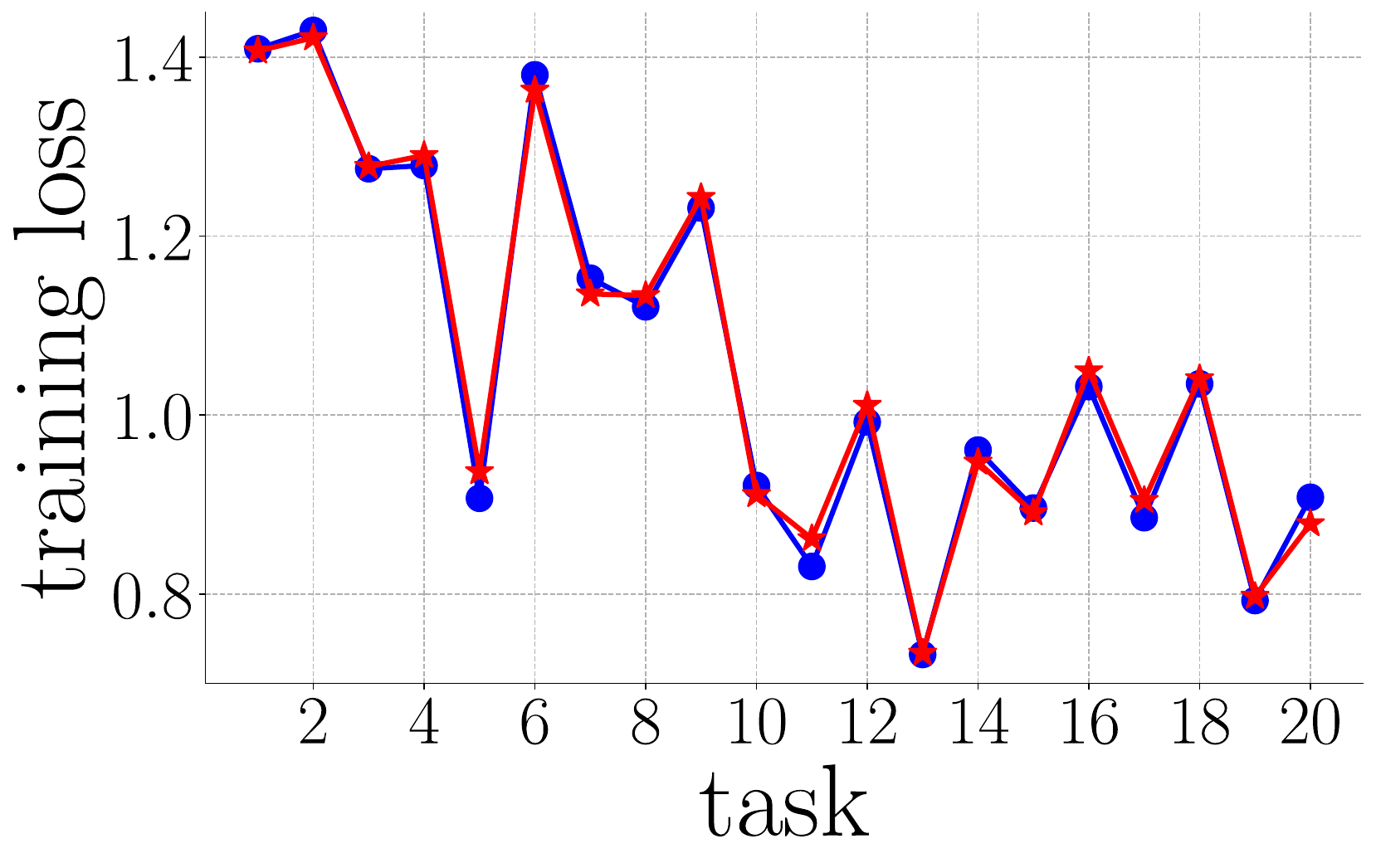}    } &
	\subfloat{\includegraphics[width=0.225\textwidth]{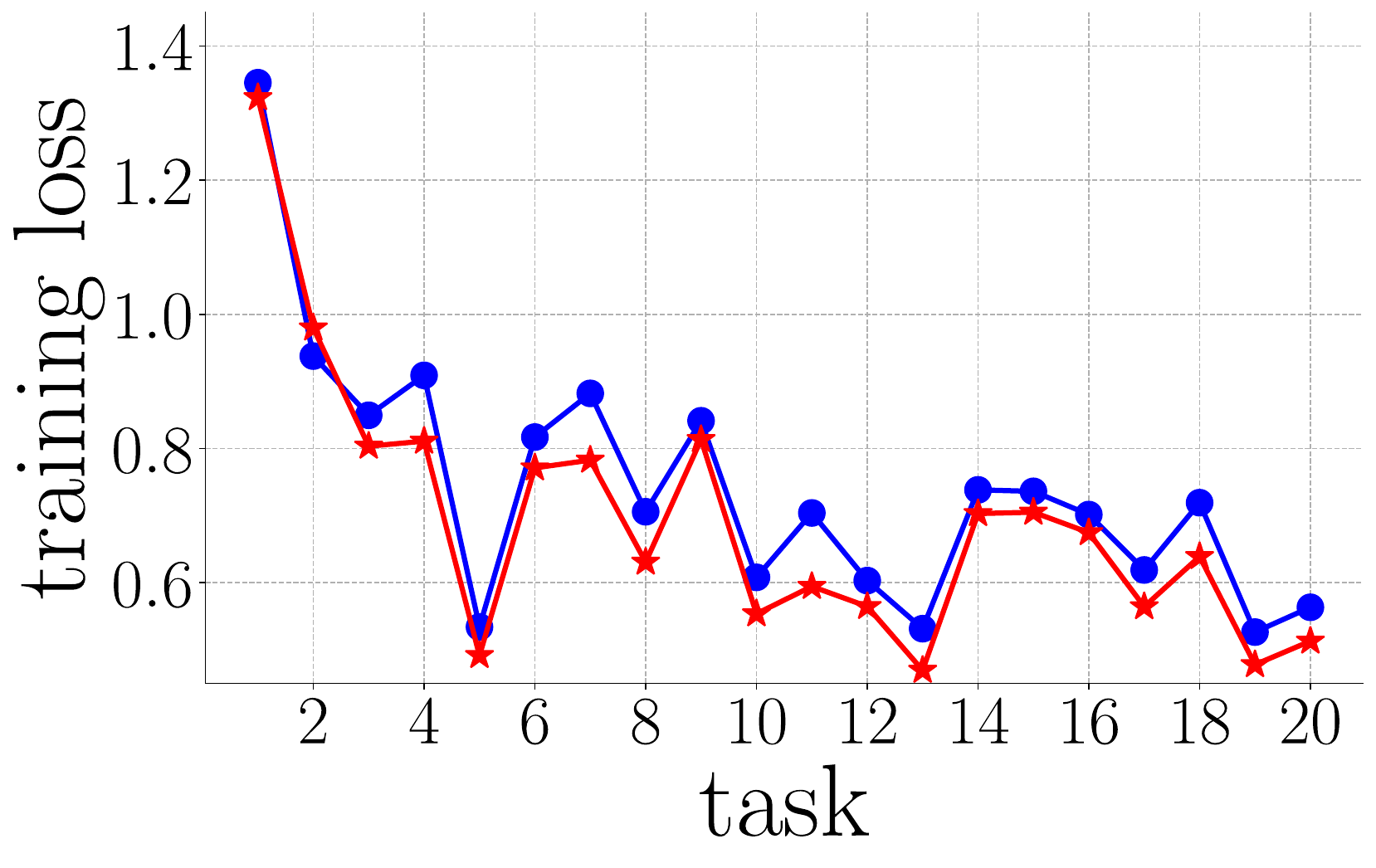}    } &
	\subfloat{\includegraphics[width=0.225\textwidth]{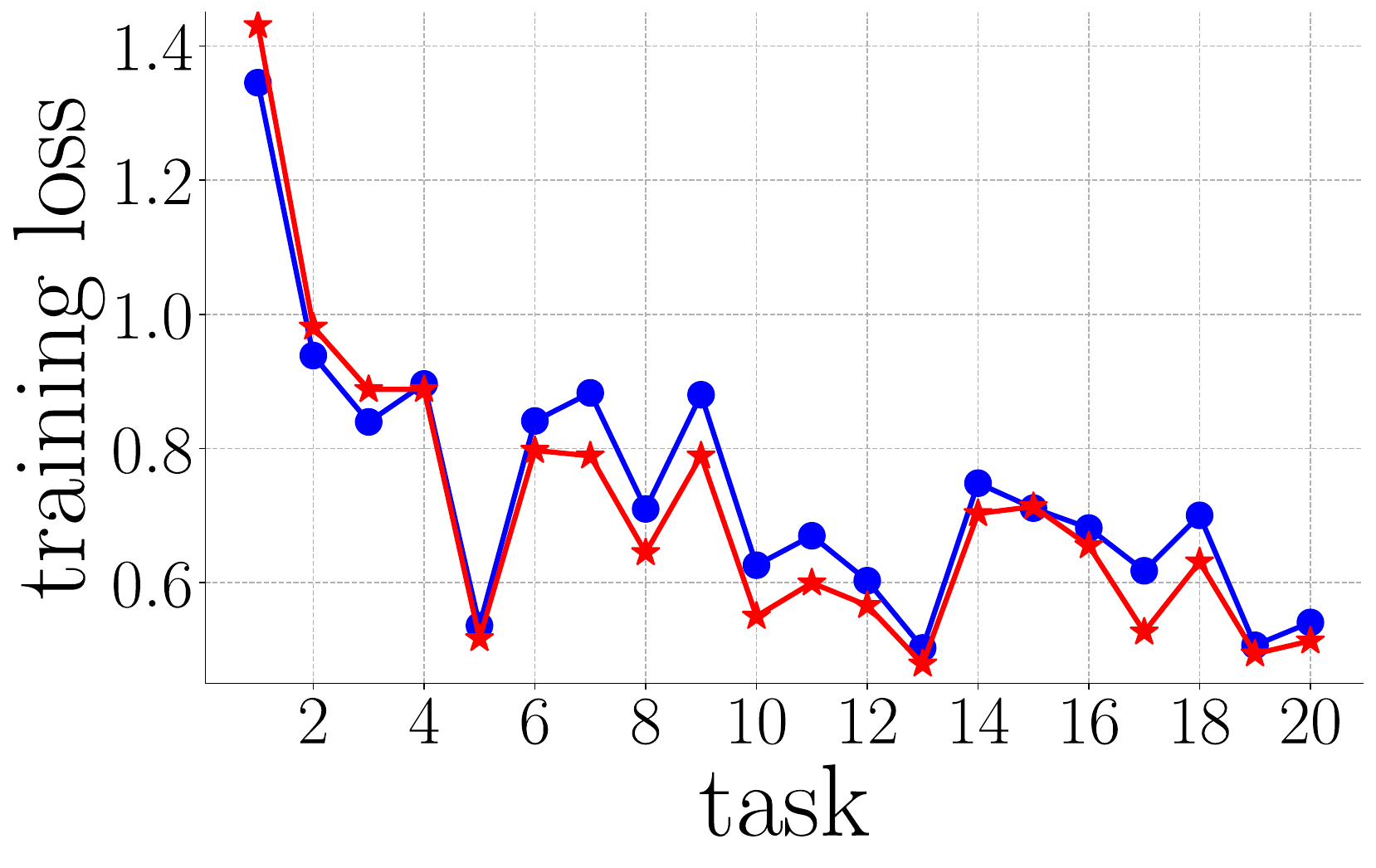}    } \\
	\subfloat{\includegraphics[width=0.225\textwidth]{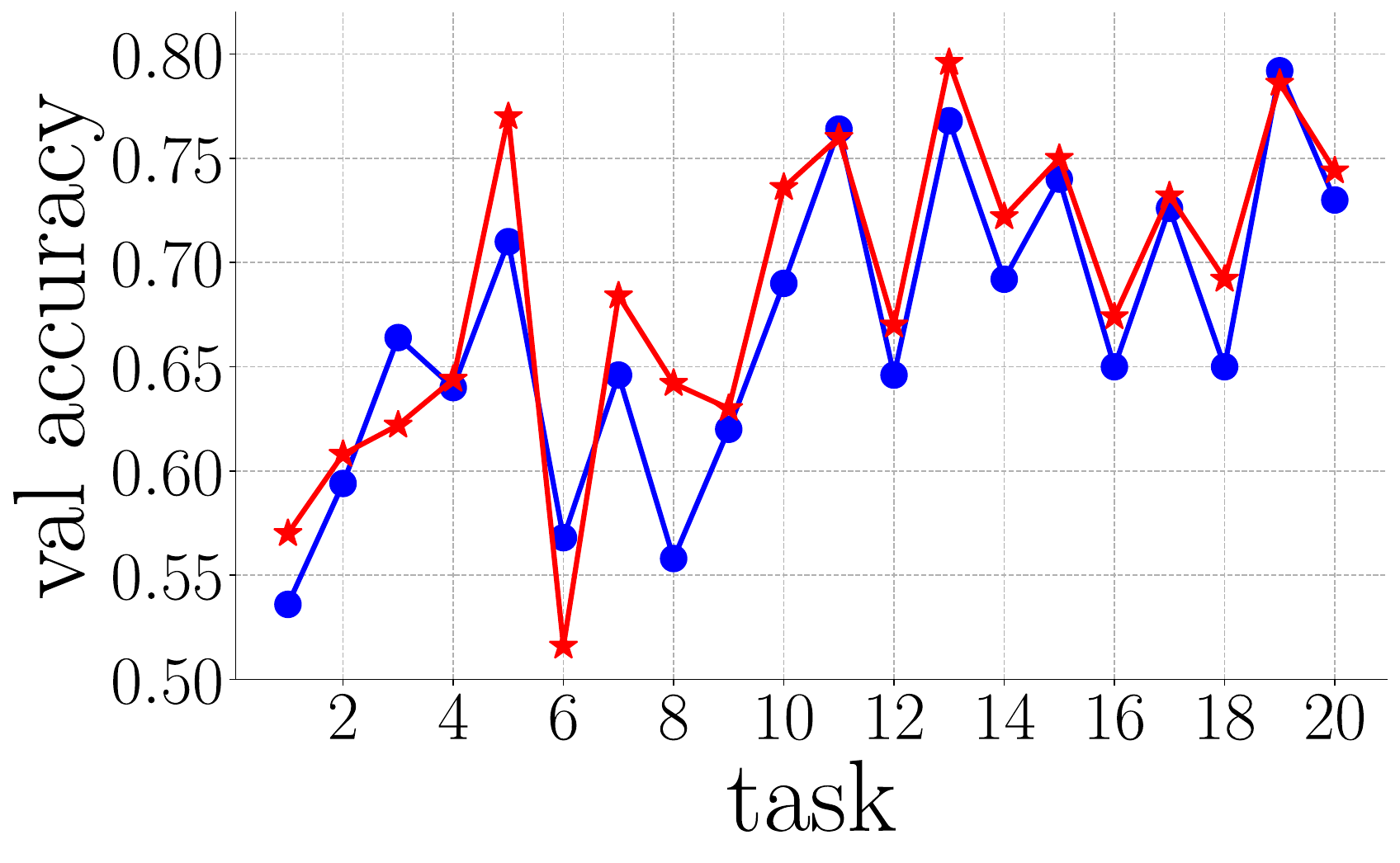}} &
	\subfloat{\includegraphics[width=0.225\textwidth]{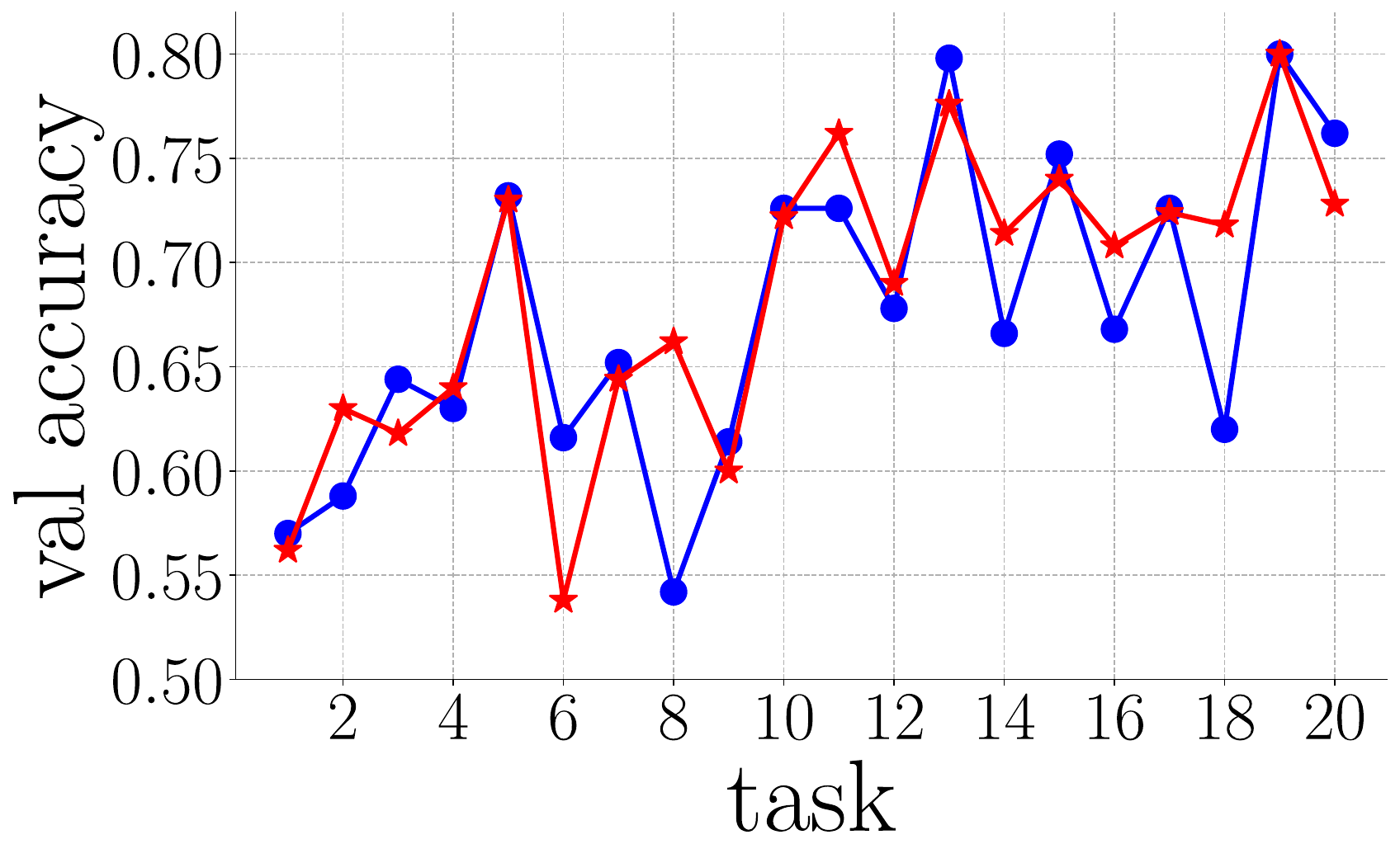}} &
	\subfloat{\includegraphics[width=0.225\textwidth]{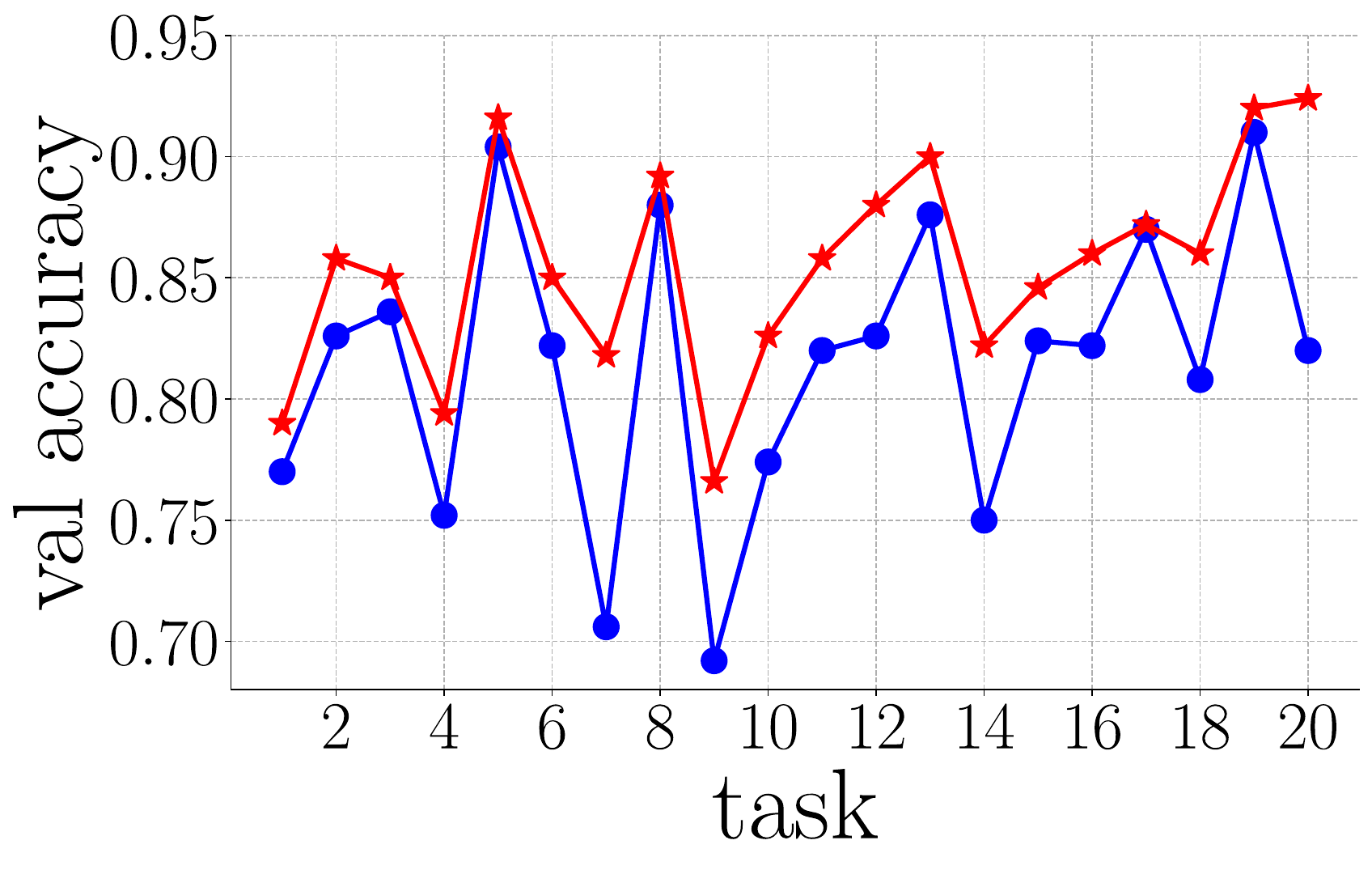}} &
	\subfloat{\includegraphics[width=0.225\textwidth]{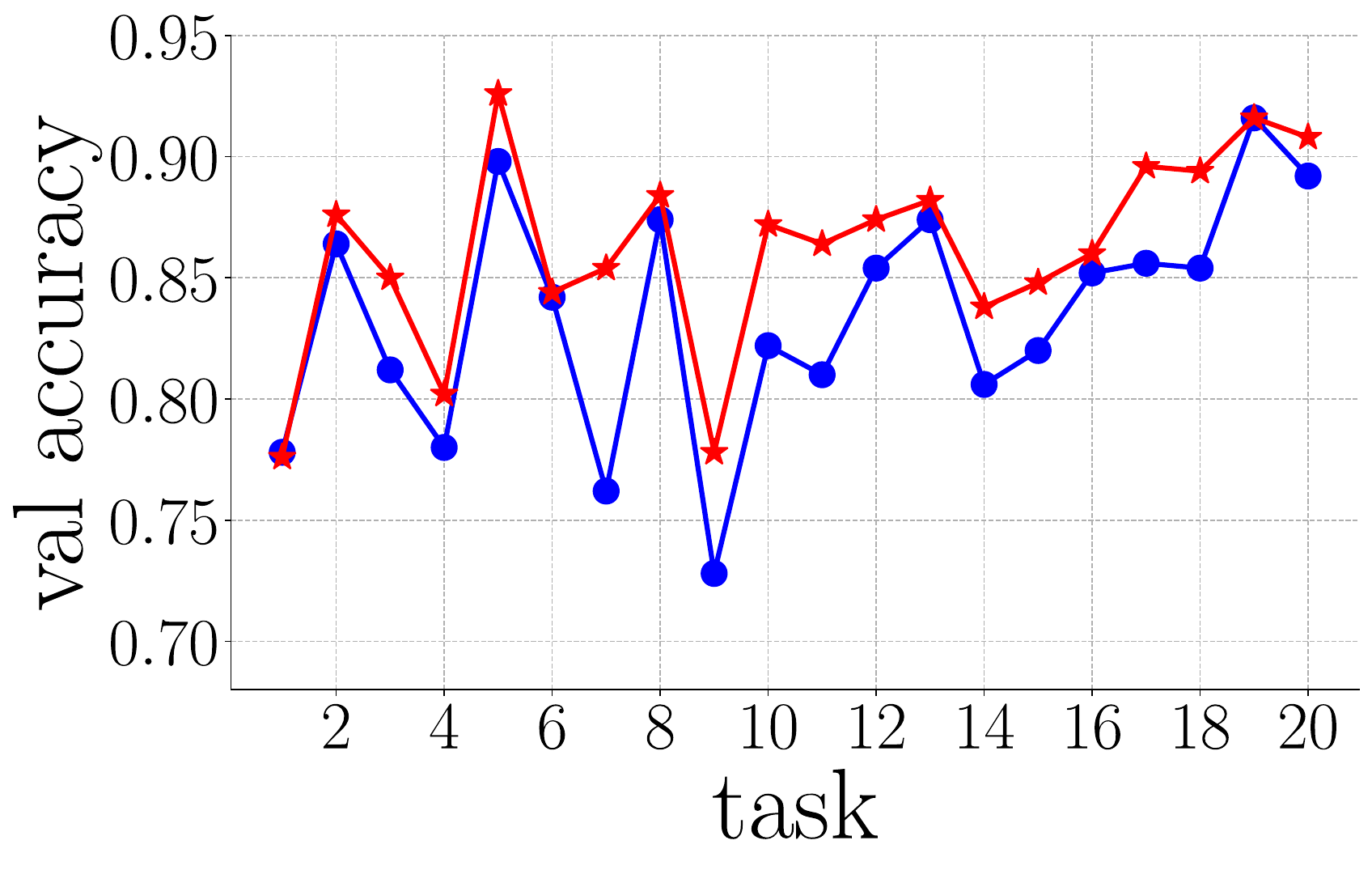}} \\
	\subfloat{\includegraphics[width=0.225\textwidth]{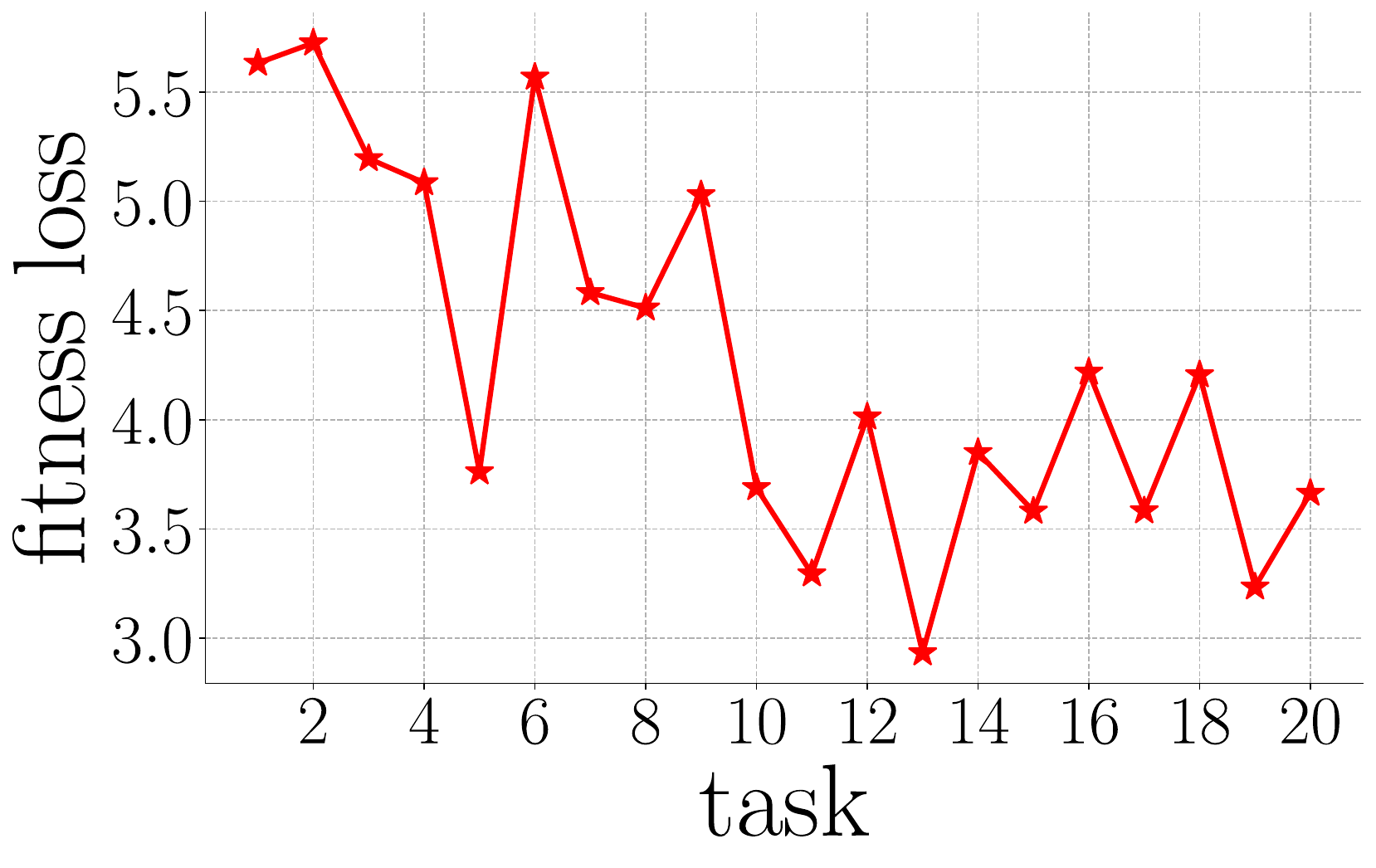} } &
	\subfloat{\includegraphics[width=0.225\textwidth]{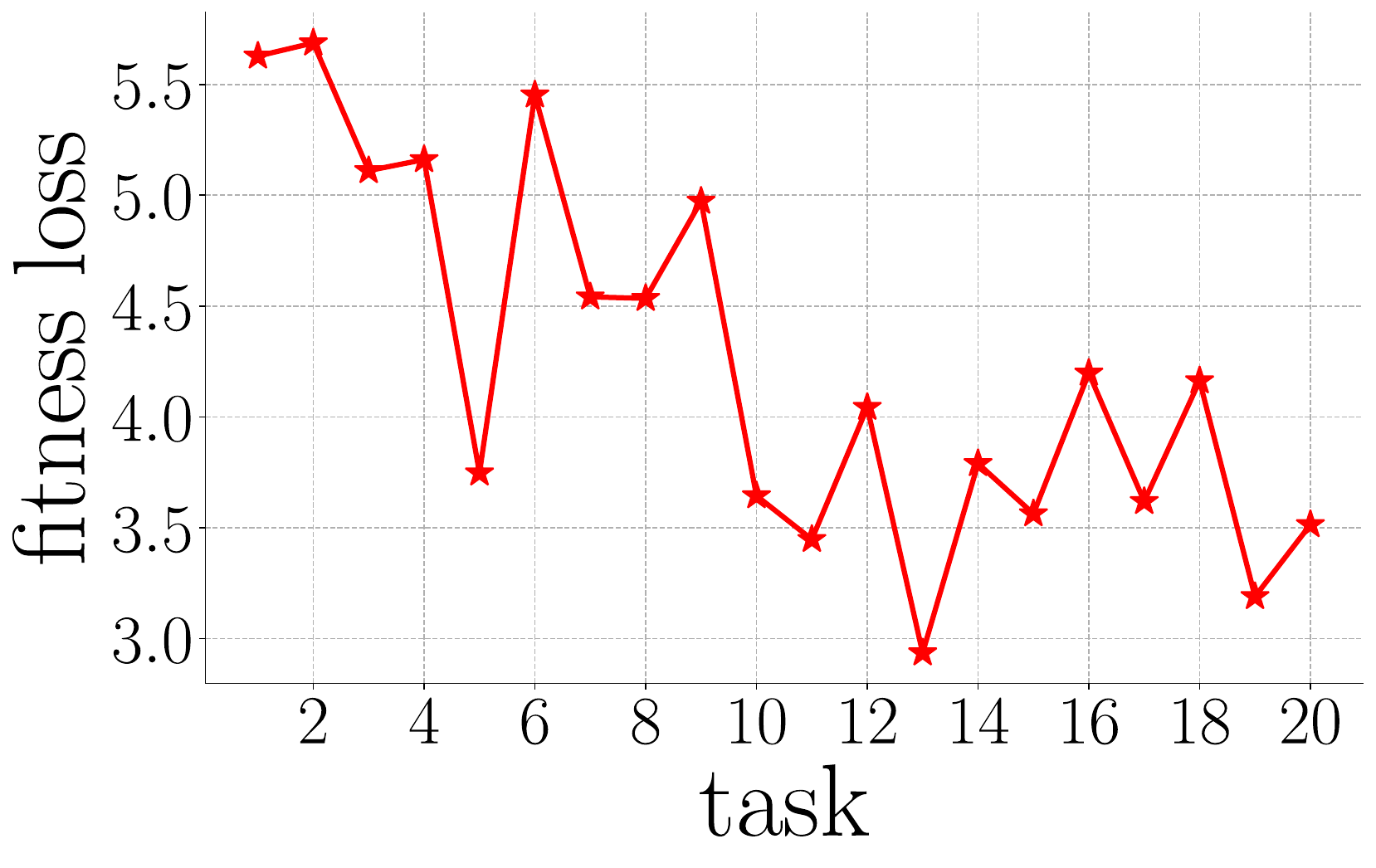} } &
	\subfloat{\includegraphics[width=0.225\textwidth]{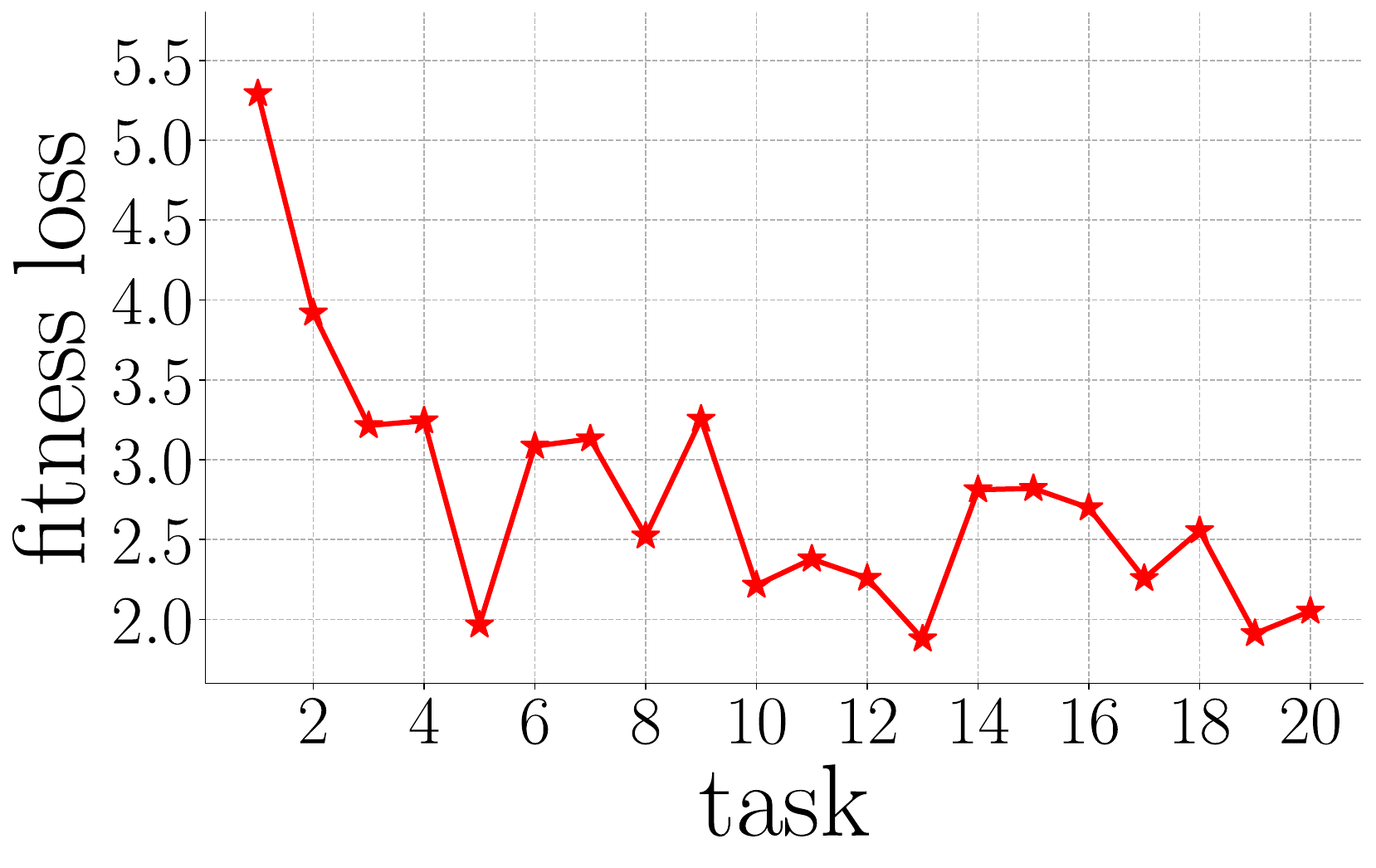} } &
	\subfloat{\includegraphics[width=0.225\textwidth]{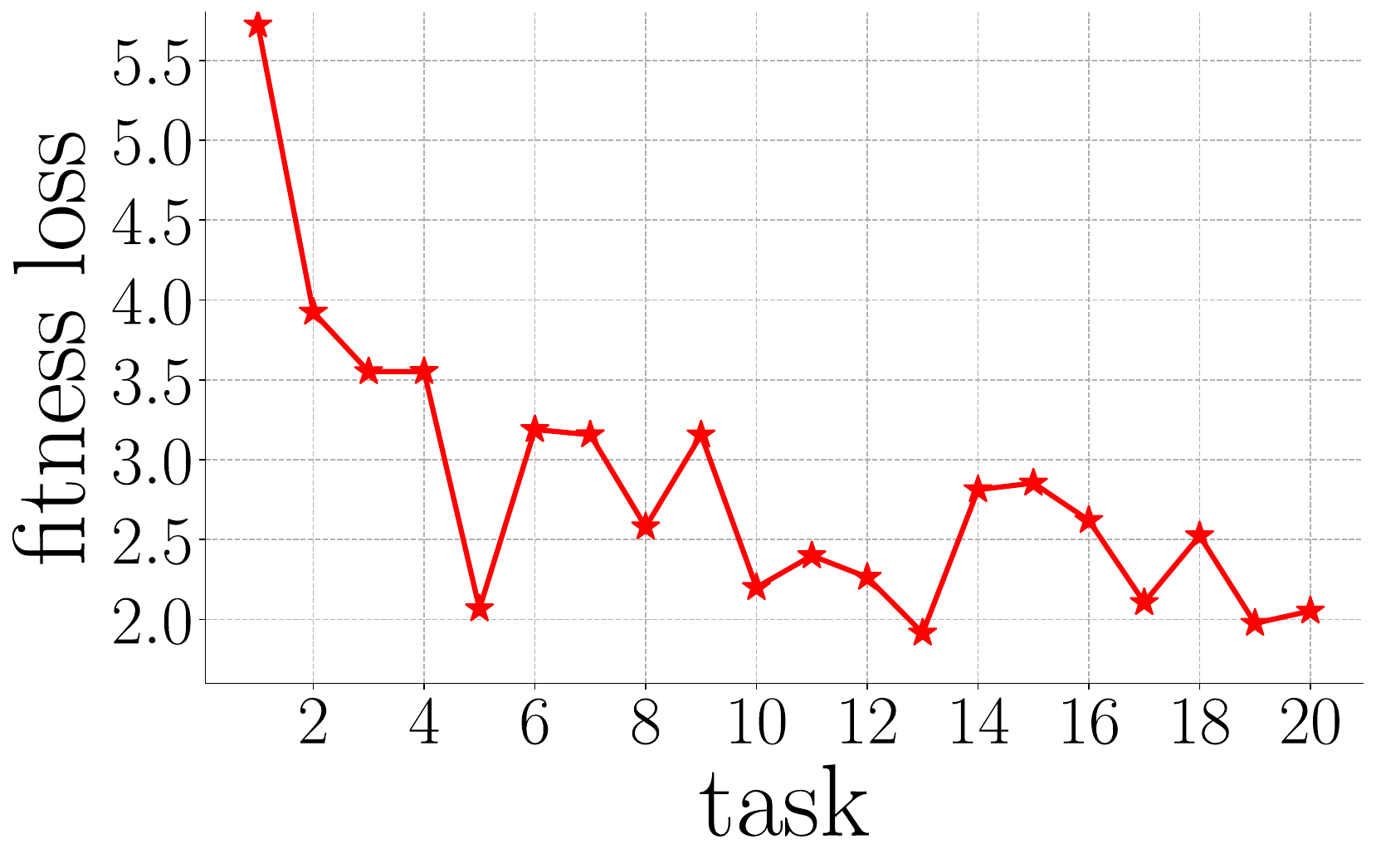} } \\
	\end{tabular}
	\vspace{-1em}
	\caption{\label{fig:loss_vs_accuracy}
    	Plots of the training loss curves (top row), the validation accuracy curves (middle row), and fitness loss curves (bottom row) on iCIFAR-100 with various pairs of methods and backbones.
    	}
\end{figure*}

\subsection{Ablation Study}

As introduced in the experimental set-up, the proposed method depends on five hyperparameters. 
This section shows the corresponding ablation studies and the results are shown in \figref{fig:abl}. 
As discussed in \secref{sec:method}, $p$ reflects the trade-off between overwhelming and generalizing. 
As $p$ increases, the accuracy drops significantly. This is as expected in the earlier discussion. 
Moreover, we can observe that the architecture of the proposed gradient learner is more critical to the proposed method in terms of accuracy, BWT, and FWT, comparing to the other hyperparameters.

\subsection{Training Loss, Validation Accuracy, and Fitness Loss}

The losses and accuracy against tasks are shown in \figref{fig:loss_vs_accuracy}. 
As shown, the proposed method can improve the predictive ability of CL models, \ie~EfficientNet GEM and EfficientNet DCL, when unlabeled data and corresponding predicted gradients are used.
The loss is decreased and the accuracy is increased.
On the bottom row, the curves of the fitness loss vs. task show that the fitness loss (\ref{eqn:fit_loss}) across tasks is minimized by the proposed gradient learner. 


\begin{figure}[!t]
	\centering
	\subfloat[ResNet GEM 1-PL]{\includegraphics[width=0.23\textwidth]{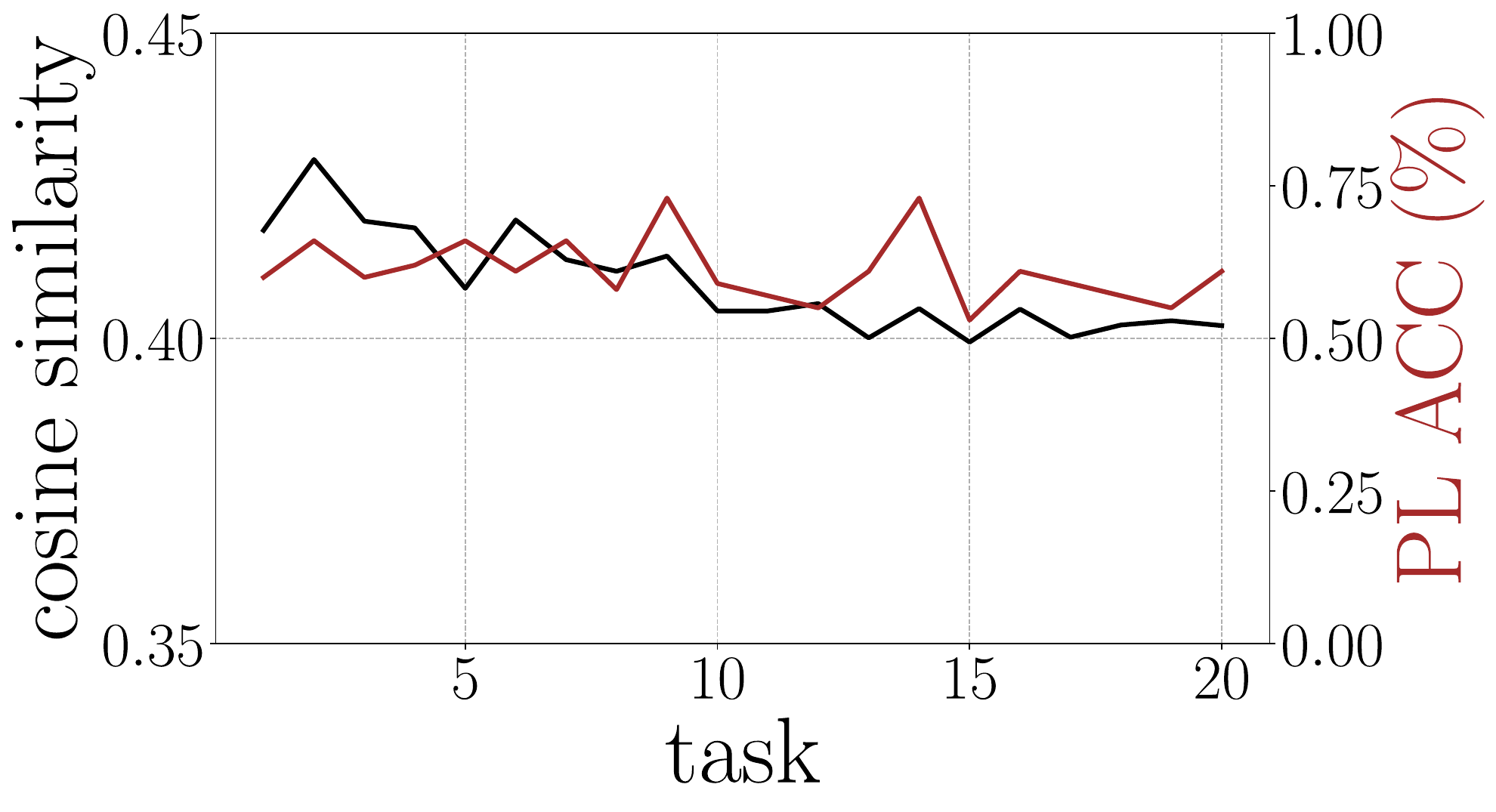}} \hfill
	\subfloat[ResNet DCL 1-PL]{\includegraphics[width=0.23\textwidth]{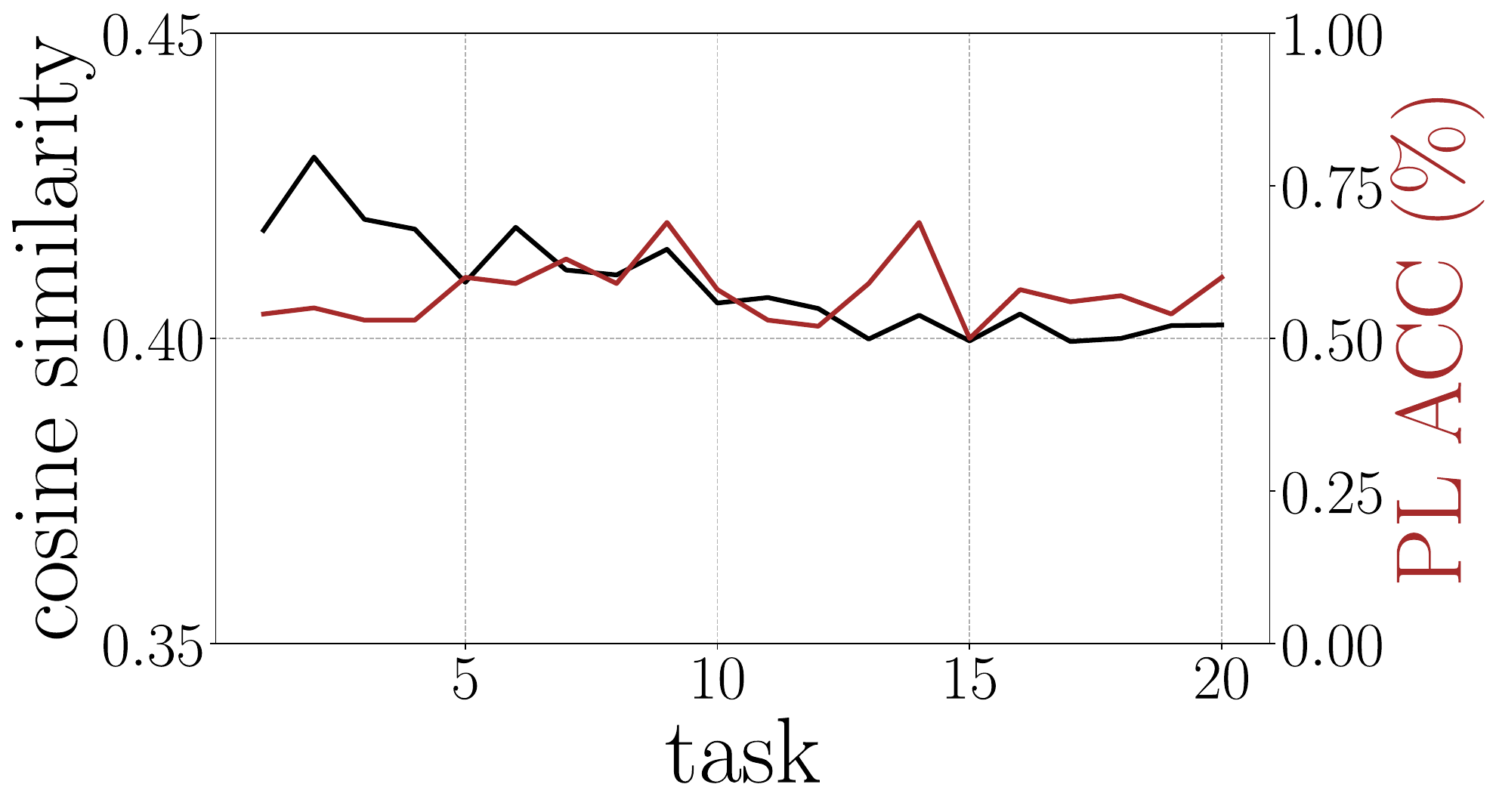}} \\
	\subfloat[ResNet GEM P-PL]{\includegraphics[width=0.23\textwidth]{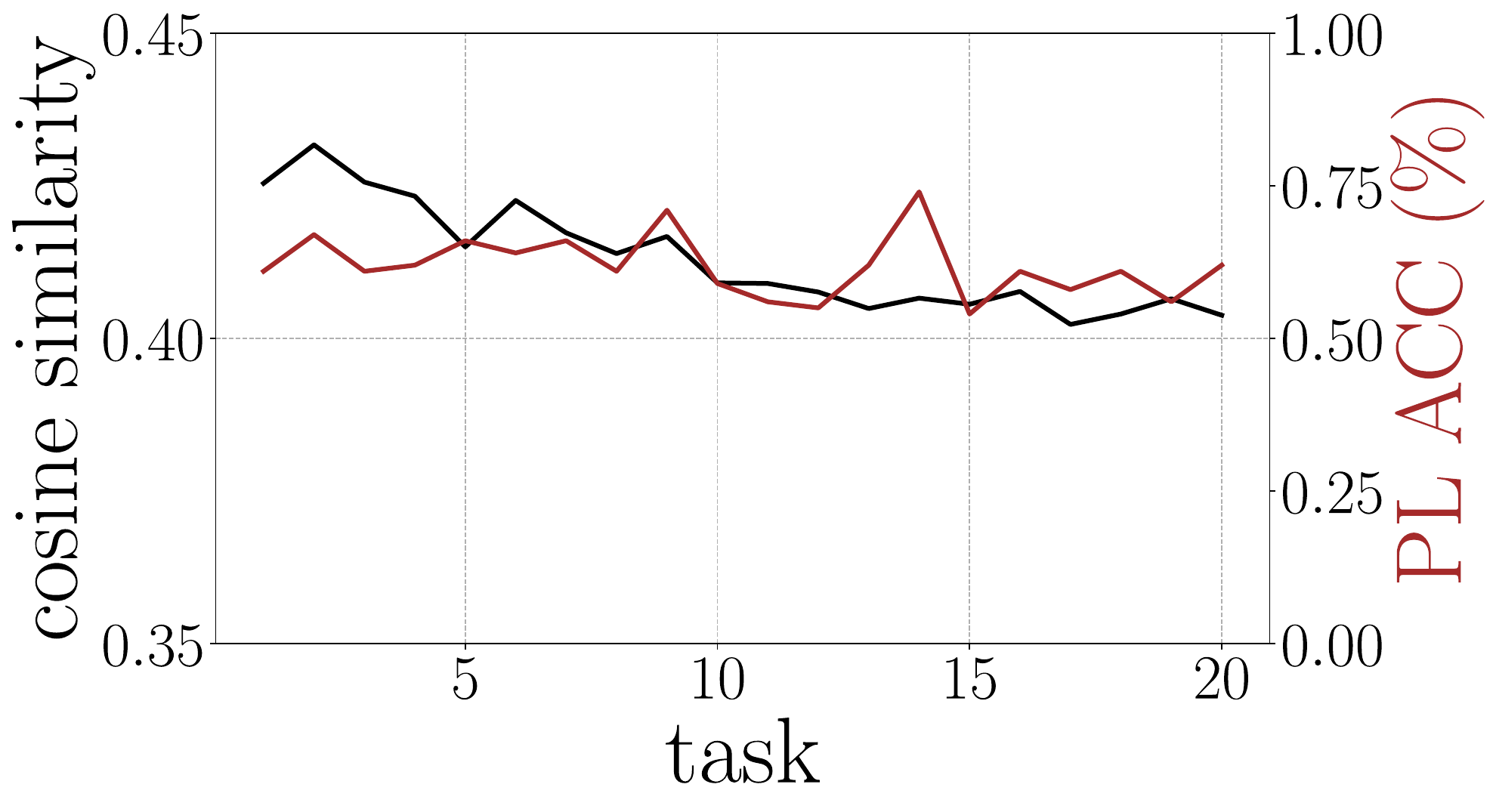}} \hfill
	\subfloat[ResNet DCL P-PL]{\includegraphics[width=0.23\textwidth]{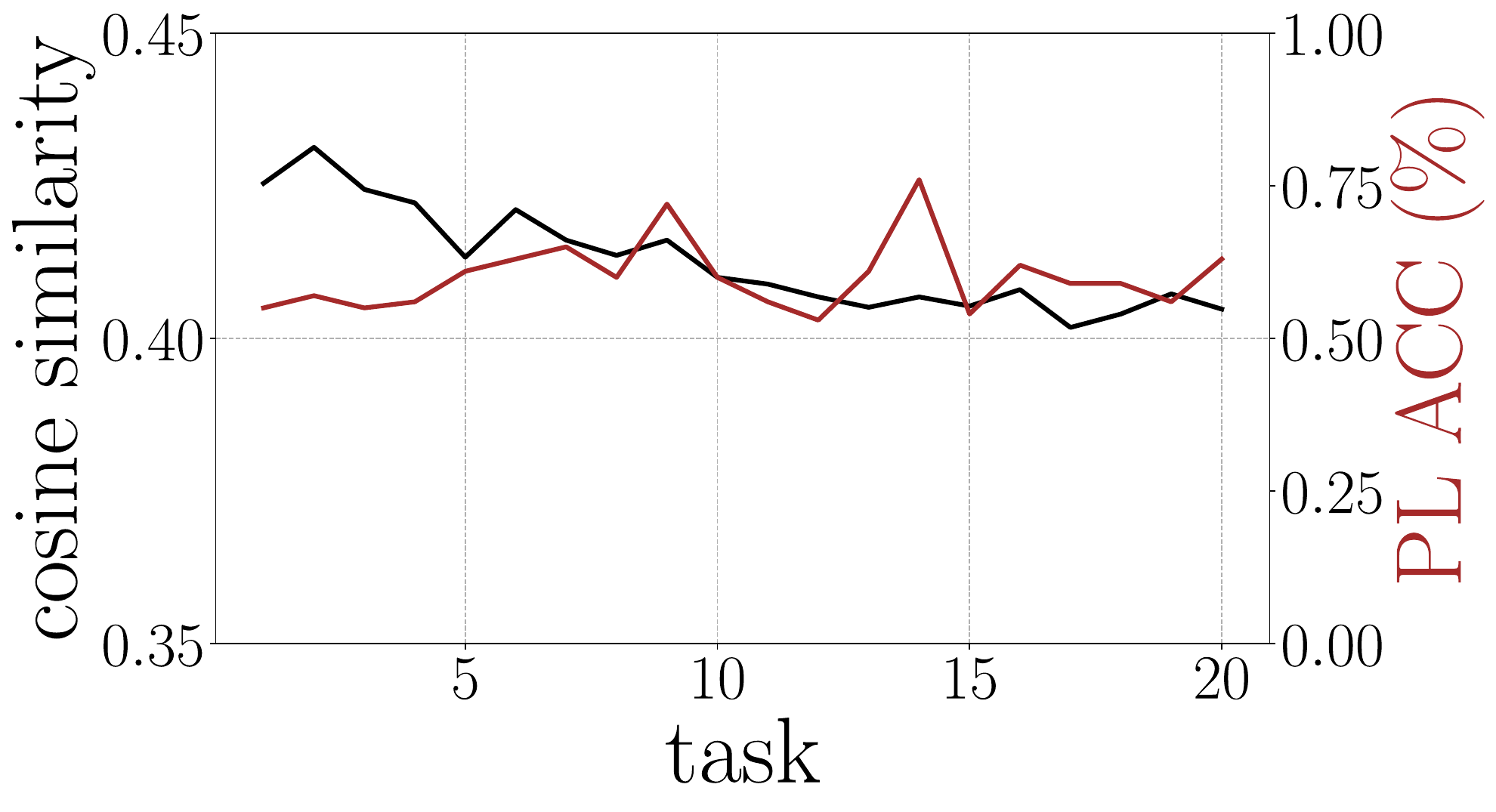}} 
    \vspace{-1ex}
	\caption{\label{fig:cosine}
		Plots of cosine similarity between the gradients generated with ground-truth labels and pseudo labels (black curve), as well as the corresponding pseudo label prediction accuracy (brown curve).
		Due to the lack of training samples and the dynamical change of visual concepts at each task, the pseudo label prediction perform badly (lower than 1\%). 
		This is consistent with the drop on the cosine similarity, which should be 1 if the predicted pseudo labels are correct.
    	}
\end{figure}

\begin{figure}[!t]
	\centering
	\includegraphics[width=0.49\columnwidth]{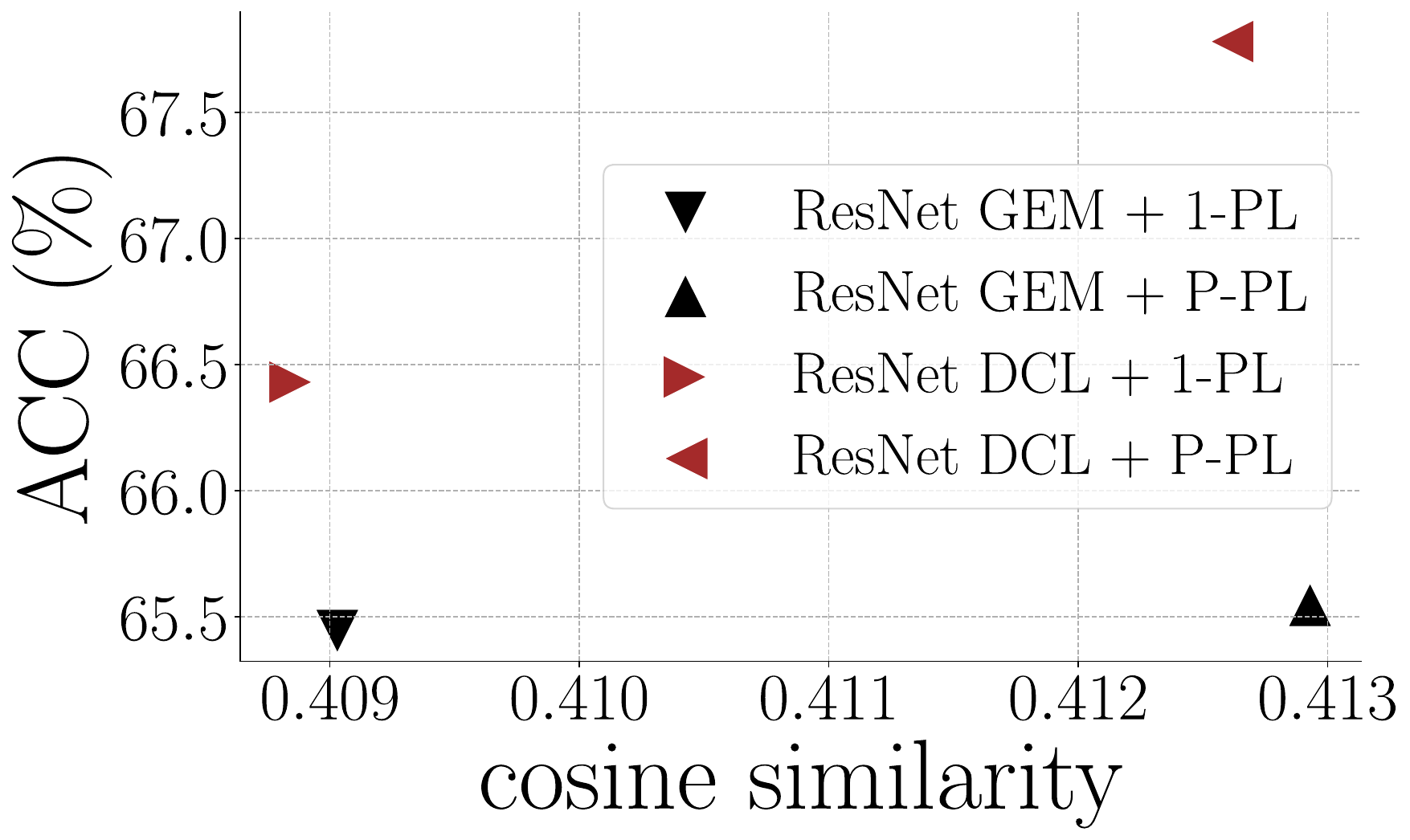}
	\includegraphics[width=0.49\columnwidth]{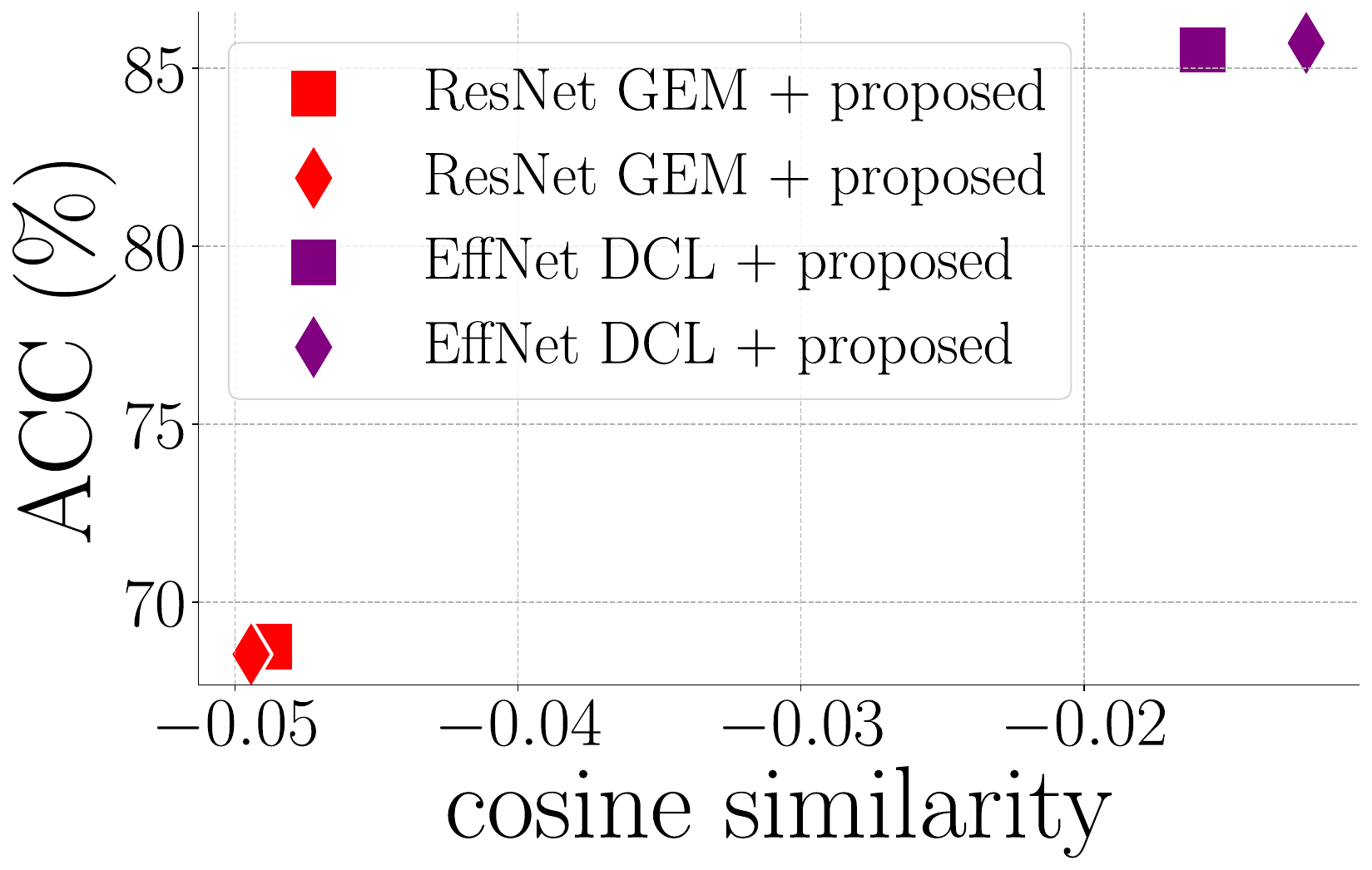}
    \vspace{-1ex}
	\caption{\label{fig:cosine_vs_acc}
		Correlation between cosine similarity and averaged accuracy. 
		Left: Experiments with 1-PL and P-PL; 
		Right: Experiments with the proposed method.
    	}
\end{figure}



\begin{figure*}[!t]
	\centering
	\subfloat[1-PL]{\includegraphics[trim={0 0 28ex 0},clip,width=0.235\textwidth]{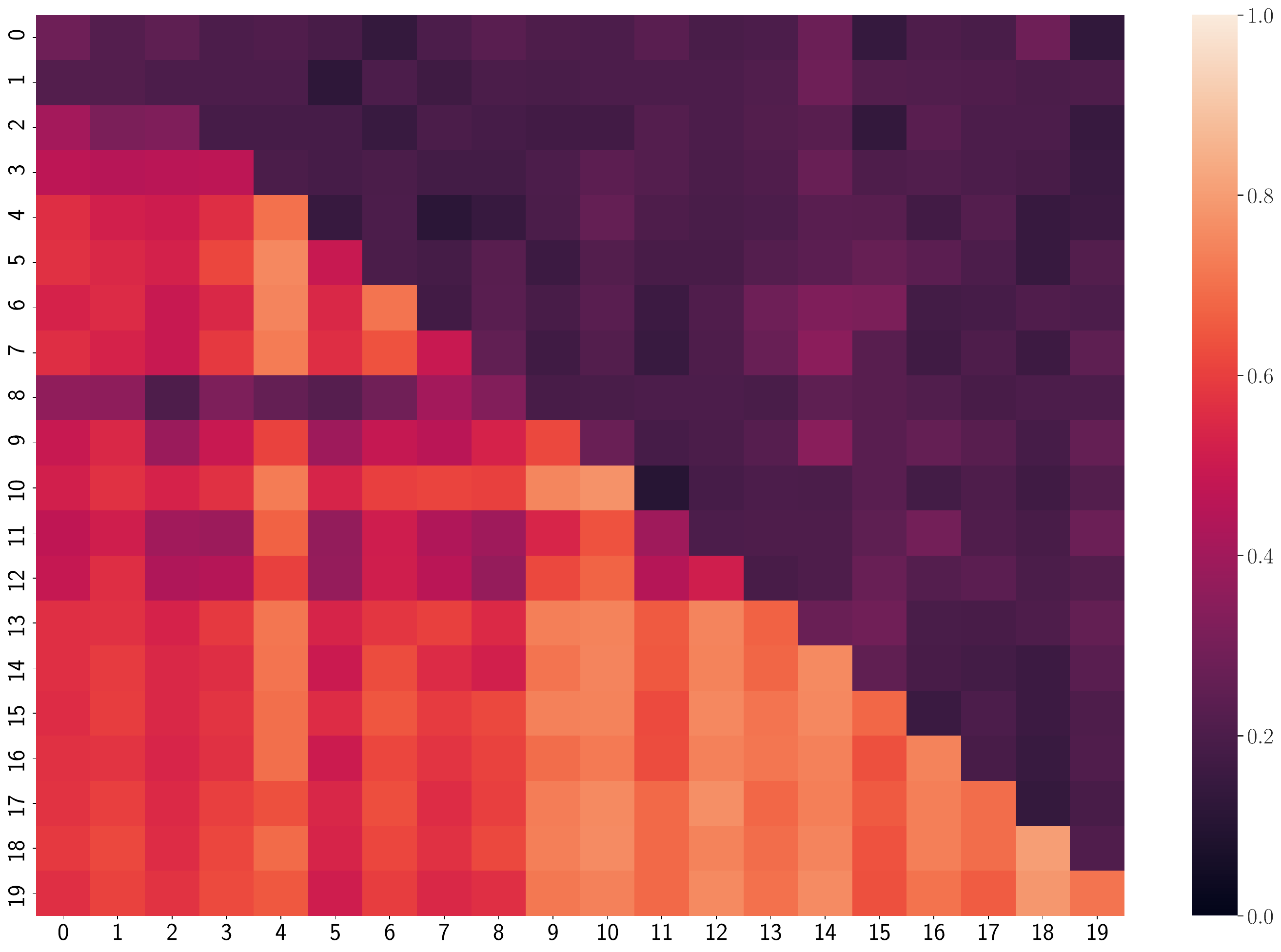}} \hfill
	\subfloat[P-PL]{\includegraphics[trim={0 0 28ex 0},clip,width=0.235\textwidth]{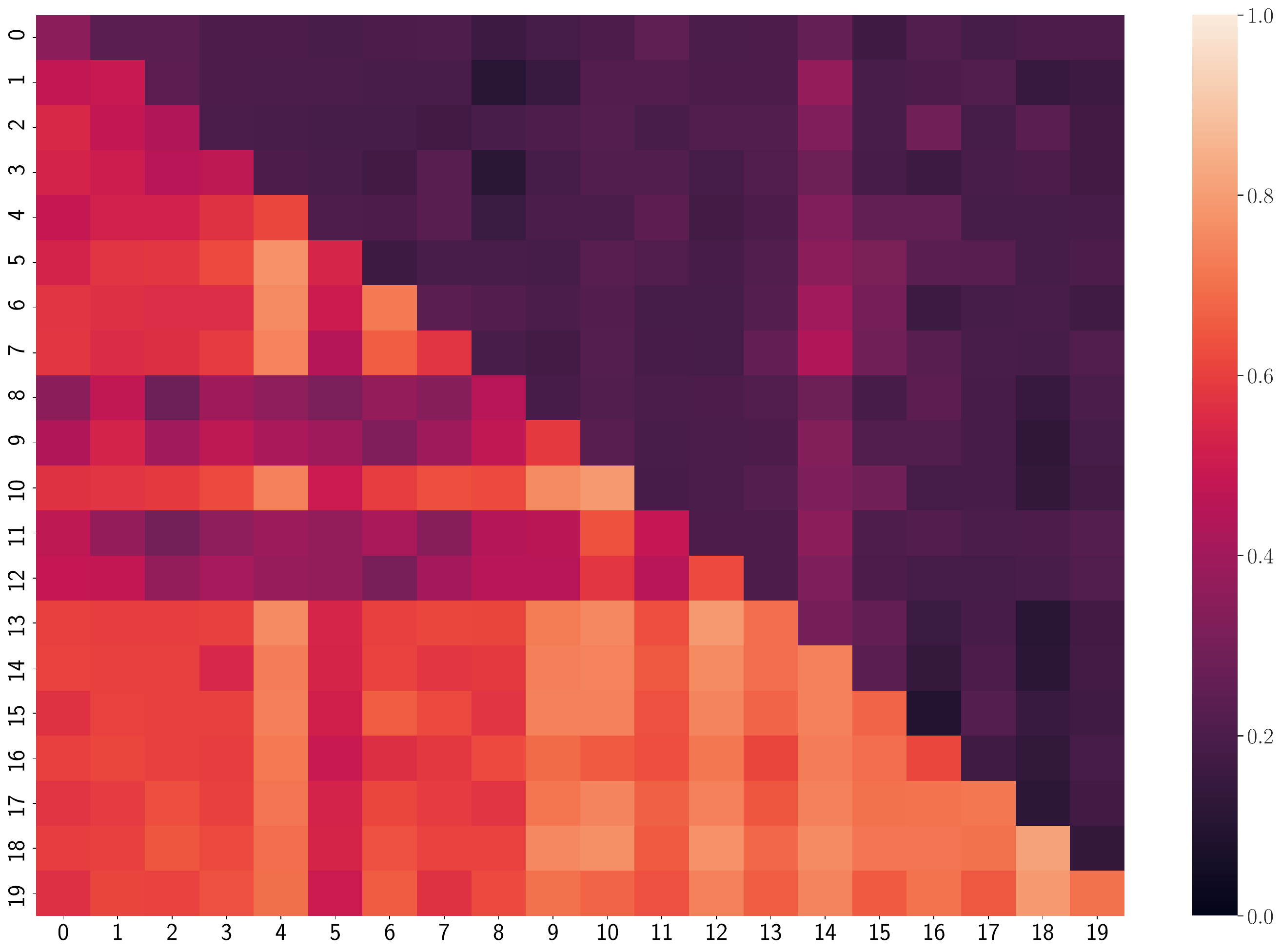}} \hfill
	\subfloat[Baseline]{\includegraphics[trim={0 0 28ex 0},clip,width=0.235\textwidth]{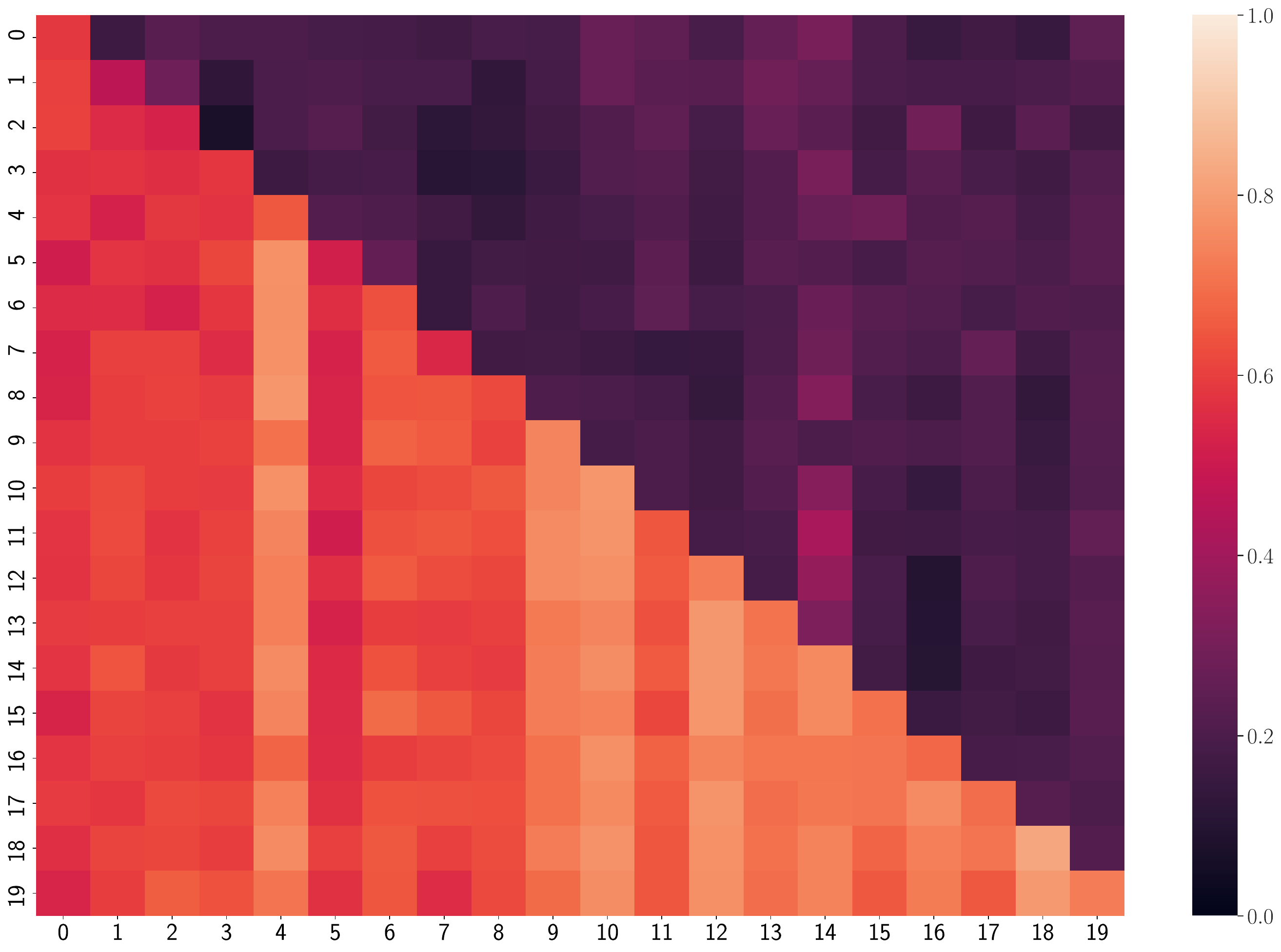}} \hfill
	\subfloat[Proposed]{\includegraphics[width=0.26\textwidth]{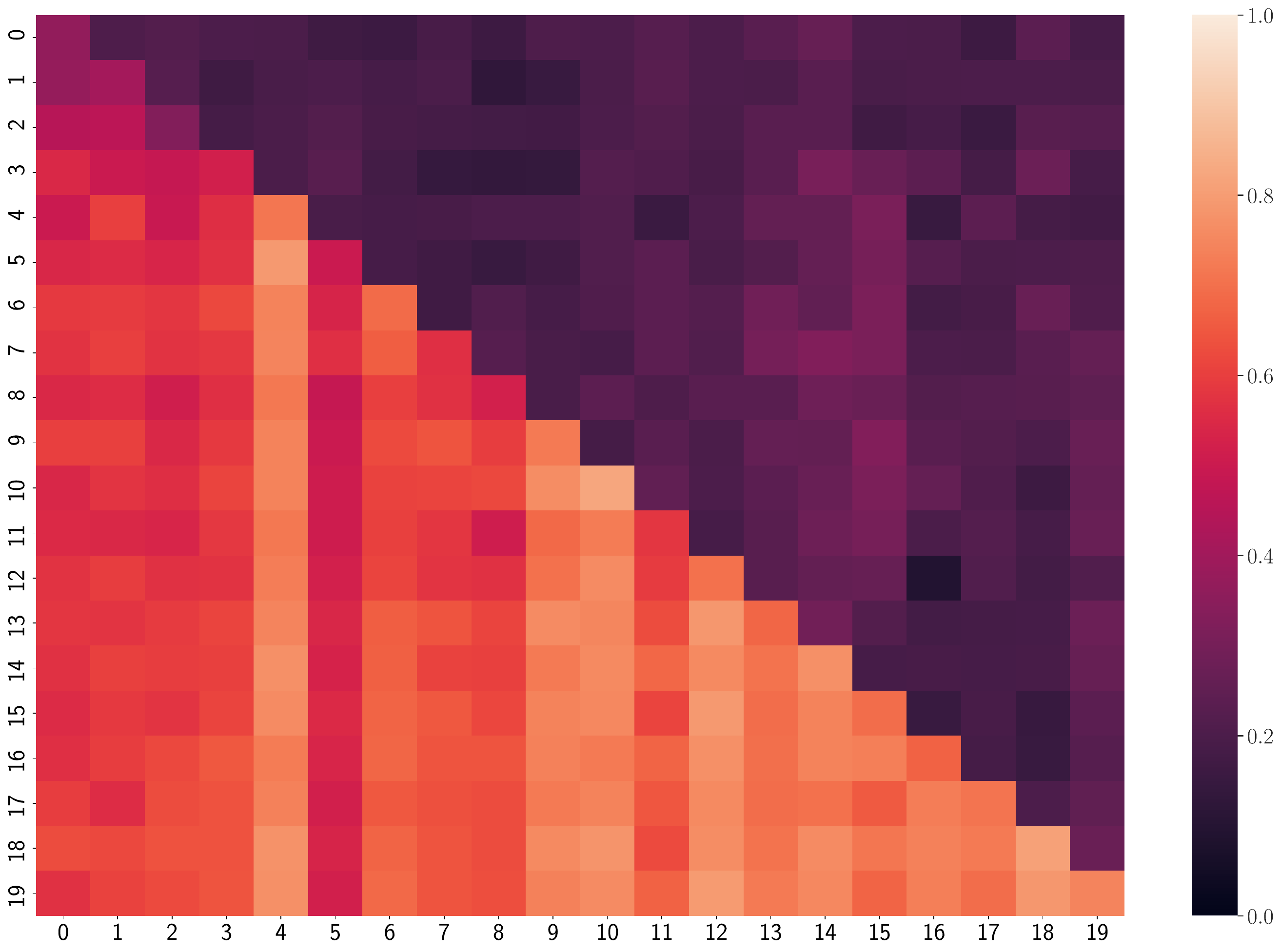}}
    \vspace{-1ex}
	\caption{\label{fig:confusion}
    	Confusion matrix generated by different methods.
    	The $i$-th row indicates it is the training stage with the $i$-th task's samples, while the $i$-th row $j$-th column indicates the model trained with the $i$-th task's samples is evaluated on the $j$-th task. ResNet GEM is used for the analysis.
	}
\end{figure*}


\begin{figure*}[!t]
	\centering
	\subfloat[$\mathcal{N}(0,1)$]{\includegraphics[trim={0 0 28ex 0},clip,width=0.235\textwidth]{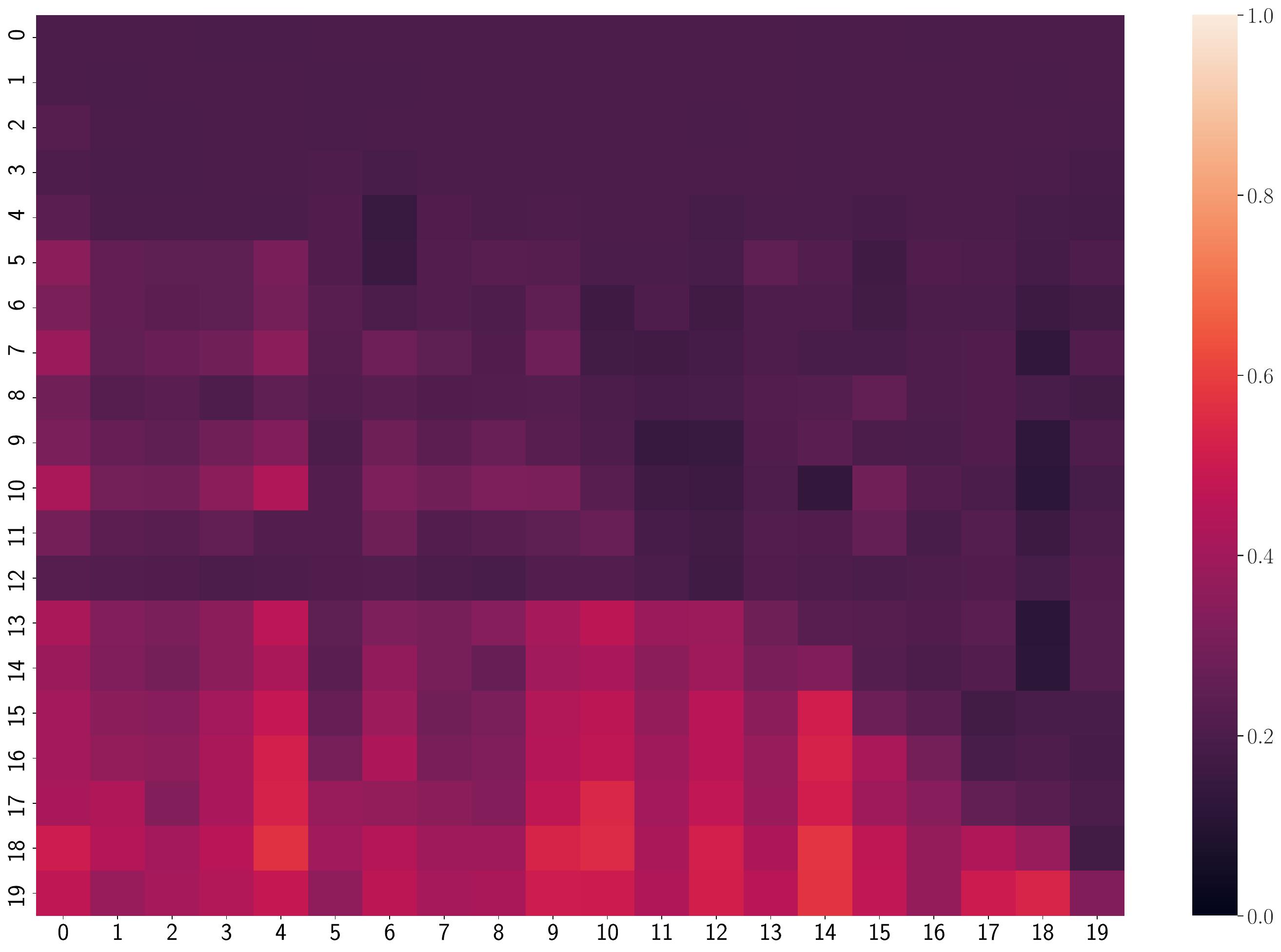}} \hfill
	\subfloat[$\mathcal{N}(0,1)$ + proposed]{\includegraphics[trim={0 0 28ex 0},clip,width=0.235\textwidth]{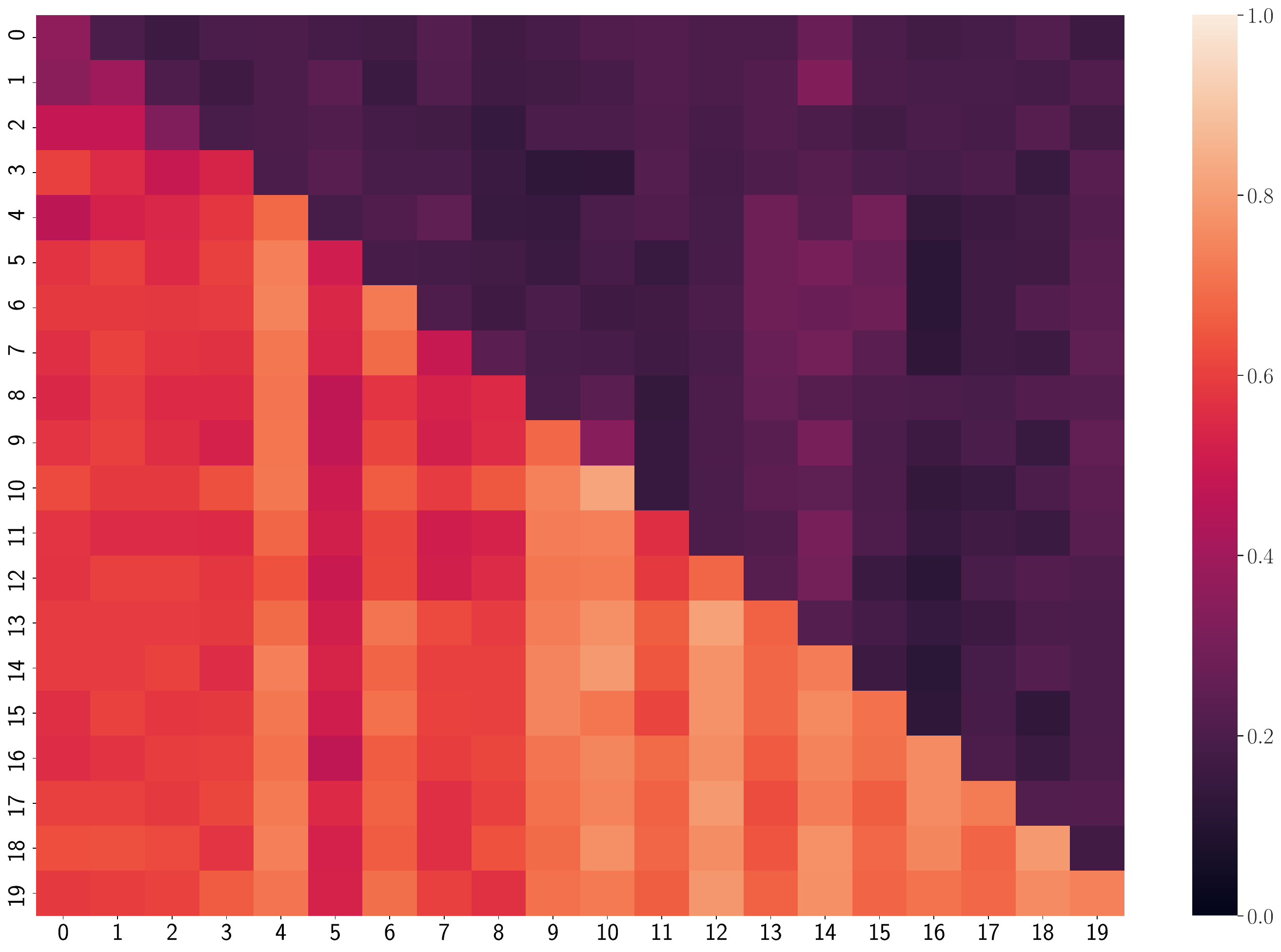}} \hfill
	\subfloat[$\mathcal{U}(-1,1)$]{\includegraphics[trim={0 0 28ex 0},clip,width=0.235\textwidth]{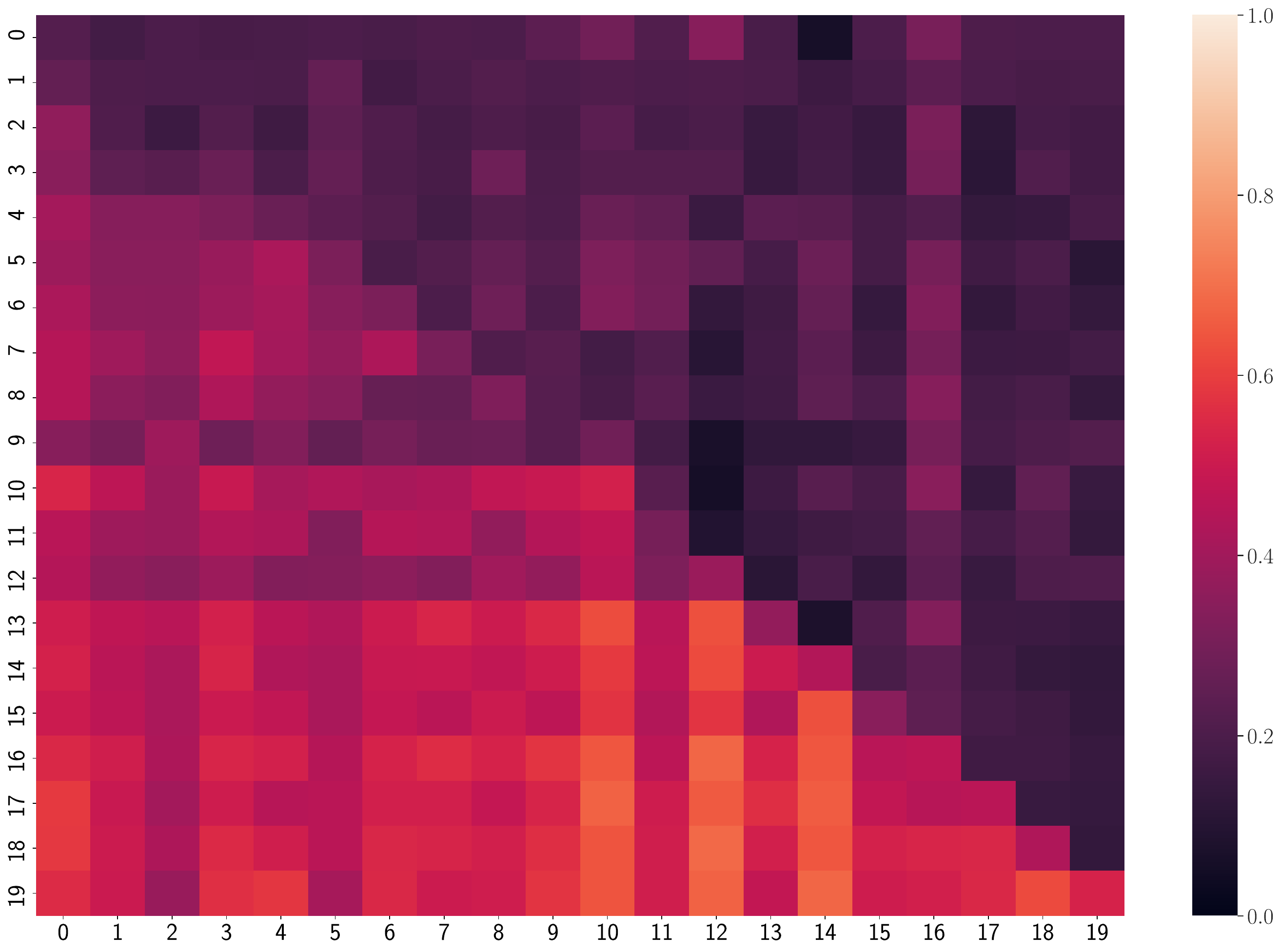}} \hfill
	\subfloat[$\mathcal{U}(-1,1)$ + proposed]{\includegraphics[width=0.26\textwidth]{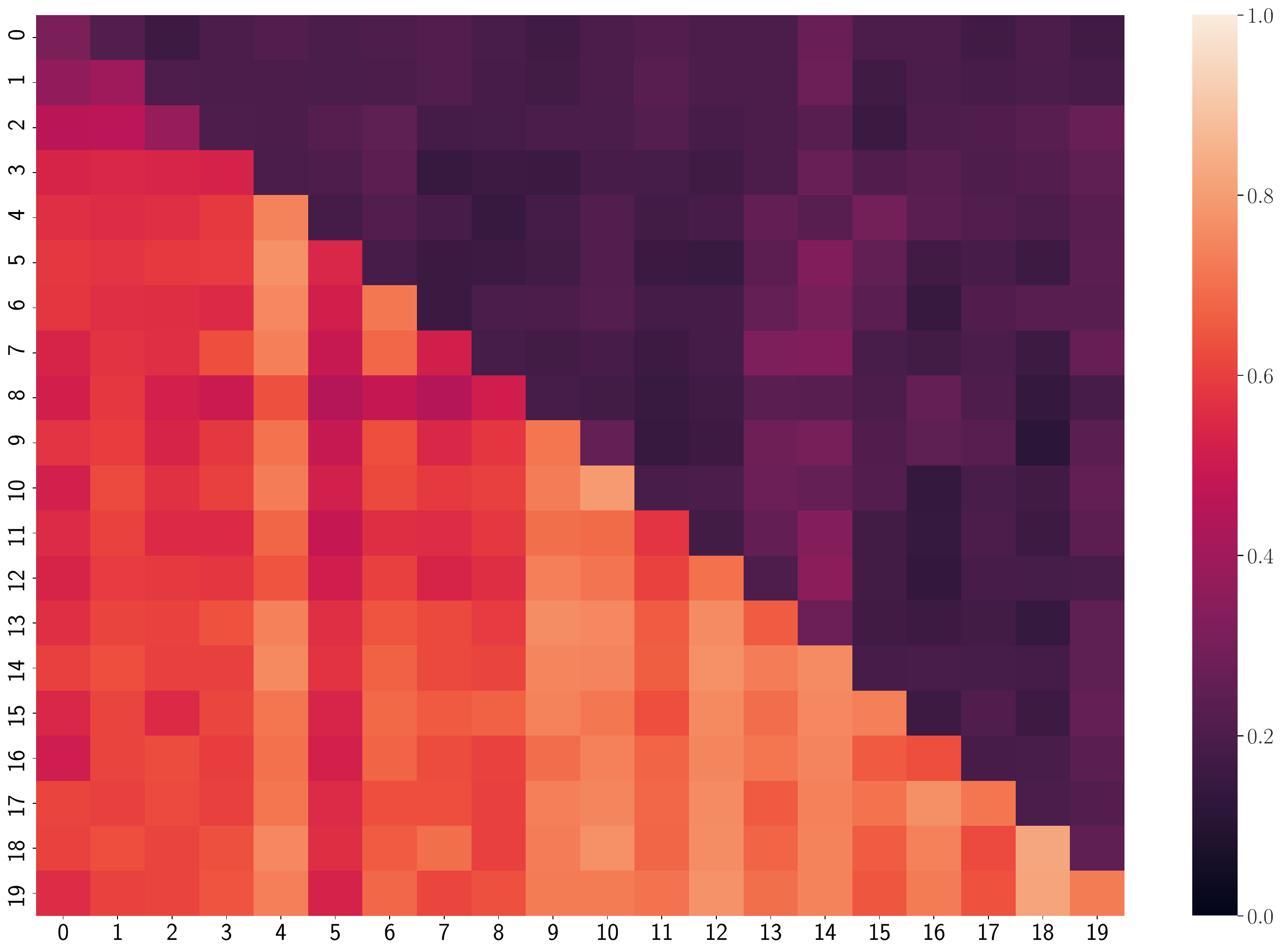}}
    \vspace{-1ex}
	\caption{\label{fig:confusion_n}
    	Confusion matrix generated with random noise. The experimental details are described in Section \ref{subsec:noise}.
	}
\end{figure*}

\subsection{Pseudo Labeling vs. Gradient Prediction}

This section examines how pseudo labeling and gradient prediction work in the CL method. 
Moreover, we investigate the correlations between the gradients generated by various methods and its performance.

To understand the efficacy of pseudo labeling methods, we first inspect the gradients generated with pseudo labels and the accuracy of pseudo label prediction on training samples. 
We take the gradients generated with ground-truth labels as a reference and compute the cosine similarity {\small $\cos(\frac{\partial \ell}{\partial z}|_{x,t,y},\frac{\partial \ell}{\partial z}|_{x,t,\hat{y}})$} between the gradients generated with two types of labels, where $x,t,y$ are training samples and $\hat{y}$ are pseudo labels. 
In this way, the discrepancy can be quantified as the cosine similarity. 
In other words, if pseudo labels are the same as the ground-truth labels, the cosine similarity between the gradients generated with pseudo labels and ground-truth labels should be 1, 
which indicates the resulting gradients are fully aligned.
As shown in \figref{fig:cosine}, the cosine similarities generated by 1-PL and P-PL are stably around 0.4.
The drop from 1 to 0.4 results from the incorrect pseudo labels. 
The accuracies of 1-PL and P-PL are lower than 1\%. The reasons for the low accuracy are two-fold. 
First, in CL, all samples are only observed once and the number of training sample w.r.t. a class is relatively small, \eg~500 on iCIFAR-100. 
Thus, there is no enough data to train a high-performance teacher model. 
Second, the classes of samples are used for training at a task are distinct from that of the other tasks. This dynamic results in the difficulty to train a strong teacher model.

Next, we examine how the gradients generated by various methods correlate with the performance.
Note that the predicted gradients aim to minimize the fitness loss (\ref{eqn:fit_loss}), while the pseudo labeling methods aim to maximize the similarity between the gradients generated with pseudo labels and ground-truth gradients labels. 
Hence, the predicted gradients are expected to differ from the ground-truth generated gradients. 
As shown in \figref{fig:cosine_vs_acc}, ResNet GEM + P-PL yields a higher cosine similarity than ResNet DCL + P-PL, but achieves a lower accuracy. 
In contrast, the proposed method's (\ie~with gradient prediction) accuracy is clearly proportional to the cosine similarity.
On the other hand, we observe that discriminative features produced by a strong backbone will lead to better predicted gradients in terms of the geometric relationship.

\begin{table}[!t]
	\centering
	\caption{\label{tbl:conn_cifar_runtime}
	    Computational complexity on iCIFAR-100. 
        }
	\adjustbox{width=0.9\columnwidth}{
	\begin{tabular}{L{24ex} C{14ex} C{14ex}}
		\toprule
		Methods & Training Time (ms/Image) & GPU Mem (MB/Image)  \\
		\cmidrule(lr){1-1} \cmidrule(lr){2-2} \cmidrule(lr){3-3} 
		ResNet GEM + 1-PL & 25 & 194  \\
		ResNet GEM + P-PL & 25 & 194   \\
		ResNet GEM & 18 &  193  \\
		ResNet GEM + MG & 14 &  194    \\
		ResNet GEM + proposed & 19 &  193   \\ \midrule
            ResNet DCL + 1-PL & 24 & 194 \\
		ResNet DCL + P-PL & 25 & 194 \\
		ResNet DCL & 15 & 193  \\
		ResNet DCL + MG & 13 & 193 \\
		ResNet DCL + proposed & 16 & 193 \\ 
        \midrule
		EffNet GEM + 1-PL & 108 &  753  \\
		EffNet GEM + P-PL & 109 &  753  \\
		EffNet GEM & 60 & 756 \\
		EffNet GEM + MG & 61 & 750 \\
		EffNet GEM + proposed & 62 & 756 \\ \midrule
            EffNet DCL + 1-PL & 108 & 763 \\
		EffNet DCL + P-PL & 108 & 763 \\
		EffNet DCL & 57 & 759 \\
		EffNet DCL + MG & 56 & 757 \\
		EffNet DCL + proposed & 61 & 759 \\ 
		\bottomrule	
	\end{tabular}}
\end{table}

\subsection{Computational Complexity for Training}

\REVISION{
\tabref{tbl:conn_cifar_runtime} reports the runtime and GPU memory for training models. For ResNet models, the training time per image ranges from 13-25 ms, with MG being fastest and P-PL being slowest. For EfficientNet models, training is generally slower, ranging from 56-109 ms per image. GPU memory usage per image is around 190-200 MB for ResNet models and 750-760 MB for EfficientNet. There is little difference between methods. In general, MG and the proposed method are faster than the other methods (\ie baseline, 1-PL, and P-PL). In particular, the proposed gradient learner method has comparable speed to MG for both ResNet and EfficientNet models, while using slightly less memory.
}

\section{Confusion Matrix}

To comprehensively understand the efficacy of the proposed predicted gradients, we visualize the confusion matrices generated by various methods on iCIFAR-100 in \figref{fig:confusion} and \ref{fig:confusion_n}. The $i$-th row of the confusion matrix indicates the test classification accuracies over 20 tasks with the model trained on the $i$-th task. Similarly, the $j$-th column indicates the results are evaluated on the test set of the $j$-th task.

As shown in \figref{fig:confusion}, leveraging extra unlabeled images with the proposed method will have lower accuracies on early tasks than the baseline as the proposed gradient learner does not have sufficient training samples for learning. With more and more training samples being observed, better predicted gradients are produced to improve the performance on late tasks. In addition, the accuracies of 1-PL and P-PL are overall lower due to the disturbance caused by the incorrect pseudo labels. As discussed in \secref{subsec:perf}, low $R_{i,i}$ leads to a high BWT score.

\tabref{tbl:noise} shows that using random noise as predicted gradients yields higher BWT than the other settings. Again, this is because the random noise disturbs the learning process, which leads to low accuracies (see \figref{fig:confusion_n}). 
More importantly, \figref{fig:confusion_n} shows that random noise + proposed is more robust than method with only random noise.

\section{Conclusion}

In this work, we study how to exploit the semantics of the unlabeled data to improve the generalizability of CL methods.
Exsisting semi-supervised (continual) learning presumes that the labels associated with unlabeled data are known to the learning process. We relax the constraint, \ie the labels associated with unlabeled data could be known or unknown to the learning process.
Correspondingly, we propose a new SSCL method, where a novel gradient learner is trained with labeled data and utilized to generate pseudo gradients when the input label is absent. 
The proposed method is evaluated in the CL and ACL settings. The experimental results show that the average accuracy and backward transfer are both improved by the proposed method and achieve state-of-the-art performance.
This implies that utilizing the semantics of the unlabeled data improves the generalizability of the model and alleviates catastrophic forgetting.
Last but not least, we provide empirical evidence to show that the proposed method can generalize to the semi-supervised learning task.

\ifCLASSOPTIONcompsoc
 \section*{Acknowledgments}
\else
 \section*{Acknowledgment}
\fi
This research is supported in part by the NSF under Grants 1908711 and 2143197, and in part by the National Research Foundation, Singapore under its Strategic Capability Research Centres Funding Initiative. Any opinions, findings, and conclusions or recommendations expressed in this material are those of the author(s) and do not reflect the views of National Research Foundation, Singapore.

\ifCLASSOPTIONcaptionsoff
  \newpage
\fi

\bibliographystyle{IEEEtran}
\bibliography{sec/references}

\begin{IEEEbiography}[{\includegraphics[width=1in,height=1.25in,clip,keepaspectratio]{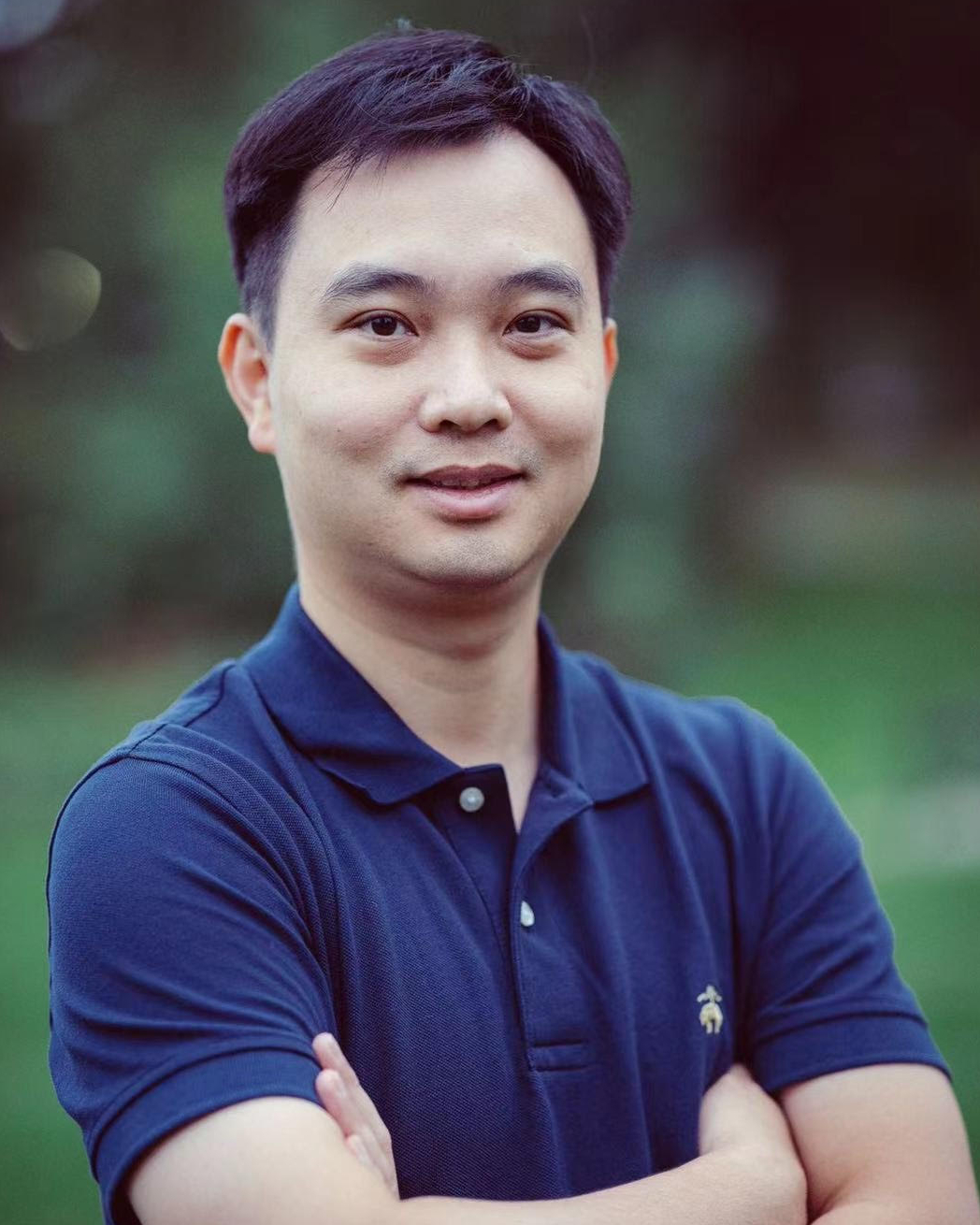}}]{Yan Luo}
    is currently a postdoctoral research fellow at the Harvard Ophthalmology AI lab. He  obtained a Ph.D. degree from the University of Minnesota (UMN), Twin Cities. Prior to UMN, he joined the Sensor-enhanced Social Media (SeSaMe) Centre, Interactive and Digital Media Institute at the National University of Singapore (NUS), as a Research Assistant. Also, he joined the Visual Information Processing Laboratory at the NUS as a Ph.D. Student. He received a B.Sc. degree in computer science from Xi'an University of Science and Technology. He worked in the industry for several years on a distributed system. His research interests include responsible AI and equitable deep learning. He is a member of the IEEE since 2023.
\end{IEEEbiography}


\begin{IEEEbiography}[{\includegraphics[width=1in,height=1.25in,clip,keepaspectratio]{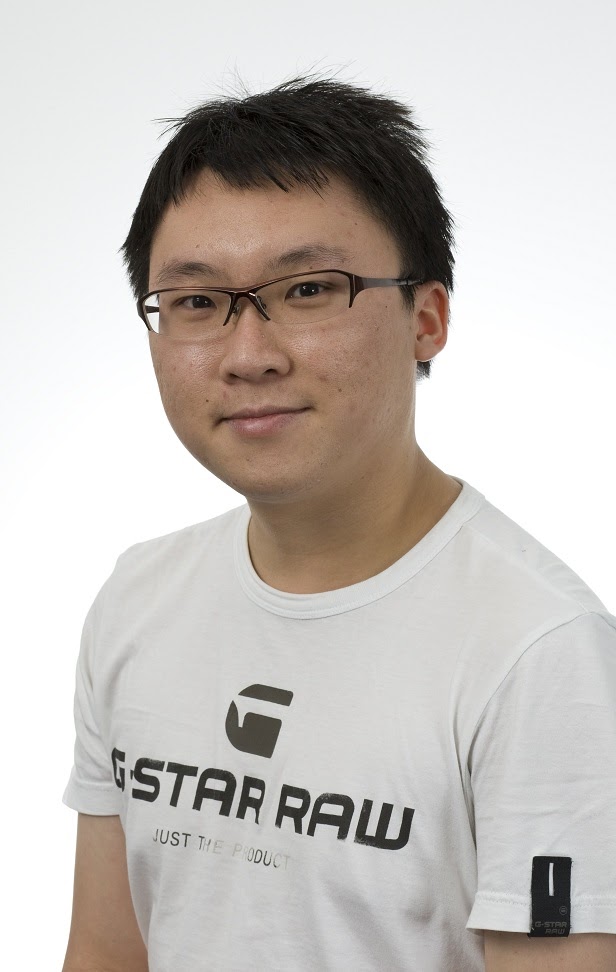}}]{Yongkang Wong} is a senior research fellow at the School of Computing, National University of Singapore. He is also the Assistant Director of the NUS Centre for Research in Privacy Technologies (N-CRiPT). He obtained his BEng from the University of Adelaide and PhD from the University of Queensland. He has worked as a graduate researcher at NICTA’s Queensland laboratory, Brisbane, QLD, Australia, from 2008 to 2012. His current research interests are in the areas of Image/Video Processing, Machine Learning, Trusted Multimodal Analysis, and Human Centric Analysis. He is a member of the IEEE since 2009.
\end{IEEEbiography}

\begin{IEEEbiography}[{\includegraphics[width=1in,height=1.25in,clip,keepaspectratio]{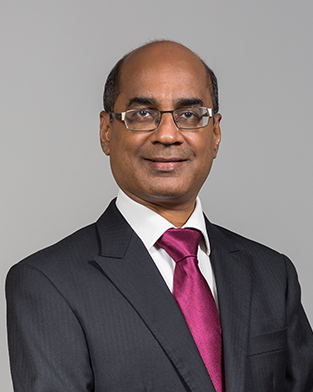}}]{Mohan~Kankanhalli} is Provost's Chair Professor of Computer Science at the National University of Singapore (NUS). He is the Deputy Executive Chairman of AI Singapore Dean of NUS School of Computing and he also directs N-CRiPT (NUS Centre for Research in Privacy Technologies) which conducts research on privacy on structured as well as unstructured (multimedia, sensors, IoT) data. Mohan obtained his BTech from IIT Kharagpur and MS \& PhD from the Rensselaer Polytechnic Institute. Mohan’s research interests are in Multimedia Computing, Computer Vision, Information Security \& Privacy and Image/Video Processing. He has made many contributions in the area of multimedia \& vision – image and video understanding, data fusion, visual saliency as well as in multimedia security – content authentication and privacy, multi-camera surveillance. Mohan is a Fellow of IEEE.
\end{IEEEbiography}

\begin{IEEEbiography}[{\includegraphics[width=1in,height=1.25in,clip,keepaspectratio]{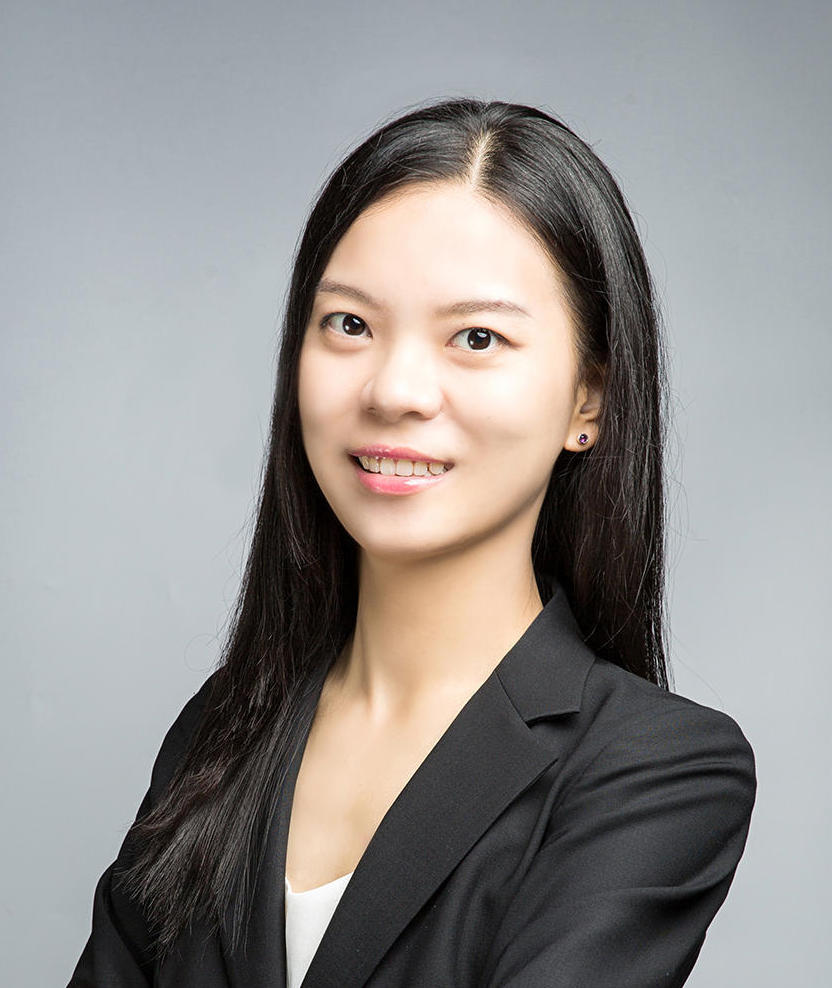}}]{Qi Zhao}
	is an associate professor in the Department of Computer Science and Engineering at the University of Minnesota, Twin Cities. Her main research interests include computer vision, machine learning, cognitive neuroscience, and healthcare. She received her Ph.D. in computer engineering from the University of California, Santa Cruz in 2009. She was a postdoctoral researcher in the Computation \& Neural Systems at the California Institute of Technology from 2009 to 2011. Before joining the University of Minnesota, Qi was an assistant professor in the Department of Electrical and Computer Engineering and the Department of Ophthalmology at the National University of Singapore. She has published more than 100 journal and conference papers in computer vision, machine learning, and neuroscience venues, and edited a book with Springer, titled Computational and Cognitive Neuroscience of Vision, that provides a systematic and comprehensive overview of vision from various perspectives. She serves as an associate editor of IEEE TNNLS, IEEE TMM, and IEEE TCDS, as a program chair WACV’22, and as an organizer and/or area chair for CVPR and other major venues in computer vision and AI regularly. She is a senior member of IEEE.
\end{IEEEbiography}


\end{document}